\documentclass[journal]{IEEEtran}

\usepackage[T1]{fontenc}
\usepackage[utf8]{inputenc}
\usepackage{amsmath,amssymb,amsfonts}
\usepackage{graphicx}
\usepackage{booktabs}
\usepackage{tabularx}
\usepackage{array}
\usepackage{caption}
\usepackage{textcomp}
\usepackage{url}
\usepackage{flushend}
\usepackage{morefloats}
\usepackage{adjustbox}
\usepackage{microtype}

\graphicspath{{figures/}}

\title{Transformation Laws in Neural Representations:\\
Structure, Realisability, and Construction}

\author{Yuan Sun%
\thanks{The author is with the School of Mathematical Sciences, Beijing Normal University,
Beijing 100875, China (e-mail: 3396897122@qq.com).}}

\usepackage[hidelinks,pdfusetitle]{hyperref}
\hypersetup{pdfauthor={Yuan Sun}}

\begin{document}
\maketitle


\begin{abstract}

How neural representations preserve the structure of input changes connects representation analysis with internal intervention. We study operable representational content through compatible actions of reference transformations on neural features. We characterise when a transformation descends through an encoder, and give a linear setting in which the defect is governed by the transformation's demand for discarded information, measured in the metric the representation induces. On a rectifier the failure to realise a transformation has two distinguishable sources --- what the source region has already made unrecoverable, and what it costs to satisfy every region the transformation visits with one operator --- and for a \textit{measured} harmonic carrier the same question has a closed answer: a linear realisation exists exactly when the retained harmonic blocks are invariant under the action. Using colour as the in-depth instance, we find that hue orbits in frozen visual features concentrate 84--88\% of their energy in the first two harmonics with rotation planes shared across shapes, that this organisation is substantially inherited from input and architecture and is reshaped by training and depth, and that the measured structure supports prediction, transport from new starting states, and composition --- with global and local realisations differing sharply in which they achieve. Guided by the measurements, we construct a compact interface whose rotation action is fixed by the structure and never fitted: it reads hue zero-shot at 3.4$^\circ$ median error on unseen shapes. Theory, structural measurement, and construction together establish transformation laws as a concrete object connecting the understanding of neural representations to their design.

\end{abstract}


\section{Introduction}

\subsection{The operational question}

A network has encoded an object. Can we act on its internal state alone, and obtain the computation the network would have produced had the object's colour actually changed? The question is well posed because input-side changes come with their own algebra: hue rotates on a circle (a group, $SO(2)$), heat diffuses by $B_t=e^{-tL}$ with $B_tB_s=B_{t+s}$ (a semigroup), and saturation or value rescale until they clip (monoids, with degenerate points where fibres collapse). World states transform as $s\mapsto\tau s$; imaging $R$ and the first $\ell$ layers of a network compose into $\psi_\ell:\mathcal S\to Z_\ell$. The question is whether the algebra survives the trip. We ask for a homomorphism between two algebras of transformations --- the physical family $\tau\in M$ with its composition law, and operators on the feature space:
\begin{equation*}\adjustbox{max width=\columnwidth}{$\displaystyle \begin{aligned}
&\rho_\ell:M\to\operatorname{End}(Z_\ell),\qquad \psi_\ell(\tau s)\approx\rho_\ell(\tau)\psi_\ell(s),\\
&\rho_\ell(\tau_2\tau_1)=\rho_\ell(\tau_2)\circ\rho_\ell(\tau_1),\qquad \rho_\ell(e)=I,
\end{aligned}\tag{$\star$}
$}\end{equation*}
that is, the square drawn in Figure 1(a).

\begin{figure*}[t]
\centering
\includegraphics[width=\textwidth]{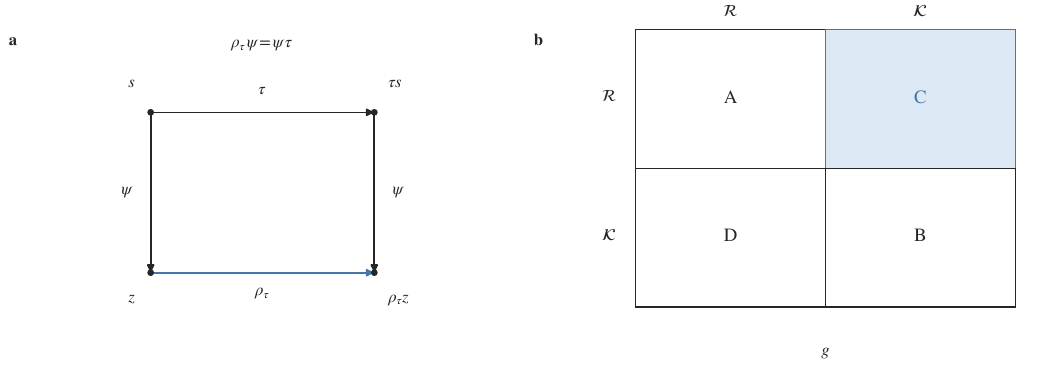}
\caption*{\textbf{Figure 1.} The correspondence between input transformations and feature-space operators. (a) A physical transformation $\tau$ acts on the state space and the encoder $\psi$ maps states to features; the operator $\rho_\tau$ is a representation of $\tau$ on the feature space exactly when the diagram commutes, $\rho_\tau\psi=\psi\tau$, which also delivers the composition law. (b) Four-block structure of the map induced by $\tau$ when the encoder is linear, in the basis that splits the input into retained ($\mathcal R$) and discarded ($\mathcal K$) coordinates. Blocks $A$ and $B$ act inside one subspace, $D$ maps retained coordinates into the discarded one, and $C$ maps discarded coordinates back into the retained subspace; only $C$ can make a feature-space action fail to exist, because it is the only block whose contribution cannot be recovered from the retained features.}
\end{figure*}

The object of study is therefore not a collection of isolated feature edits but a \textit{representation of the transformation algebra}: the composition law must survive into the feature space, not only the individual actions. Does such a $\rho_\ell$ exist, at which layers, in what form, and when does it degrade?

\subsection{Three results}

\textbf{Existence is an information condition, and its failure is priced twice over.} An induced action exists at a site exactly when the transformation preserves that site's fibres: states the encoder has merged must stay merged after the transformation. Existence is therefore a property of the encoder--transformation pair, and it is a single condition rather than two, because once the action is well defined its composition law is inherited from the physical one. In the linear layer the same condition has an explicit form --- a realising operator exists iff the transformation preserves the encoder's kernel --- and its four-block decomposition shows that \textit{only} the block that routes discarded information back into the retained subspace obstructs: discarding information is free, contradicting it is not. When a realising operator exists it is unique, being the solution of $\rho W=Wg$ with $W$ of full row rank; when it does not, the best achievable score is set by the mixing ratio \textit{in the metric the representation induces}, $1-T_{RMS}=\kappa_W/\sqrt{1+\kappa_W^2}$ under stated moment conditions, verified to $0.0091$ over a $\kappa$ sweep spanning sixteen orders of magnitude. Two consequences are counter-intuitive and both are measured: only mixing-back is poisonous, and retained rank is nearly irrelevant --- $\rho(\kappa,T_F)=-0.968$ against $\rho(\text{effective rank},T_F)=+0.502$ on trained weights. The law's tightness condition is explicit, and real backbones do not meet it; the paper reports the deviation rather than presenting a controlled-regime identity as a calibrated predictor. On a rectifier the failure to realise has two distinguishable sources: what the source region has already made unrecoverable on a transition cell, and what it costs to satisfy every cell the transformation visits with one operator. \textit{Earned at} \S{}3.2--3.5, \S{}8.3.

\textbf{The measured organisation fixes which realisations can exist.} A continuous rotation action on features is block-diagonal in a suitable basis, so the measurement question is which blocks carry the energy and whether their planes are shared. For colour, the first two harmonics carry $84$--$88\%$ of the orbit energy and the planes are shared across shapes at $\cos\ge0.96$, and that form is prescriptive: for a carrier $B$ with action $A_\Delta$, a fixed linear realisation exists iff $A_\Delta\ker B\subseteq\ker B$, and then the least-squares optimal map is $K^\star=BA_\Delta B^\dagger$, whose residual is the energy of the retained-block directions the rotation carries back into view. The criterion is a \textit{design} condition for a compiled carrier --- it says which coordinate sets may be kept and what a chosen truncation costs. On the measured features the synthesis is injective, so closure holds exactly and the residual failure of a global fit is attributable to content dependence and tail instead; the paper keeps that distinction rather than blurring it. \textit{Earned at} \S{}4.2--4.6.

\textbf{From the correspondence to operation: what it costs, and what it builds.} Global linear families fail where a local, phase-conditioned transporter succeeds; chains compose within a bound set by the truth action's Lipschitz constant in the consumer's seminorm, at a price that is measured rather than assumed; and the measured form fixes an interface whose action is never fitted, whose error has three named sources of which only one is removed by training, and whose operating domain is published as a threshold rule. \textit{Earned at} \S{}5, \S{}7.

The rest of the paper develops these three results, measures the structure that fixes the realisable forms, and compiles that structure into an interface.

\subsection{Reading is not transforming}

The question is independent of attribute readout. Take $x=(u,v)$, let $f$ keep $u$, let the task read $u$, and let $\tau$ swap $u$ and $v$: the representation is perfectly task-sufficient --- the attribute reads out exactly --- and completely transformation-insufficient, because the transformed state is unrecoverable from what was kept. Reading an attribute and carrying its transformation law are different properties with different conditions, and the standard instruments of representation analysis measure the former. The symptom is sharper than a gap in coverage: a random orthogonal operator, fitted to nothing, can attain a higher projection score against a true hue rotation than a correctly fitted operator while destroying downstream behaviour entirely (\S{}5.5). Geometric proxies cannot by themselves tell a correspondence from a decoy, which is why every capability claim in this paper is graded against ground-truth targets and downstream behaviour.

\subsection{What the measurements find}

Colour gives the question a concrete instance. Densely sampled hue orbits in frozen vision features concentrate $84$--$88\%$ of their energy in the first two harmonics, with rotation planes shared across shapes (principal-angle cosine at least $0.96$) while amplitudes carry the shape-specific content; much of this is inherited from the input and from architecture rather than learned, and it erodes with depth. The organisation is strong enough to act on: operators built from it predict the features of genuinely re-rendered hue-rotated inputs to $0.17$--$0.40\%$ relative error, more than an order of magnitude tighter than the copy baseline. Three questions follow: why does this structure exist, which realisations can act on it, and what do training and depth do to it?

\subsection{From structure to implementation}

Predicting an orbit from a fixed anchor, transporting from a starting state never seen, and composing several operations into one true total transformation are capabilities of increasing difficulty, and the measured structure supports them unequally: global families fail where a local phase-conditioned transporter succeeds. Operability also changes through the network --- it erodes with depth along the readout--transformation pairing, not uniformly. Finally, the measured structure is prescriptive: it specifies a compact interface whose action is fixed rather than fitted, and \S{}7 shows that interface working on unseen shapes, never-fitted parameters and real regions.

\subsection{Contributions}

\setcounter{enumi}{0}
\begin{enumerate}
\item \textbf{Theory: when a transformation law is realisable at all.} A realisability theory in three layers. \textit{General}: an induced action exists exactly when the transformation preserves the encoder's fibres, and existence already yields the homomorphism (\S{}3.2: the universal property of the quotient, with its two consequences). \textit{Single-region}: a linear operator exists iff the kernel is preserved, and its defect obeys the metric-weighted closed form of \S{}1.2 (Theorem 1, Theorem 2). \textit{Piecewise-linear}: on a rectifier encoder the effect decomposes exactly into a within-regime term and a term carried by the pairs of regions the transformation moves between; whether one fixed linear operator can satisfy every visited pair is a decidable joint system (Propositions C, D), and for the fixed-linear class the attainable score has an exact form, with the gate-change share as a measured proxy for its slack (Proposition E). Four further statements cover realisation and composition: a carrier admits a fixed linear action exactly when its retained blocks are closed, in which case the least-squares optimal map is the carrier action seen through the truncation (Propositions F--G); a closed orbit has no exact coordinate-monotone realisation (Theorem 3); chain error obeys a three-regime accumulation law set by the Lipschitz constant of the truth action in the consumer's seminorm (Theorem 4); and a carrier with band-limited orbits and shared rotation planes realises the fixed block-diagonal action with error bounded by the read-in error, the action mismatch and the target tail (Theorem 5).

\item \textbf{Measurement: a protocol that decides whether an intervention realises a transformation} (\S{}3.7). Four quantities grade a candidate operator --- feature-target error, chain target, composition defect, consumer error --- and three premises gate every evaluation before scoring begins: the reference must be defined on the input, the consumer's classes must survive the transformation, and cross-transformation comparisons must match displacement. Every table carries its baseline by construction: copy for transport, the trivial depth rule for attribution, random and shuffled operators for capability.

\item \textbf{Empirical law: the organisation of colour transformations and its erosion.} Orbits are low-order and shared, substantially inherited rather than learned, and operable in a way that erodes with depth along the readout--transformation pairing rather than along the algebra, measured on four pretrained backbones through four depths and two algebras (\S{}6.1), with the price of a chain measured at a fixed total transformation (\S{}5.4).

\item \textbf{Construction: a structure-fixed interface and its measured uses} (\S{}7). Because the measured structure is prescriptive, it fixes an interface whose action is never fitted: $c=E_\theta(z)$ with $A_\Delta=\operatorname{diag}(I,R(\Delta),R(2\Delta))$. It reads hue zero-shot on unseen shapes, generalises to never-fitted parameters, and carries measured uses --- augmentation, composition, detection and microsecond region editing --- each against ground truth and its own baseline (Tables 15 and 16).
\end{enumerate}

\textbf{Table 1} collects the claims of this paper with the evidence that carries each one and its scope.

\begin{table*}[t]
\centering
\small
\caption*{\textbf{Table 1.} Main results at a glance: each row states the claim, the evidence that carries it, and its scope.}
\begin{tabularx}{\textwidth}{@{}>{\raggedright\arraybackslash}X >{\raggedright\arraybackslash}X >{\raggedright\arraybackslash}X@{}}
\toprule
\textbf{Claim} & \textbf{Key evidence} & \textbf{Scope} \\
\midrule
An induced action exists iff fibres are preserved, and existence yields the homomorphism & \S{}3.2 (one-line observation) & general encoders and algebras \\
Realisation is governed by the mixing ratio in the induced metric; only mixing-back is poisonous; retained rank nearly irrelevant & $\kappa$ sweep $4.5\times10^{-16}\to9.34$, closed form within $0.0091$; $\rho(\kappa,\allowbreak{}T_F)=-0.968$ against $\rho(\text{rank})=+0.502$ & linear encoders, stated moments \\
Hue orbits concentrate in harmonics $k\le2$; rotation planes are shared across shapes & $84$--$88\%$ of orbit energy; $\cos\ge0.96$ & measured colour instance \\
The organisation is substantially inherited from input and architecture & untrained $0.976$ vs trained $0.868$ under one protocol & colour instance \\
A carrier is usable iff its retained blocks are closed; then the optimal fixed linear map is the projected carrier action & closure criterion verified on 40 random selections; $K^\star=PAP^{\mathsf T}$ to $9\times10^{-16}$; $10\times$ tighter local transport & colour sites \\
Chain error obeys a three-regime accumulation law & hue chains flat; heat chains degrade monotonically ($0.760\to0.481$); the regime at a site is measured in the consumer's seminorm (Table D.22) & any realised family \\
The gate-change concentration orders the measured transfer, and a change of local rule counts only where the consumer reads it & $\rho=-0.886$ for the consumer-visible share against $+0.49$ (n.s.) for the flip count & piecewise-linear encoders, measured \\
Operability declines with depth along the readout--transformation pairing & eight of eight backbone--family curves; re-cut control $24/24$ negative; ResNet-50 visible-to-random ratio $9.40\to1.14$ for hue against $1.43\to0.68$ for heat & four backbones, real images \\
Learning moves operability only through $\Delta\kappa$ & $3.8\times$ compression and an $80$--$90^\circ$ row-space rotation leave $\Delta T_F\in\{-0.000,\allowbreak{}+0.026\}$ & controlled linear setting \\
A structure-fixed interface reads hue zero-shot and edits regions in microseconds & $3.4^\circ$ synthetic / $8.0^\circ$ real high-concentration; detector AP50 $0.37$ unseen vs $0.81$ seen; $20.9$ $\mu$s vs $38.1$ ms & colour instance \\
The operating domain is knowable in advance & $0.770$ vs $0.532$, gap CI $[0.128,\allowbreak{}0.345]$, AUC $0.660$ & 750 pre-registered real regions \\
\bottomrule
\end{tabularx}
\end{table*}

\section{Related Work}

Four lines of work touch the question this paper asks, and each answers a different part of it. Equivariance measurement scores how much of a transformation a representation carries; symmetry-based architectures build that transformation in; colour and scale-space theory fix what the reference transformations analytically are; and causal abstraction asks when an intervention on a representation corresponds to one on the world. No single line contains the join this paper makes --- existence conditions, realisation form, composition price, and construction --- and \S{}2.5 states the gap line by line.

\subsection{Transformations in existing representations}

The oldest question in this line is whether a representation has \textit{learned} the transformation it is meant to respect. Lenc and Vedaldi [1] measure equivariance and equivalence by fitting a probe-space linear map and scoring how well it commutes with the transformation, which is also the object this paper fits as a baseline operator. The line has since become quantitative: the Lie derivative [2] scores \textit{local} equivariance error along a transformation's flow; Bruintjes et al. [3] ask what governs learned equivariance across layers and find architecture-level dependence; Romero and Lohit [4] learn partial equivariances rather than assuming a group; Brehmer et al. [5] ask whether equivariance matters at scale; and He et al. [6] trace how data augmentation shapes the representations themselves. Equivariant self-supervision [7], [8] identifies or encourages equivariant embeddings rather than measuring them after the fact.

What this line supplies is the empirical object: a frozen representation whose response to a known transformation can be measured, and the practice of grading that response against a re-rendered reference. What it does not contain is an \textit{information condition}. Equivariance is scored, never asked to exist; the score conflates the part of the algebra the representation keeps with the part it merges away; and no score is defined so that its composition over a chain is meaningful. Our fibre condition supplies the missing information condition --- an induced action exists exactly when the encoder's fibres survive the transformation, and existence then delivers the composition law for free --- and \S{}5.4 grades a chain against the truth-level composite rather than treating it as a test of a group axiom. The measurement-side consequence is sharp: because the scores presuppose an operator family, a family can be self-consistent and still fail the transformation, and the standard proxies can rank an unfaithful operator first (\S{}5.5).

\subsection{Symmetry-based and equivariant representations}

A second line builds the symmetry in. Group-equivariant convolutions [9], steerable variants [10], and their colour-specific descendants --- CEConv [11], learning colour-equivariant representations [12], and exact colour equivariance by hypertoroidal covering [13] --- construct layers whose response to the group is exact by design. A parallel line studies what such networks can express and when pointwise activations preserve the symmetry [14], and symmetry-based disentanglement defines group-structured latent factors from the data [15]. Classical representation theory [16] supplies the harmonic-block decomposition this paper uses as its candidate \textit{forms} (\S{}3.6).

These methods assume the transformation and its symmetry group a priori and are strongest exactly where that assumption is right. The complementary situation --- an opaque, already-trained network whose transformation structure is not known --- is the one this paper addresses: we do not build the action, we measure whether one exists, of what form, and at what price, and the measurement then serves as the blueprint for an interface (\S{}7). The two directions meet where a measurement produces a specification: the fixed block-diagonal action of \S{}7.2 is not an architectural prior but a measured premise, and carrier sufficiency (Thm 5) states when such a compiled action is sufficient. There is also a difference in what is claimed: an equivariant architecture asserts a symmetry of the \textit{whole} network, whereas a measurement report can be, and here often is, weaker and more specific --- this action, on this site, for this consumer, with this defect.

\subsection{Colour and perceptual transformations}

Colour and scale have a mature analytic theory of what their transformations are: the axioms of scale-space [17], [18] fix the diffusion family this paper uses as its dissipative instance, and colour-appearance and HSV-family coordinates fix the hue rotation that supplies its group instance. We treat these as the implemented physics of the rendered domain --- a hue rotation and a diffusion semigroup that are exact at the pixel level (Assumption A1) --- and take no position on perceptual uniformity or spectral rendering.

The distinction that matters for us is algebraic rather than perceptual, and it is where this line is silent: the same perceptual space contains a \textit{group} (hue), a \textit{semigroup} (diffusion), and \textit{monoids with degenerate points} (clipped scalings), and the three behave differently under composition. Prior colour-equivariant work targets recognition performance under colour shift; this paper targets the transformation law itself --- its orbit structure, its degeneracies, and what of it remains operable --- and uses the algebra, not the attribute, as the independent variable. That choice is what makes the dissipative family's failure informative rather than embarrassing: the heat semigroup composes exactly at the truth level, and the obstruction appears only when we ask a representation to carry it (\S{}4.7).

\subsection{Causal abstraction and latent linearisation}

The fourth line asks when a high-level intervention corresponds to a low-level one. Causal abstraction [19], [20] gives the interventional-consistency condition; our fibre condition is, formally, that condition specialised to transformation actions. The latent-linearisation line asks when a single direction is a well-defined intervention --- the linear representation hypothesis and its geometry [21], the observation that steering vectors can be steerable but not decodable [22], and the non-identifiability of steering vectors without a stated reference [23], surveyed for representation engineering by Wehner et al. [24] --- with certification-style evaluation proposed for interventional claims [25] and paired-scene benchmarks for compositional faithfulness [26]. Koopman-style approaches [27] ask when a nonlinear system admits an invariant finite-dimensional subspace on which the evolution is linear, and design or learn a lift into it; our band-limited global family is a truncated instance of that lift applied to a \textit{physical} transformation of a frozen vision encoder. What that line does not supply --- and what this paper adds --- is an existence criterion for a \textit{given} representation rather than a designer's lift, together with the transfer from temporal dynamics to non-temporal transformations of images.

Specialising to transformation actions buys three things this line does not contain. First, the \textit{whole algebra} must descend, not single interventions --- and the composition law is then priced quantitatively by the accumulation law, which an intervention-at-a-time condition cannot express. Second, identifiability becomes an object rather than a caveat: the kernel and stabilisers of the representation name exactly which part of the algebra survives, and lumpability of a partition is the classical version of the same condition [28], with the quotient construction itself standard [29], [30]. Third, the question continues past existence to realisation form, family capacity, and construction: we do not assume a linear realisation exists, we measure when it does (\S{}5.2), prove that the natural per-dimension class cannot realise a closed orbit (Thm 3), and build the working object from the measurements (\S{}7).

\subsection{What none of the four lines contains}

The gap is a gap of \textit{question}, not of technique, and it can be said in four sentences. Equivariance measurement answers how much of a transformation a representation carries, but not whether an action exists, which part of the algebra survives, or what a chain of transformations costs. Equivariant architectures answer how to build a prescribed symmetry in, but say nothing about the structure an already-trained opaque network carries. Colour and scale-space theory fix what the reference transformations analytically are, but not whether a learned representation carries them, nor at what price. Causal abstraction and latent linearisation answer when an intervention is well defined, but at the level of single interventions: they contain neither the whole algebra nor its composition price, and they stop before realisation form and construction.

The join is the paper's contribution: a compatibility condition that decides existence, a decomposition that prices realisation, a measured organisation that fixes the form of the operator, and a construction whose action is fixed by that measurement rather than fitted. Appendix G states the nearest collision against each of our claims.

\section{Transformation Laws through Neural Encoding}

This section builds the theory that answers the paper's first question: operable content has explicit structural conditions --- how the encoder merges states decides whether an input action can be defined on features at all; which transformations remain distinguishable, and in what form an action can be realised, require further structure.

\subsection{Actions, encoding, and operable content}

Let $\mathcal{S}$ be a reference state space, M a transformation family acting on it (group, monoid, or semigroup), R: $\mathcal{S}$ $\to$ $\mathcal{X}$ an imaging or rendering map, f: $\mathcal{X}$ $\to$ $\mathbb{R}$\textasciicircum{}d a network site, and $\psi$ = f$\circ$R with image Z\_$\psi$ = $\psi$($\mathcal{S}$). Two spaces each carry an algebra of transformations: the physical space, where M acts with its composition law, and the feature space, where we seek a family of endomorphisms mirroring it.

\textbf{Definition 1 (representation of the transformation algebra on features).} A \textit{representation of M at the site} is a map $\rho$: M $\to$ End(Z\_$\psi$) with $\rho$($\tau$)$\psi$(s) = $\psi$($\tau$s) for all s, $\tau$ --- so that $\rho$($\tau$ $_2$ $\tau$ $_1$) = $\rho$($\tau$ $_2$)$\circ$ $\rho$($\tau$ $_1$) and $\rho$(e) = I follow from the action's laws. (Approximate versions --- $\|$ $\rho$($\tau$)$\psi$(s) $-$ $\psi$($\tau$s)$\|$ $\le$ $\varepsilon$ in a stated norm over a stated state distribution --- are priced under composition in \S{}5.4, Theorem 4.)

The exact induced action $\rho$\_$\tau$ (when it exists) and a fitted operator G\_$\tau$ from some family $\mathcal{P}$ are different objects throughout: the first belongs to the theory, the second to experiment, and a failure of the second is not, by itself, a fact about the first.

\textbf{Definition 2 (representation level vs behavioural level).} The representation above acts on features. One level down, a consumer F: $\mathbb{R}$\textasciicircum{}d $\to$ $\mathbb{R}$\textasciicircum{}m (a probe, a head, a detector) sees only the quotient of Z that it reads; a \textit{behavioural representation} exists when the action descends to that quotient --- F($\rho$($\tau$)z) depending on z only through F(z) and matching F($\psi$($\tau$s)). The representation level asks what the features carry; the behavioural level asks what any given observer of the features can enact. They have independent conditions (\S{}3.2), and confusing them is the root of the proxy failures in \S{}5.5. Figure 1 fixes the two objects the paper compares: the action of a transformation on the world and the operator that stands in for it on features, and panel (b) shows which part of a linear encoder is responsible when the second cannot exist.

\textbf{Nontriviality: which part of the algebra survives.} A constant representation satisfies ($\star$) trivially --- every transformation acts as identity. A representation of a transformation algebra is therefore informative only together with an account of \textit{which part of the algebra it represents faithfully}. Two objects carry this: the \textit{action kernel} --- the transformations acting trivially on all of Z\_$\psi$, the part of the algebra the encoding has merged away --- and the \textit{stabiliser} of a single state, the transformations fixing that state alone. For a group, the represented object is exactly the effective quotient M/ker. Either way, "invariance" and "information loss" become precise: they are the kernel and the stabilisers of the representation. Both are concrete in colour: an orbit built only from the second harmonic cannot distinguish hues $180^\circ$ apart, since $R(2\Delta)$ has kernel $\Delta=180^\circ$; and the grey axis ($s=0$) is stabilised by every hue rotation --- "grey has no hue" is an algebraic fact of the colour solid, and \S{}4.4 measures its operational extent. Two further sites --- six site$\times$parameter points --- in our real-network grid are causally unreachable in the same sense: the action exists in the world but is invisible at the site, whose transportability is exactly $0.000$ against a non-zero input-side displacement (Appendix C.3).

\subsection{Compression--transformation compatibility}

\textbf{Observation (fibre compatibility, and the inherited composition law).} The assignment $\rho_\tau(\psi(s)):=\psi(\tau s)$ is well defined exactly when the fibre condition holds,
\[\adjustbox{max width=\columnwidth}{$\displaystyle (\mathrm F)\qquad \psi(s_1)=\psi(s_2)\ \Longrightarrow\ \psi(\tau s_1)=\psi(\tau s_2),$}\]
and when it holds the composition law comes for free: for $z=\psi(s)$, $\rho_{\tau_2}\rho_{\tau_1}z=\rho_{\tau_2}\psi(\tau_1s)=\psi(\tau_2\tau_1s)=\rho_{\tau_2\tau_1}z$, and $\rho_e\psi(s)=\psi(s)$. This is the universal property of the quotient $Z_\psi$ written for a transformation family. We record it as an observation rather than a theorem because its content is definitional; it is the \textit{consequences} that carry weight, and there are two.

\textbf{Consequence 1 (existence is one condition, not two).} The two requirements of ($\star$) --- well-definedness and the composition law --- are not independent burdens. A representation either exists and is automatically homomorphic, or it does not exist; there is no third case in which an action exists but fails to compose. The behavioural level descends by the same logic, and its condition is the consumer-side analogue of (F):

\[\adjustbox{max width=\columnwidth}{$\displaystyle \begin{aligned}
&(\star\text{-behavioural})\qquad F(z_1)=F(z_2)\\
&\qquad\Longrightarrow\ F(\rho(\tau)z_1)=F(\rho(\tau)z_2)\quad\text{for every }\tau\in M.
\end{aligned}
$}\]

Where it holds, the induced map on the consumer's outputs is well defined and, by the same one-line argument, a homomorphism; where it fails, no action on the consumer's visible state can be faithful however good the feature-level fit is, which is exactly how a full feature-space operation can succeed while the same intervention, read only through particular outputs, fails (\S{}5.5).

\textbf{Consequence 2 (a composition experiment cannot falsify the group axiom).} Because the composition law is inherited rather than assumed, no network-level composition experiment tests it. What such an experiment measures is the defect of a \textit{fitted family} --- which is why \S{}5.4 grades a chain against the truth-level composite and against a direct refit, and not against the group axioms.

\textbf{A direct check of the fibre condition (F).} The condition can be tested without fitting any operator. Two inputs that a frozen consumer maps to the same output must, if the correspondence exists, stay equal after the same physical transformation. Fixing the tolerance at the $q$-quantile of the consumer's base pair-distance distribution --- so that the random-pair preservation rate is $q$ by construction --- the fraction of base-equal pairs that remain equal after the transformation is $0.25$--$0.46$ for hue and $0.58$--$0.82$ for heat at $q=0.01$, across four frozen backbones, far above the chance rate that the tolerance quantile fixes by construction. Existence therefore holds approximately rather than exactly, and more strongly for the dissipative family, which merges states instead of moving them. The full grid, including the per-site rates and the effect sizes, is Table D.23. Figure 2 shows the same check per backbone and per site.

\begin{figure*}[t]
\centering
\includegraphics[width=\textwidth]{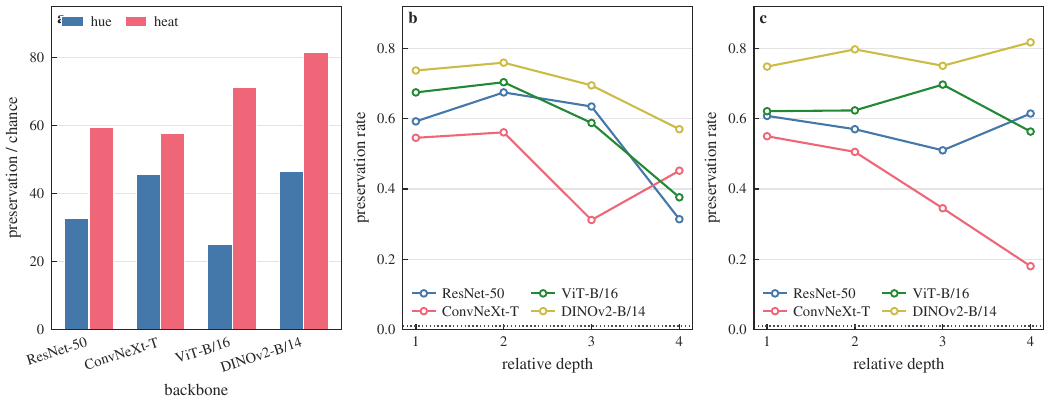}
\caption*{\textbf{Figure 2.} The fibre condition tested directly. (a) Consumer-level preservation of base-equal pairs at tolerance quantile $q=0.01$, expressed as a multiple of the chance rate, for four backbones and both algebras; the chance rate is $q$ by construction. (b) Preservation at each probed site under hue rotation and (c) under heat diffusion, one curve per backbone, against the chance line (dotted); per-site preservation runs from $0.18$ to $0.82$.}
\end{figure*}

\textbf{The organising principle.} Compression matters through its relation to the transformation, not through retained dimension. A transformation that mixes discarded information back into the retained subspace is the only source of principled obstruction. This single principle generates the paper's theory, and \textbf{Table 2} lists what it derives and where each consequence is earned.

\begin{table*}[t]
\centering
\small
\caption*{\textbf{Table 2.} What the organising principle derives, and where each consequence is earned.}
\begin{tabularx}{\textwidth}{@{}>{\raggedright\arraybackslash}X >{\raggedright\arraybackslash}X >{\raggedright\arraybackslash}X >{\raggedright\arraybackslash}X@{}}
\toprule
\textbf{\#} & \textbf{Consequence} & \textbf{Derived how} & \textbf{Earned} \\
\midrule
T1 & A representing action exists iff the transformation preserves the encoder's fibres & mixing-back is the only obstruction to well-definedness & \S{}3.2 (observation) \\
T2 & The defect law $1-T_{RMS}=\kappa_W/\sqrt{1+\kappa_W^2}$ in the induced metric & error = the $C$-block's energy against the demanded displacement, weighted by the encoder & \S{}3.4, Thm 2 (proof + sweep) \\
T3 & Retained rank is nearly irrelevant; mixing is decisive & the law contains $\kappa_W$ alone; rank enters only through where the split is drawn & \S{}3.4 (rank sweep vs $\kappa$ sweep) \\
T4 & Training moves operability only through $\Delta$ $\kappa$ & learning changes the encoder; the law sees only the induced mixing & \S{}6.4 (corollary + trajectories) \\
T5 & The deviation of the linear law at trained sites is a change of local rule & Prop D decomposes the effect; the deviation is carried by the diversity of the realised rules & \S{}3.5, \S{}6.5 (proof + 42 sites) \\
T6 & Depth erosion follows the pairing between the transformation and the readout & rule diversity is created by gate changes (Prop D), and a change of rule counts only where the consumer reads it & \S{}3.5, \S{}6.3 (proof + 2$\times$2 + visible share) \\
T7 & The fixed-linear class attains the exact score $T_F^{\max}$, and the gate-change share is a measured proxy for its slack & Prop E & \S{}3.5 (proof + six-point check) \\
T8 & A carrier admits a fixed action iff its retained blocks are closed, and then the optimal linear map is the projected carrier action & Prop F, Prop G & \S{}4.6 (proof + 40 random selections) \\
\bottomrule
\end{tabularx}
\end{table*}

T1--T4 and T7--T8 are proved or proved-then-verified; T5--T6 are consequences of the piecewise-linear layer of \S{}3.5 read on real networks. One principle reaching existence, pricing, learning, and dynamics --- each consequence checked where it is derived.

\subsection{Two sources of realisation error}

With Z = $\psi$(S), Y\_$\tau$ = $\psi$($\tau$S), squared loss, and a candidate family $\mathcal{P}$:

inf\_\{G$\in$ $\mathcal{P}$\} E$\|$G(Z) $-$ Y\_$\tau$ $\|$ $^2$ = E$\|$Y\_$\tau$ $-$ E[Y\_$\tau$|Z]$\|$ $^2$ + inf\_\{G$\in$ $\mathcal{P}$\} E$\|$G(Z) $-$ E[Y\_$\tau$|Z]$\|$ $^2$.

The first term is irrecoverable information --- what the encoding discarded that the transformation moves; the second is family error --- how far the chosen $\mathcal{P}$ is from expressing the conditional expectation; finite-sample estimation adds a third, standard term. The two failures have different remedies and different meanings, and a minimal example keeps them apart: with z = u and v = (u$^2$ $-$1)/$\surd$2, the nonlinear map (u $\mapsto$ (u, u$^2$ $-$1)/$\surd$2) recovers v exactly from u alone --- no information is missing --- while every linear map fails; the irreducible term is zero and the family term carries everything. A fitted residual therefore measures the sum, and attribution between the terms is an experiment, not an assumption (\S{}3.7).

\textbf{Two normalised scores, and they are not the same number.} The headline score throughout this paper is the \textit{mean-ratio transfer}
\[\adjustbox{max width=\columnwidth}{$\displaystyle T_F\;=\;1-\frac{\mathbb E\lVert F(Gz)-F(\tau z)\rVert}{\mathbb E\lVert F(z)-F(\tau z)\rVert},$}\]
while the quantity a least-squares fit actually optimises is the \textit{RMS-criterion transfer}
\[\adjustbox{max width=\columnwidth}{$\displaystyle T_{RMS}\;=\;1-\sqrt{\frac{\mathbb E\lVert F(Gz)-F(\tau z)\rVert^{2}}{\mathbb E\lVert F(z)-F(\tau z)\rVert^{2}}}.$}\]
Both are ratios in which $1$ matches the real transformed route and $0$ is no better than the no-op; they differ only in whether the average is taken before or after the ratio, and on the measured grids they agree to about one percent ($T_F=0.602$ against $T_{RMS}=0.603$ at the first ResNet-50 site). The distinction matters for exactly one statement: Theorem 2 is a theorem about $T_{RMS}$, because the closed form is derived from the squared-error normal equations. Every table states which of the two it reports, and no comparison in this paper mixes them.

The same two-source structure appears concretely on a rectifier, and there it is measurable. A rectifier network partitions its input space into \textit{activation regions} $R_r$ indexed by the gate pattern, and on each region the site map is affine, $\phi(x)=A_rx+b_r$; the gate is recoverable from the site tensor itself, since $z=\mathrm{ReLU}(a)$ gives $G=\mathbf{1}[z>0]$. Write $u=\phi(x)=A_rx+b_r$ for the source feature and $v=\phi(Tx)=A_sTx+b_s$ for its image under a linear input transformation $T$, and index by \textit{transition cell} $(r,s)$ the inputs that start in region $r$ and land in region $s$. The realisation error of a fixed linear map decomposes into two terms with different meanings,
\[\adjustbox{max width=\columnwidth}{$\displaystyle \begin{aligned}
\min_K\ \mathbb E\|Ku-v\|^2={}&\underbrace{\sum_{r,s}w_{rs}\min_{K_{rs}}\mathbb E\big[\|K_{rs}u-v\|^2\,\big|\,(r,s)\big]}_{\text{within-cell}}\\
&+\underbrace{\sum_{r,s}w_{rs}\,\mathbb E\big[\|(K_{rs}-K^\star)u\|^2\,\big|\,(r,s)\big]}_{\text{sharing cost}},
\end{aligned}
$}\]
with $w_{rs}$ the cell weights, $K_{rs}$ the cell-wise least-squares operators and $K^\star$ the pooled one; the split is the orthogonal decomposition of a pooled regression into within-cell and between-cell parts.

\textit{The first term need not vanish}, and its source is broader than information loss. On a cell $u$ and $v$ are both affine in $x$, which does not make $v$ affine in $u$: if the source rule has already discarded what the target rule needs --- one region covering everything, with $u=x_1$ and $v=x_2$ --- no choice of operator helps, and that is irrecoverability localised to a cell. But a positive within-cell term can also come from the restricted form of the family (a zero-intercept linear fit where the cell relation has an offset, say), so the within-cell term is the error of the \textit{best member of the family on that cell}, not a measure of discarded information. When the source rule does determine the target rule on the cell and the family can express it, the first term vanishes and the whole error is the cost of sharing.

\textit{The second term is what a change of gate pattern can create.} A fixed linear $K$ realises the transformation on the visited cells iff the joint system of \S{}3.5 is consistent; since that system is linear in $K$, consistency is a rank condition, and its least-squares residual is the data-weighted sharing cost when the cell feature moments are matched --- the general statement is the moment-weighted version of the decomposition above. Two qualifications belong to it: the cells must cover the region with non-empty interior, or span it, for a zero finite-sample residual to imply the matrix identity on the region; and the criterion concerns the \textit{visited} cells, which is what a measurement can see.

\subsection{A quantitative linear model}

The linear case is the theory's exact layer, and it is the \textbf{single-region} case of \S{}3.5. Write the compression with retained block u and discarded block v, and the transformation in the same coordinates as [A C; D E]: A moves retained to retained, C mixes discarded into retained, D moves retained to discarded, E acts within the discarded block.

\textbf{Theorem 1 (existence is kernel preservation).} A linear operator realising the transformation on the retained features exists \textbf{iff} the transformation preserves the encoder's kernel (proof in Appendix B).

\textbf{Theorem 2 (defect law).} Under Assumptions B and C the best score is governed by the mixing ratio \textit{in the metric the representation actually induces}: with $A$ and $C$ the retained blocks and $W$ the encoder, the least-squares optimum satisfies
\[\adjustbox{max width=\columnwidth}{$\displaystyle \begin{aligned}
1-T_{RMS}&=\frac{\lVert WC\rVert_F}{\sqrt{\lVert W(A-I_{\mathcal R})\rVert_F^2+\lVert WC\rVert_F^2}}
=\frac{\kappa_W}{\sqrt{1+\kappa_W^2}},\\
\kappa_W&:=\frac{\lVert WC\rVert_F}{\lVert W(A-I_{\mathcal R})\rVert_F}.
\end{aligned}
$}\]
When $W$ is an isometry on the retained subspace --- in particular on whitened features --- this is $\kappa=\lVert C\rVert_F/\lVert A-I\rVert_F$ with the optimum at $A$; in general the optimum is $W$-conjugate. \textit{Proof.} The residual is $WCx_K$ and the displacement is $W(A-I)x_R+WCx_K$; under isotropy $x_R\perp x_K$, so $\mathbb E\lVert WCx_K\rVert^2=\lVert WC\rVert_F^2$ and $\mathbb E\lVert W(A-I)x_R\rVert^2=\lVert W(A-I)\rVert_F^2$, and dividing through gives the displayed form. $\square$

Two readings are counter-intuitive, and the first is metric-independent: \textbf{only mixing-back is poisonous} --- the $D$ block loses information harmlessly while $C$ contradicts it, so \textit{discarding information is not what hurts} --- and \textbf{retained rank barely matters}, since the split enters only through $\kappa_W$. The metric sets the magnitude rather than the mechanism, which is why a law calibrated at one site need not transfer to another with a different $W$. \textbf{Table 3} carries the controlled verification and the scope of the law.

\begin{table*}[t]
\centering
\small
\caption*{\textbf{Table 3.} Controlled verification of Theorems 1--2, and the measured scope of the law. The statistic is named per row: the $\kappa$ sweep is the $T_{RMS}$ of Theorem 2 (the quantity a least-squares fit optimises), the other rows are the mean-ratio $T_F$ of \S{}3.3; the two differ by about a percent where both are recorded.}
\begin{tabularx}{\textwidth}{@{}>{\raggedright\arraybackslash}X >{\raggedright\arraybackslash}X >{\raggedright\arraybackslash}X >{\raggedright\arraybackslash}X@{}}
\toprule
\textbf{check} & \textbf{range} & \textbf{effect on the score} & \textbf{closed-form deviation} \\
\midrule
kernel-preserving $g$ (Thm 1) & retained rank 8 / 48 & $T_F = 0.9999992$ / $0.9999943$, residual $2.6\times10^{-5}$ vs displacement $34.3$ & --- \\
$\kappa$ sweep at fixed rank & $4.5\times10^{-16}\to 9.34$ & $T_{RMS}$ $1.0000\to 0.0041$ (mean-ratio $T_F$ $1.0000\to 0.0041$) & max $0.0091$ for $T_{RMS}$, $0.0102$ for $T_F$ (both at $\kappa\approx1.2$) \\
retained-rank sweep at fixed geometry & rank $8\to48$ & $0.374\to0.665$ (about $2\times$) & --- \\
anisotropy sweep at fixed $W,\allowbreak{}g$ & --- & $0.069\to0.429$ & --- \\
trained linear weights, 32 cells & --- & $\rho(\kappa,\allowbreak{}T_F)=-0.968$ vs $\rho(\text{eff. rank},\allowbreak{}T_F)=+0.502$ & median $0.014$ \\
\bottomrule
\end{tabularx}
\end{table*}

\textbf{Scope of the law.} The closed form is exact under isotropy (Assumption C), and its tightness condition is stated in Proposition B.3: under a general covariance the identity holds only when a cross-covariance term vanishes, and Corollary B.4 shows the score then depends on the input covariance. Real sites are strongly anisotropic, so the law is a statement about the controlled regime rather than a calibrated predictor for real backbones. On the 42 reachable sites of the depth grid, with three estimators of the retained subspace and five ranks and no constant fitted to the score, the prediction is biased high by $+0.40$ to $+0.65$ and orders no better than the naive baselines --- a metric mismatch, since each site brings its own $W$ and its own covariance. What transfers is the mechanism (alignment rather than retained rank) and the diagnostic reading of the deviation; the magnitude is a per-site quantity. Figure 3 shows the law in the controlled setting where it is exact, together with the two candidate explanations it rules out, retained rank and the amount of discarded energy.

\begin{figure*}[t]
\centering
\includegraphics[width=\textwidth]{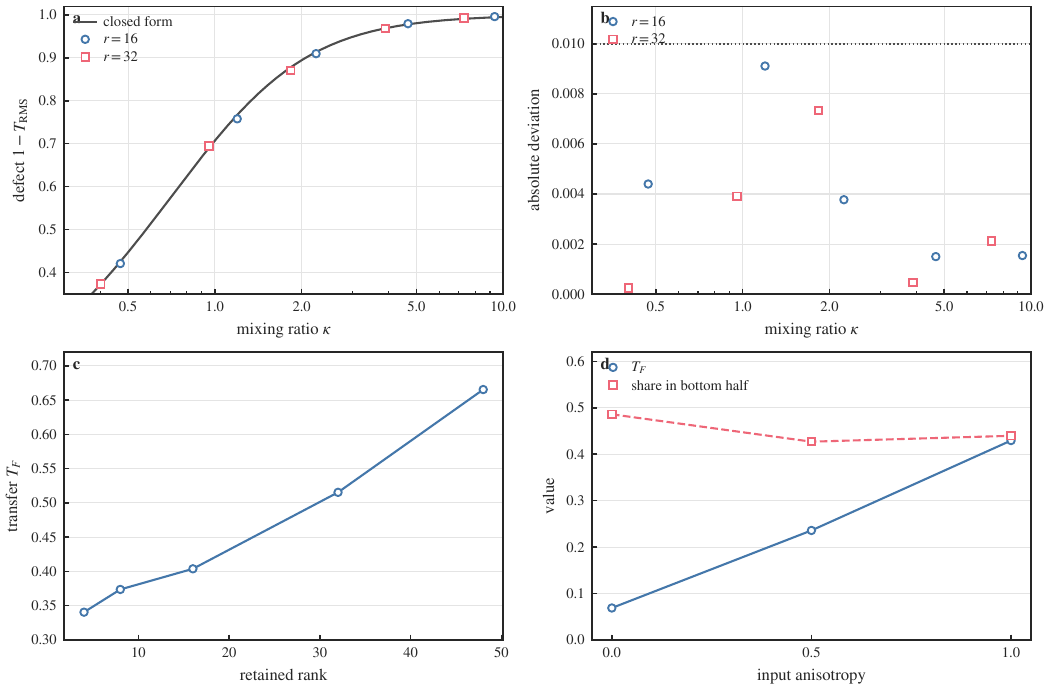}
\caption*{\textbf{Figure 3.} The defect law over its measured range. (a) Defect of the best linear realisation, $1-T_{\mathrm{RMS}}$, against the mixing ratio $\kappa$ for retained ranks 16 (circles) and 32 (squares) on a logarithmic axis; the solid line is the closed form of Theorem 2 in the isometric setting, $\kappa/\sqrt{1+\kappa^2}$. The sweep covers $\kappa=0.40$--$9.34$, and the vertical axis is truncated to the swept range; the two $\kappa\to0$ entries of the same file lie outside it, being the kernel-preserving case with defect below $2\times10^{-6}$. (b) Absolute difference between measurement and closed form over the same sweep against a dotted line at $0.01$; the deviation is largest ($0.0091$) near $\kappa=1.2$ and an order of magnitude smaller at both ends. (c) Transfer against retained rank at fixed geometry: the score moves by about a factor of two while $\kappa$ moves in the opposite direction. (d) Transfer, and the share of the displacement carried by the lowest-variance half of the directions, against input anisotropy: the score rises from $0.069$ to $0.429$ while the share stays between $0.43$ and $0.49$, so the score follows the geometry of the split rather than the discarded energy.}
\end{figure*}

\subsection{The piecewise-linear layer: what a real encoder does}

\S{}3.4 is the single-region case. A rectifier network is not linear on its domain: it partitions the input space into \textbf{activation regions} $R_G$ indexed by the gate pattern $G(x)=\mathbf{1}[a(x)>0]$, and on each region the site map is affine, $\phi(x)=A_Gx+b_G$. Define the \textbf{crossing set} $\mathcal C_\tau=\{x: G(\tau x)\neq G(x)\}$ --- the inputs whose activity pattern the transformation changes.

\textbf{Proposition C (region-transition criterion).} Let the site be affine on each visited region, $\phi=A_rx+b_r$, let $T$ be the input transformation, and let $(r,s)$ range over the realised transitions --- the cells on which $x$ starts in region $r$ and $Tx$ lands in region $s$, each with non-empty interior or spanning its region. Then a fixed linear $\rho$ with $\rho\phi=\phi\circ T$ on the visited cells exists \textbf{if and only if} the joint system
\[\adjustbox{max width=\columnwidth}{$\displaystyle \rho A_r=A_sT,\qquad \rho b_r=b_s\qquad\text{for every realised }(r,s)$}\]
is consistent. \textit{Proof.} On a cell, $\phi(Tx)=A_sTx+b_s$ and $\rho\phi(x)=\rho(A_rx+b_r)$; both sides are affine in $x$, and an identity between affine maps on a set with non-empty interior extends to the maps, so the system is necessary; conversely a solution satisfies it on every cell. $\square$

\textbf{Corollary C.1 (no crossing).} If $T$ leaves the gate pattern of each visited region unchanged, the realised transitions are the diagonal cells and the system reduces to $\rho A_r=A_rT$, $\rho b_r=b_r$ for every visited $r$. \textit{Proof.} With the gate unchanged the realised transitions are the diagonal cells, so the joint system of Proposition C reduces to its diagonal part and nothing else is imposed. $\square$

For a single visited region this is Theorem 1's kernel condition rewritten --- the linear layer is the single-region case.

\textit{Consequence.} \textbf{Even with no gate change at all, the multi-region structure alone can obstruct a fixed linear action}: the visited regions must be simultaneously conjugate to the transformation. This is why enlarging the operator family buys little --- measured, a ridge-plus-$3\times3$-residual family differs from the global linear family by at most $0.087$ at every ResNet-50 site and $0.11$ at every ConvNeXt site, with a median gap of $0.005$ over the 32-site grid; the two separate only where both fail (DINOv2 heat $block8$: $-0.40$ against $-0.99$).

\textbf{Proposition D (the decomposition, and where rule diversity comes from).} With $z=\mathrm{ReLU}(a)$, $a'=a+\Delta a$, $G=\mathbf{1}[a>0]$ and $G'=\mathbf{1}[a'>0]$,
\[\adjustbox{max width=\columnwidth}{$\displaystyle z'-z=G\odot(a'-a)+(G'-G)\odot a',$}\]
and on a unit whose gate changes the second term is $+a'$ (off$\to$on) or $-a'$ (on$\to$off). \textit{Proof.} $z'=G'\odot a'=G\odot a'+(G'-G)\odot a'$ and $z=G\odot a$. $\square$

\textit{Neither direction of implication between gate changes and failure holds.} Take $\phi(x)=(\mathrm{ReLU}\,x,\mathrm{ReLU}(-x))$ with $Tx=-x$: every nonzero input changes one gate, and the fixed operator that swaps the two coordinates realises the transformation \textbf{exactly}, since $\phi(-x)=K\phi(x)$ for all $x$. For the other direction take the scalar site $\phi(x)=\mathrm{ReLU}(x_1)+\mathrm{ReLU}(x_2)$ with $T=\mathrm{diag}(2,3)$, which \textit{preserves every gate pattern}, since positive scalings do not change signs. On the region where the first unit is active, $u$ and $v$ are related by one factor; where the second is active, by the other. Each region is exactly realisable on its own, yet a single scalar $K$ would have to equal both factors at once. No gate change occurs and no shared operator exists. Gate changes are therefore the \textit{source} of rule diversity, not the obstruction; the obstruction is whether one operator satisfies every rule the transformation actually visits, which is the joint system of Proposition C.

\textit{Consequence (which quantity governs).} The within-regime term is linear in $\Delta a$ and is what a linear-class action can reach. The gate-change term has magnitude $|a'|$ on the changed units, so it is not a constant that vanishes with the perturbation, and it carries a dependence on the discrete region index whose reachability by one fixed linear operator is the joint system of Proposition C rather than a matter of counting flips. What governs the achievable score is therefore the \textit{concentration} of the change and its visibility to the consumer, not the number of gate flips. Measured (ResNet-50, three depths $\times$ two algebras): the flip count does not order the score ($\rho=+0.49$, not significant), the gate-change share --- the share of the change carried by $\Delta_{\mathrm{flip}}$ --- does ($\rho=-0.83$), and the consumer-visible share orders it most tightly of all ($\rho=-0.886$; significance levels in Appendix D.7; the measured associations are proxies for the sharing cost of Proposition C, not the cost itself).

\textbf{Proposition E (achievable score of the fixed-linear class).} Let $F$ be the consumer, $\lVert\cdot\rVert_F$ the seminorm its response induces, and $d=z'-z$. The best score attainable by any family whose members are fixed linear maps of $z$ is
\[\adjustbox{max width=\columnwidth}{$\displaystyle T_F^{\max}=1-\frac{\lVert\mathrm{Proj}_F d^\perp\rVert_F}{\lVert\mathrm{Proj}_F d\rVert_F},$}\]
where $d^\perp$ is the residual of the $L^2$ projection of $d$ on the consumer-visible linear span of $z$: the class reproduces the part of the change that is linearly predictable from $z$ and nothing else. \textit{Status.} This exact form is proved for linear consumers; no claim is made here for families outside the fixed-linear class. The proxy is evaluated against the measured transfer at six (site, algebra) points in \S{}6.3.

\textbf{The gate-change share is a proxy, not a bound.} Writing the gate-change term of Proposition D as $\Delta_{\mathrm{flip}}:=(G'-G)\odot a'$, its consumer-visible share is a good proxy for the slack above whenever that term is the part of $d$ not linearly predictable from $z$ --- the generic case, and the reason the share orders the measured transfer scores of \S{}6.3. The proxy is an identity in neither direction: in the counterexample above every gate changes and the slack is zero, and a consumer blind to the change has zero visible share whatever the slack. The measured association is what the proxy licenses; the two-source decomposition of \S{}3.3 is what the failure actually consists of.

This is the layer in which the paper's real-network results live. The deviation of the linear law at trained sites (\S{}6.5) is a change of local rule --- a quantity the framework predicts, not an anomaly --- and the depth trend (\S{}6.1) is the statement that the transformation visits more distinct rules as depth grows, which \S{}6.6 measures.

\subsection{From existence to form}

Existence says nothing about form; representation theory supplies the candidate. If a continuous finite-dimensional linear action of the circle exists on the features, then in an appropriate basis it decomposes into invariant blocks and harmonic rotation blocks --- so the measurement question is sharp: which harmonics carry the energy, are the rotation planes shared, how do amplitude and phase divide the content? The theory does not fix the answers (it does not predict that k $\le$ 2 should dominate), but it tells the measurement what to look for, and \S{}4 answers with numbers. The same premise warns against over-reading coordinates: the natural basis may be elliptical, frequencies need not be orthogonal, and multiple copies of one frequency may mix --- all of which matter for the operator comparisons of \S{}5.

\subsection{Operational measurements}

Four quantities grade any candidate operator: the \textit{feature-target error} (against re-rendered features), the \textit{chain target} (against the truth-level composite), the \textit{composition defect} ($\rho$($\tau$ $_2$)$\rho$($\tau$ $_1$) against $\rho$($\tau$ $_2$ $\tau$ $_1$)), and the \textit{consumer error} (downstream outputs). The no-op (copy) baseline is mandatory --- a transformation that barely moves the features makes copying look like transport --- and so is power: where the transformation does not move what the consumer reads, no operator can be graded and the site is excluded rather than scored. Full protocols: Appendix C. Every results table in this paper carries its baseline column by construction --- copy for transport, the trivial depth rule for attribution, random and shuffled operators for capability --- and the four-quantity rule above is the evaluation criteria box that all capability claims pass through.

Three premises gate every evaluation before scoring begins. \textit{Reference identifiability}: where an attribute is not defined on a state --- a multicoloured region has no global hue --- readouts must fail without error, and no operator can repair a missing reference (\S{}4.4, \S{}7.7). \textit{Consumer-class integrity}: if a transformation tears apart states that a consumer merges, no action on the consumer's visible state can be faithful, however good the feature-level fit (\S{}5.5). \textit{Cross-transformation comparability}: $T_F$ is normalised by the no-op's own error, so a transformation that displaces the consumer further is scored more leniently \textbf{for the same absolute error}. At matched sites the two families' no-op errors differ by a fixed factor (heat is $0.81\times$ hue) while their absolute transport residuals agree to within about a fifth; the apparent group-versus-semigroup gap in $T_F$ ($0.588$ against $0.471$ at layer 2) is therefore a \textbf{denominator effect}, not a statement about how well the transformation is carried. The per-layer decomposition is Table C7. Any comparison of $T_F$ across transformations must report or match the displacement.

All three premises are properties of the encoding and the task, not of any candidate operator --- which is why they are checked first.

The theory now knows when an action can exist and what form to look for. The next section shows what real colour features actually present.

\textbf{Table 4} fixes what is general and what is colour in one place.

\begin{table*}[t]
\centering
\small
\caption*{\textbf{Table 4.} Claim--scope: what is general, what is specific to the colour instance, and where each is earned.}
\begin{tabularx}{\textwidth}{@{}>{\raggedright\arraybackslash}X >{\raggedright\arraybackslash}X >{\raggedright\arraybackslash}X@{}}
\toprule
\textbf{Claim} & \textbf{Scope} & \textbf{Where earned} \\
\midrule
Existence condition and inherited composition law, risk decomposition & General: all encoders and transformation algebras & \S{}3.2--3.3 \\
Kernel-preservation criterion, metric-weighted defect law & \textbf{Single-region (linear) encoders}, stated moment conditions & \S{}3.4 (proofs + controlled verification) \\
Region-transition criterion (Prop C, Cor. C.1), gate-change decomposition (Prop D), exact score of the fixed-linear class (Prop E) & \textbf{Piecewise-linear encoders}; E for the fixed-linear class, with its proxy checked in \S{}6.3 & \S{}3.5 (proofs) \\
Accumulation law (three-regime bound, Thm 4) & Any realised family, truth action with Lipschitz constant L & \S{}5.4 (proof; the regime at each site is measured in the consumer's seminorm, Table D.22) \\
No exact coordinate-monotone realisation of a closed orbit (Thm 3) & Monotone parameter families in a fixed basis & \S{}5.2 (proof + measured best fit) \\
Harmonic concentration, shared scaffold, inheritance & The measured colour instance (and, in specified cells, heat / clipped scalings / multi-attribute and spatial-rotation controls) & \S{}4, \S{}6; Appendix D \\
Carrier closure and optimal linear realisation (Props F--G) & Any carrier and action; a \textbf{design} condition on the retained coordinates & \S{}4.6 (proofs + 40 selections) \\
Carrier sufficiency (Thm 5) and the interface & Any carrier, action and read-in; the measured conditions size its terms, and it is instantiated for colour & \S{}7.3 (proof + measured conditions) \\
\bottomrule
\end{tabularx}
\end{table*}

The linear row is the theory's exact layer and the piecewise-linear row is the layer the measured networks live in; the fourth row is measurement, general as method and colour-specific as fact. Heat flow, clipped scalings, the multi-attribute grid, and spatial rotation appear throughout as the second-algebra and boundary evidence; the full protocol is instantiated for colour and for the composition, pairing and boundary arms of the others.

\section{The Shared Structure of Colour Orbits}

This section reports the paper's central empirical finding: colour variation in the measured frozen features has a shared, low-order organisation --- a candidate foundation for acting across objects.

\subsection{Controlled orbit measurements}

Orbits are rendered from parameters (hue swept densely at fixed (s,v) or per-shape), passed through frozen backbones (four architectures, one of them self-supervised [31], and a YOLO detector), and featurised at the probed sites with DC removed and pose averaged. The protocol table (Appendix C) fixes every definition once; we report results directly. One scope boundary, stated once: the physical transformations of this paper are transformations in rendering parameter space (hue rotation in HSV-like coordinates, heat diffusion on the image grid) --- the synthetic--real gap is measured, not assumed away, and it is visible in the numbers (3.4$^\circ$ synthetic median vs 8.0$^\circ$ on the high-concentration real regions, \S{}7.4).

\subsection{Low-order harmonic organisation}

\textbf{Table 5} collects the organisation measurements that \S{}4.2--\S{}4.5 rely on. Three readings carry the section: the organisation is \textbf{low-order and shared} (harmonics $k\le2$, rotation planes shared across shapes at $\cos\ge0.96$, phase coherence surviving depth while amplitude concentration halves), it is \textbf{substantially inherited} rather than learned (an untrained network preserves the algebraic shape more purely than a trained one, and no affine remixing of the untrained features can synthesise it), and it has \textbf{concrete degeneracies} that bound the operating domain (a $k=2$-only orbit cannot distinguish hues $180^\circ$ apart; the grey axis is stabilised by every hue rotation). Figure 4 shows what that organisation consists of: two harmonics carry almost all of the energy, their planes are shared across shapes, and the low-saturation end of the domain is where the orbit weakens while keeping its plane.

\begin{figure*}[t]
\centering
\includegraphics[width=\textwidth]{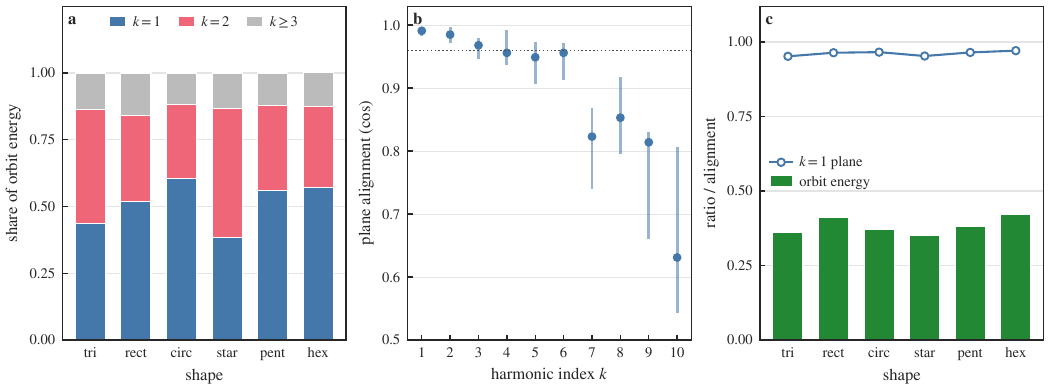}
\caption*{\textbf{Figure 4.} The organisation of hue orbits in the frozen features. (a) Composition of the centred orbit energy by harmonic index for each shape of the dense orbit suite; the first two harmonics carry $0.84$--$0.88$ of the energy in every shape and the tail ($k\ge3$) never exceeds $0.16$. (b) Alignment between each shape's rotation plane and the reference shape's, per harmonic index, shown as the range over the five non-reference shapes with the median marked and a dotted line at $\cos=0.96$; alignment is $0.97$--$1.00$ for $k\le2$ and falls to $0.54$--$0.92$ for $k\ge7$, so the shared planes belong to the low harmonics. (c) The low-saturation limit: bars give the orbit energy at the grey end of the saturation range relative to the mid-saturation ring, and the line gives the alignment of the fundamental plane between the two rings; the orbit loses most of its energy while its harmonic plane survives, which bounds the operating domain.}
\end{figure*}

\begin{table*}[t]
\centering
\small
\caption*{\textbf{Table 5.} The shared structure of colour orbits: spectrum, scaffold, inheritance and degeneracy.}
\begin{tabularx}{\textwidth}{@{}>{\raggedright\arraybackslash}X >{\raggedright\arraybackslash}X@{}}
\toprule
\textbf{quantity} & \textbf{value} \\
\midrule
harmonic share, $k=1{+}k=2$ & $84$--$88\%$ of orbit energy \\
PCA participation ratio & $\approx 3$ \\
spectral centroid & $2.0$--$2.8$ \\
rotation-plane sharing across shapes & principal-angle $\cos\ge 0.96$; band-normalised $\ge 0.95$ \\
phase coherence, trained early $\to$ late layer & $0.971\to 0.906$ \\
phase coherence, input baseline (pixels, RGB) & $0.9997$ \\
amplitude concentration over the same step & halves \\
inheritance: untrained vs trained, same protocol & $0.976$ vs $0.868$ \\
degeneracy: a $k=2$-only orbit cannot distinguish & $\Delta=180^\circ$ \\
grey axis $s=0$ & stabilised by every hue rotation \\
\bottomrule
\end{tabularx}
\end{table*}

\subsection{A shared scaffold with content-dependent envelopes}

The organisation decomposes cleanly into what is shared and what is specific. Rotation planes are shared across shapes while amplitudes carry the shape-specific content, and the division of labour survives depth \textbf{asymmetrically}: the phase that lets one operator serve every shape persists, while the radial structure that would calibrate magnitude erodes. These two measurements are the construction's conditions arriving as numbers --- band-limitation and plane sharing are the two measured conditions the interface's construction uses, quantified here before \S{}7 builds on them (Table 5).

\subsection{Nontriviality and degenerate states}

The abstract kernel and stabiliser of \S{}3.1 have concrete colour content. An orbit carrying only the second harmonic cannot distinguish hues $180^\circ$ apart, and the measured orbits carry both harmonics precisely where hue identity is preserved; the grey and black axes are stabilised by hue rotation, and the discriminative range contracts exactly as saturation approaches the axis. Degeneracy is therefore not noise around the structure but part of the structure: \textbf{the operable domain has a boundary}, and \S{}7.7 prices it for use.

\subsection{Inheritance and reorganisation}

Where does the organisation come from? Under the identical protocol an untrained network preserves the orbit's algebraic shape more purely than a trained one, and no affine remixing of the untrained features can synthesise that shape. Both scores are high --- training does not remove the structure. The conclusion rests on the \textbf{direction of the gap together with the control}: the shape exists at initialisation, exceeds its trained form in algebraic purity, and cannot be created by an affine remap, so its source is the architectural prior and the input statistics, and what training adds is selection and reorganisation rather than creation. Section 6 picks up what training and depth then do to operability. Figure 5 compares eight arms under one protocol and shows that the organisation is present before training, which is why the paper attributes it to the input statistics and the architecture rather than to learning.

\begin{figure*}[t]
\centering
\includegraphics[width=\textwidth]{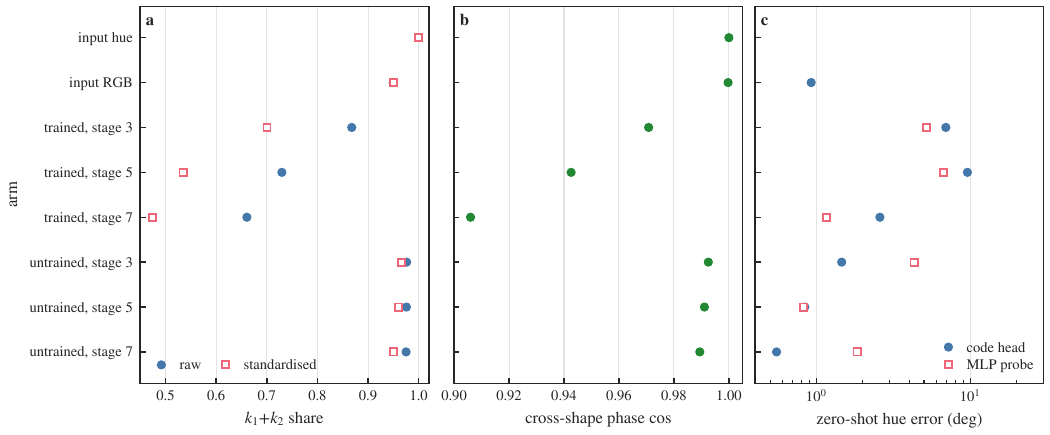}
\caption*{\textbf{Figure 5.} Where the organisation comes from: eight arms under one protocol. (a) Share of orbit energy in the first two harmonics, raw and after standardisation; the untrained sites of the same architectures are at least as concentrated as the trained ones. (b) Cross-shape phase coherence: input pixels are exact by construction, and the untrained sites sit closer to that than the trained ones. (c) Zero-shot hue error on unseen shapes from the code head (circles) and from an MLP probe on the same features (squares), on a logarithmic axis. One seed per arm.}
\end{figure*}

\subsection{When the measured carrier supports a linear action: a design condition}

The measurements above describe the \textit{form} of the orbit. What that form licenses has an exact answer, and it is a condition on the carrier one chooses to keep. Summarise the measured organisation by a carrier map $B$ from $m$ carrier coordinates $c$ (the harmonic blocks) into the feature space, $z=Bc$, and suppose the input transformation acts on the carrier by a known $A_\Delta$ --- for hue, the block rotations $R(k\Delta)$ on the $k$-th pair.

\textbf{Proposition F (closure criterion).} There exists a fixed linear $K$ with $KB=BA_\Delta$ if and only if $A_\Delta\ker B\subseteq\ker B$. \textit{Proof.} $KB=BA_\Delta$ is solvable iff $BA_\Delta$ vanishes on $\ker B$. $\square$

Closure is a property of the \textit{retained blocks}, decidable from the measured organisation alone: $\ker B$ consists of the harmonic directions the carrier does not keep, and the criterion asks whether the rotation carries them back into the retained ones.

\textbf{Proposition G (optimal linear realisation).} Let $\mathbb E[cc^\top]=I$, $z=Bc$ and $z'=BA_\Delta c$. Then $K^\star=BA_\Delta B^\dagger$ is a least-squares minimiser and
\[\adjustbox{max width=\columnwidth}{$\displaystyle \min_K\ \mathbb E\|Kz-z'\|^2=\big\|BA_\Delta P_{\ker B}\big\|_F^2,$}\]
which vanishes exactly when Proposition F holds. For a coloured carrier, $\mathbb E[cc^\top]=\Sigma_c$, the residual is $\|BA_\Delta\Sigma_c^{1/2}P_{\ker(B\Sigma_c^{1/2})}\|_F^2$ with $K^\star=BA_\Delta\Sigma_cB^\top(B\Sigma_cB^\top)^{\dagger}$ and $\Sigma_c^{1/2}$ the \textbf{symmetric} square root. The projector must be taken \textit{after} whitening: correlation lets a retained coordinate predict a discarded one, so the optimal predictor is not the Euclidean projection onto $\ker B$. The two expressions coincide only for $\Sigma_c\propto I$. \textit{Proof.} $K^\star z-z'=BA_\Delta(B^\dagger B-I)c=-BA_\Delta P_{\ker B}c$, because $B^\dagger B$ is the orthogonal projector onto the row space of $B$; taking expectations with $\mathbb E[cc^\top]=I$ gives the Frobenius norm, and the coloured case follows by whitening. $\square$

Three consequences make this a design statement. \textit{Existence:} the same expression that decides whether an action exists also prices its failure. \textit{Error source:} what a fixed linear map on the carrier cannot do is exactly to undo the carrier directions the rotation carries back into view --- the merged harmonics re-entering the retained subspace, the harmonic form of the mixing-back block of \S{}3.4. \textit{Design:} a carrier is usable iff its retained coordinate set is invariant under the action, which for a rotation with one two-dimensional block per frequency means keeping or dropping each conjugate pair \textbf{as a pair}. (The pair rule is stated for a full continuous rotation action written in one block per frequency, with selection along those coordinates; for a single fixed angle, or when frequencies repeat and their blocks may mix, the right object is the invariant subspace rather than the pair.)

Two numerical checks fix the scale, and the full columns are Table D.21. On the three-dimensional code $z(h)=(\cos h,\sin h,\cos 2h)$ --- information-complete for the hue and entirely within the first two harmonics --- the best fixed linear map leaves residual $0.96$ and closure fails, because $\cos2h$ alone is not invariant: the rotation sends it into the missing $\sin2h$. Adding that one coordinate makes the action numerically exact with $A_\Delta=\operatorname{diag}(R(\Delta),R(2\Delta))$, and $K^\star$ is then the block rotation itself. Over random coordinate selections of a six-dimensional carrier, the minimiser equals the projected carrier action $PA_\Delta P^\top$ to machine precision, and the closure verdict coincides with the sign of the residual in every case.

\textbf{What the criterion does and does not explain on measured features.} The same analysis, run on features rather than on a synthetic code, is informative in a different way. Taking the carrier of the known input hue and estimating the synthesis $B$ from the features, the reconstruction tail is a few percent at two sites, the closure defect is numerically zero --- because with a feature dimension far larger than five the synthesis is injective, so the exact carrier model is closed automatically --- and a global linear map fitted on half the hues nevertheless leaves a large relative error on the other half (Table D.21). On these sites the binding constraint is therefore neither closure nor the tail but \textit{cross-content inconsistency}: the carrier-to-feature map is content-dependent, which is the same fact as the shape-specific envelopes of \S{}4.3 and the local-versus-global gap of \S{}5.3. Dropping the $\sin2h$ coordinate raises the tail and leaves the closure defect at zero for an injective synthesis, as the pair rule predicts.

That reading and the content-conditioning result of \S{}5.2 are compatible only if their settings are kept apart, and the paper states both rather than renaming one. Content conditioning did not rescue global forms in 48 matched-capacity contrasts, including an oracle partition by the true displacement direction (\S{}5.2); that experiment conditions a \textit{fitted operator} on content, at matched capacity, on the arms it names. The cross-content statement above is about a \textit{linear map on a measured carrier} whose synthesis is estimated per site, and it names content dependence as what remains after closure and tail are removed. The two differ in model, data, conditioning information and capacity: the content-conditioning contrasts fit a content-conditioned operator family at matched capacity on the arms they name, on synthetic orbits and COCO regions (\S{}5.2), whereas the cross-content statement here concerns a global linear map on a per-site carrier synthesis estimated from dense, controlled orbits. One conclusion follows from them jointly: a content-conditioned family that fails while \textit{using} content information is evidence that the failure is not cured by adding content \textit{capacity}. What the criterion is for, in this paper, is narrower and unambiguous --- it says which coordinate sets a compiled carrier may keep, and what a chosen truncation costs. The compiled interface of \S{}7 keeps all six coordinates, so the condition is satisfied there by construction; what it constrains is any \textit{reduced} carrier, and the dropped-coordinate variant of Table D.21 shows the price of a truncation that is not invariant under the action.

The measured organisation therefore has operational content in a precise sense. Harmonic concentration (\S{}4.2) says a \textit{small} carrier suffices; the shared planes (\S{}4.3) say one $A_\Delta$ serves every shape; and Proposition F says which coordinates must be kept for the action to exist at all. \S{}5 starts from that carrier, and \S{}7 builds the interface on it.

\subsection{Beyond hue: the protocol's reach}

The protocol is not colour-specific, and three further algebras run through it: \textbf{heat as the second full family}, \textbf{clipped scalings as the monoid boundary}, and two perimeter families whose status is labelled rather than claimed. \textbf{Table 6} collects the algebras, what is exact at the truth level, how strongly each family moves a consumer, and how each behaves with depth; the three findings that matter are that heat is a \textit{true} semigroup whose chains degrade while its truth-level law composes exactly, that clipping is a measured failure of composition rather than an assumed one, and that spatial rotation is a \textbf{zero-power} family for an invariant head --- an evaluation-premise failure, not a negative result about representations.

\begin{table*}[t]
\centering
\scriptsize
\caption*{\textbf{Table 6.} The transformation families, their algebras, and what is measured for each.}
\setlength{\tabcolsep}{2pt}
\begin{tabularx}{\textwidth}{@{}>{\raggedright\arraybackslash}X >{\raggedright\arraybackslash}X >{\raggedright\arraybackslash}X >{\raggedright\arraybackslash}X >{\raggedright\arraybackslash}X >{\raggedright\arraybackslash}X@{}}
\toprule
\textbf{family} & \textbf{algebra} & \textbf{truth-level check} & \textbf{consumer response} & \textbf{depth profile} & \textbf{status} \\
\midrule
hue, $\Delta=90^\circ$ & group, $SO(2)$ & exact, re-rendered & reference arm & erodes with depth (\S{}6.1) & in-depth instance \\
heat, $B_t=e^{-tL}$ & semigroup & composes to $1.8\times10^{-7}$ in spectral norm; a truncated Gaussian kernel departs by 22\% & logit effect $0.199/0.707/0.912$ and top-1 flip $0.08/0.48/0.81$ at $\sigma=0.5/1/2$ & the canonical \textbf{spatial} action's residual \textit{rises} $0.389\to0.737\to0.852$ & second full family \\
clipped saturation / value & monoid & non-composition measured, not assumed & headroom $\ge1$: realised $0.174$, error $0.0143$; headroom $<0.33$: clipped fraction $1.0$, realised $0.0$ & --- & boundary \\
multi-attribute grid & mixed (continuous, cyclic, discrete) & --- & per-cell transfer and power & synthetic ViT-B/16 (3 seeds); real DINOv2 / ViT-B/16 (3 seeds each, image-grouped split) & perimeter \\
spatial rotation & group & exact (rot90 round-trip $0.0$) & \textbf{zero} against an invariant head --- an evaluation-premise failure & --- & boundary: needs a rotation-sensitive consumer \\
\bottomrule
\end{tabularx}
\end{table*}

Two further families mark the protocol's perimeter. The \textit{multi-attribute grid} crosses attribute type (continuous, cyclic, discrete) with operator family, with per-cell transfer and power, on synthetic ViT-B/16 (three seeds) and real DINOv2/ViT-B/16 regions (three seeds each, image-grouped split --- boundary status). \textit{Spatial rotation} is the equivariance boundary: an equivariant backbone read by an invariant head has zero power for rotation --- the transformation moves nothing the consumer reads, exactly the evaluation-premise failure \S{}3.7 excludes --- so measuring rotation requires a three-tier consumer design: an ordinary CNN, an augmentation-invariant consumer as a low-power control, and a $C_4$-equivariant consumer with a rotation-sensitive head. Colour is the in-depth instance, heat is the second full family, and these families define the perimeter of what the protocol reaches here.

The measured structure is shared, low-order, and partly inherited. Whether it can be operated depends on how the action is realised --- the next section compares realisations directly.

\section{Realising the Structure: Global Maps, Local Transport, and Composition}

\subsection{Prediction, transport, and composition}

Three tasks, graded separately throughout: \textbf{prediction} --- from a fixed anchor, forecast features at other angles; \textbf{transport} --- apply a specified increment from a starting state never used in fitting; \textbf{composition} --- apply several operations whose product must equal one true total transformation, graded against the ground-truth composite target as well as for internal consistency. Each has its own holdouts (Appendix C). The three tasks are not three benchmarks but the three faces of Definition 1 weighed separately: prediction tests approximation along the orbit, transport tests transfer to states never fitted, and composition tests the homomorphism law $\rho$($\tau$ $_2$ $\tau$ $_1$) = $\rho$($\tau$ $_2$)$\circ$ $\rho$($\tau$ $_1$) itself.

\subsection{Global realisations}

Five natural global families were fitted at matched budgets, and exactly one of them combines fidelity with composition: the band-limited family predicts features of genuinely re-rendered inputs to \textbf{0.17--0.40\% relative MSE}, up to $16.6\times$ tighter than the copy baseline. Content conditioning does not rescue global forms (48 matched-capacity contrasts, including an oracle partition by the true displacement direction); \S{}4.6 states what that negative result does and does not imply about the cross-content reading of the residual error at measured sites. \textbf{Table 7} gives the numbers.

One of the rejected families fails \textbf{by theorem rather than by observation}.

\textbf{Theorem 3 (a closed orbit has no exact coordinate-monotone realisation).} A family $\{G_t\}$ whose members are coordinate-wise monotone in a fixed basis and whose dependence on the family parameter is monotone --- $G_t(z)_i=f_{i,t}(z_i)$ with $(t,u)\mapsto f_{i,t}(u)$ non-decreasing in both arguments --- realises an orbit only if every coordinate of that orbit is monotone in the parameter. Hence a closed non-degenerate orbit admits no exact realisation by such a family, and each coordinate that decreases by $A$ over intervals of length $\ell$ contributes $L^2$ error at least $A^2\ell/4$. (Proved in Appendix B.5.1; the two lemmas are B.7 and B.8.)

The theorem says what the class cannot do; the measurement says what it costs. On the reference orbit the best coordinate-monotone fit carries $L^2$ error $998$ against $924$ for the best fixed linear map (Table 7), and both fail by a wide margin, while the carrier family that keeps the conjugate pair is exact. Segmenting the basis --- letting it follow the field locally, \S{}5.3 --- is the escape the theorem permits and the measurements confirm.

\begin{table*}[t]
\centering
\small
\caption*{\textbf{Table 7.} Global realisations at matched budget, each against the copy floor (\S{}3.7).}
\begin{tabularx}{\textwidth}{@{}>{\raggedright\arraybackslash}X >{\raggedright\arraybackslash}X >{\raggedright\arraybackslash}X >{\raggedright\arraybackslash}X@{}}
\toprule
\textbf{family} & \textbf{fidelity vs copy} & \textbf{composition} & \textbf{verdict} \\
\midrule
symmetric & $0.09$--$0.64\times$ & fails & rejected \\
antisymmetric & $3.07\times$ & fails & rejected \\
band-limited multi-frequency & $0.1$--$0.6\times$ & fails & rejected \\
low-order discrete ($\mathbb Z_2/\mathbb Z_3$) & $\le 0.41$ & fails & rejected \\
monotone per-dimension & no exact realisation of a closed orbit (Thm 3); best fit measured, $L^2=998$ on the reference orbit & fails by theorem and by measurement & rejected \\
\textbf{band-limited, $K=3$--$4$} & \textbf{$0.17$--$0.40\%$ relative MSE; $6.7$--$16.6\times$ over copy} & passes & \textbf{the working global family} \\
content-conditioned (48 matched contrasts) & mean gain $-0.167$, best $+0.017$ & --- & structurally ineffective, not merely weak \\
\bottomrule
\end{tabularx}
\end{table*}

Segmenting --- letting the basis follow the field locally --- is exactly the escape this theorem permits, which is the local transporter of \S{}5.3.

\subsection{Local state-conditioned realisations}

The successful realisation is \textbf{local and phase-conditioned}: the hue circle is divided into $40^\circ$ segments, each carrying a generator estimated from paired endpoint features, with the current phase read from the representation and the segment generator applied --- a nonlinear, state-dependent composite. It fits an order of magnitude tighter than the best global family and closes loops to $0.4\%$ on unseen shapes and starting states. \textbf{Table 8} collects the comparison. Two further gaps --- the median advantage of a local step over a global single direction at a fixed site, and how that gap grows with depth --- are not reported here, because no producing artefact exists for them.

\begin{table*}[t]
\centering
\small
\caption*{\textbf{Table 8.} Local versus global realisation at matched budget.}
\begin{tabularx}{\textwidth}{@{}>{\raggedright\arraybackslash}X >{\raggedright\arraybackslash}X >{\raggedright\arraybackslash}X@{}}
\toprule
\textbf{quantity} & \textbf{value} & \textbf{} \\
\midrule
local vs copy, per site & $1.55$--$15.94\times$ &  \\
global vs copy, same sites & $0.08$--$0.84\times$ &  \\
generator: third-order Fourier fit in phase & $R^2 = 0.87$ & method box, Appendix E \\
orbit bending rate across layers & $1.91\to2.35$ & Appendix D \\
\bottomrule
\end{tabularx}
\end{table*}

\textit{One mechanism hypothesis fits this evidence: the obstacle to global linearisation is shape--colour resonance rather than curvature or non-sharing. Leave-one-shape-out linearisation succeeds $4/4$ at the mid-level site and fails $2/4$ at the deep site, exactly where phase organisation destabilises --- the signature a resonance account predicts. It is stated as a hypothesis, not a result. Figure 6 contrasts the two realisations site by site and prices the composition of a fixed total transformation.}

\begin{figure*}[t]
\centering
\includegraphics[width=\textwidth]{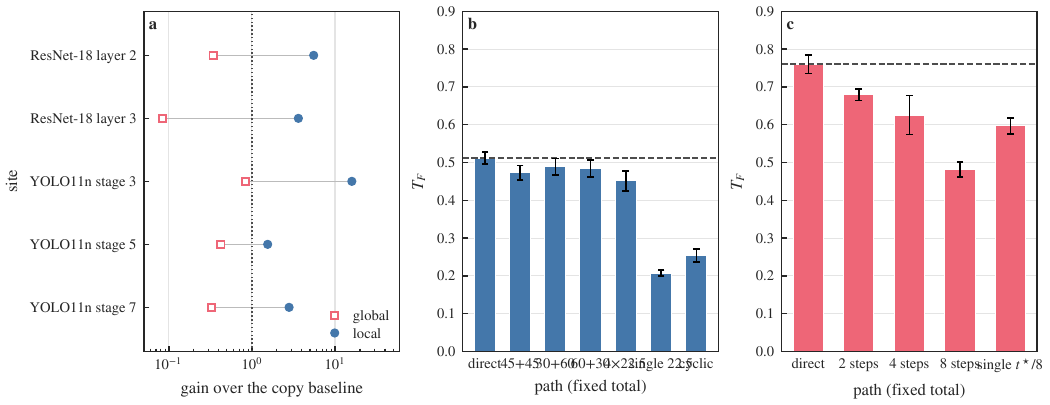}
\caption*{\textbf{Figure 6.} Which realisations work, and what composing costs. (a) Gain over the copy baseline of the best local phase-conditioned transporter (circles) and of the best global family (squares) at five sites, logarithmic axis, with a dotted line at one (no better than copying); the local construction is $1.55$--$15.9\times$ the copy baseline while the global family is $0.08$--$0.84\times$. (b) Transfer at a fixed total hue rotation as a function of how the rotation is split into a chain, with the direct fit as a dashed reference; bars are means over three seeds and whiskers one standard deviation. (c) The same for the heat semigroup, whose chains degrade monotonically with length.}
\end{figure*}

\subsection{Composable transport and its price}

Composition is where the algebra is tested. Holding the \textbf{total} transformation fixed and varying only its decomposition --- two- and four-step paths, an order swap, a cyclic round trip that must return to the identity, and, for heat, chains of two, four and eight steps against a single-step substitute control and a direct fit as the upper reference --- chains come within a few points of a direct fit for both algebras while the single-step substitute does not, and heat degrades monotonically in chain length. Path \textit{consistency} and path \textit{fidelity} come apart: the global linear family agrees most tightly between paths and transports worst, so a family of operators can be perfectly self-consistent and still fail to implement the transformation. \textbf{Table 9} names every path and carries every number, including the saturation caveat (the heat single step scores high because that total sits near saturation) and the cyclic round-trip measure.

\begin{table*}[t]
\centering
\small
\caption*{\textbf{Table 9.} Composition at a fixed total transformation (hue $90^\circ$; heat $t^\star=1.0$), plain-CNN stage 1, three seeds, zero split overlap between fit and evaluation.}
\begin{tabularx}{\textwidth}{@{}>{\raggedright\arraybackslash}X >{\raggedright\arraybackslash}X >{\raggedright\arraybackslash}X >{\raggedright\arraybackslash}X@{}}
\toprule
\textbf{total} & \textbf{path} & \textbf{ridge $T_F$} & \textbf{conv-res $T_F$} \\
\midrule
hue $90^\circ$ & direct fit at $90^\circ$ & 0.285 $\pm$ 0.008 & 0.511 $\pm$ 0.019 \\
hue $90^\circ$ & $45+45$ & 0.237 $\pm$ 0.013 & 0.472 $\pm$ 0.023 \\
hue $90^\circ$ & $30+60$ & 0.249 $\pm$ 0.012 & 0.488 $\pm$ 0.027 \\
hue $90^\circ$ & $60+30$ (order swap) & 0.245 $\pm$ 0.014 & 0.484 $\pm$ 0.028 \\
hue $90^\circ$ & $22.5\times4$ & 0.253 $\pm$ 0.024 & 0.451 $\pm$ 0.032 \\
hue $90^\circ$ & single $22.5^\circ$ ($1/4$ of total) & 0.131 $\pm$ 0.019 & 0.207 $\pm$ 0.010 \\
hue $90^\circ$ & cyclic $+120/-120$ (identity) & 0.067 $\pm$ 0.029 & 0.254 $\pm$ 0.021 \\
heat $t^\star=1.0$ & direct fit at $t^\star$ & 0.610 $\pm$ 0.107 & 0.760 $\pm$ 0.030 \\
heat & two steps & 0.514 $\pm$ 0.132 & 0.679 $\pm$ 0.019 \\
heat & four steps & 0.437 $\pm$ 0.112 & 0.625 $\pm$ 0.063 \\
heat & eight steps & 0.372 $\pm$ 0.114 & 0.481 $\pm$ 0.025 \\
heat & single step ($t^\star/8$) & 0.572 $\pm$ 0.025 & 0.597 $\pm$ 0.025 \\
\bottomrule
\end{tabularx}
\end{table*}

One law organises the whole table, and it is a statement about the \textit{truth} action rather than about any operator.

\textbf{Theorem 4 (accumulation law).} Let the truth action $\tau$ be $L$-Lipschitz in the consumer's seminorm, and let a realised family have single-step error $E_1=\varepsilon$ uniformly on the states visited. Then the $n$-step chain error satisfies
\[\adjustbox{max width=\columnwidth}{$\displaystyle \begin{aligned}
E_n&\le\varepsilon+L\,E_{n-1},\qquad\text{equivalently}\\
E_n&\le\varepsilon\frac{L^n-1}{L-1}\ (L\neq1),\qquad E_n\le n\varepsilon\ (L=1).
\end{aligned}
$}\]
\textit{Proof.} Insert and subtract the truth's intermediate state: $\rho(\tau)^n z-g_{\tau^n}z=\big[\rho(\tau)(\rho(\tau)^{n-1}z)-g_\tau(\rho(\tau)^{n-1}z)\big]+\big[g_\tau(\rho(\tau)^{n-1}z)-g_\tau(g_{\tau^{n-1}}z)\big]$; the first bracket is at most $\varepsilon$ by the uniform single-step bound, the second at most $L\,E_{n-1}$ by Lipschitzness. Iterating gives the closed forms. Appendix B carries the full statement and the uniformity hypotheses. $\square$

The bound separates three regimes by one property of the truth: $L<1$ (contractive) gives a \textbf{bounded} geometric accumulation $\varepsilon/(1-L)$; $L=1$ (isometric) gives \textbf{linear} accumulation $n\varepsilon$; $L>1$ (expansive) gives \textbf{exponential} growth $L^n$. The two tested families differ in how their chains behave, and the accumulation law is the tool that separates the possible causes: a contraction caps the chain error, an isometry accumulates it linearly, an expansion grows it exponentially. Which regime a site occupies must be \textit{measured} in the consumer's seminorm, not inferred from the generator. Two readings follow, and they are worth separating. Heat's monotone degradation in Table 9 is therefore \textbf{not} amplification by the truth action --- a contraction forbids that --- but a growing single-step error as the chain wanders into states where the fitted family is worse; and hue's flat chains, under the loosest of the three bounds, mean that its single-step error is already small rather than that the bound protects it. The regime that would actually blow up, $L>1$, is not instantiated in this paper. In all three cases the composition law itself is automatic; what the accumulation law prices is the realisation error. Estimating $\varepsilon$ and $L$ independently of the measured chain error, and instantiating the expansive class ($L>1$), where the bound predicts exponential growth, are not done here.

\textbf{Metric distortion at real sites.} The one number Theorem 4 needs is how the truth action distorts distances in the consumer's seminorm, and it is measurable. Over pairs of held-out crops, ResNet-50, four depths and both algebras, the median ratio of feature distances after and before the transformation is below one for heat and near one for hue in the \textbf{site} metric --- the qualitative prediction of the generators --- but the same ratio in the \textbf{consumer's} response sits slightly above one for both algebras at every depth (Table D.22). An input-space contraction is not a contraction of what a downstream consumer reads. That is why Theorem 4 is stated with $L$ in the consumer's seminorm and why a site's regime is reported as a distribution of typical stretch rather than read off the generator.

\subsection{Functional consequences of the realised action}

Capability claims are graded on ground-truth targets and downstream behaviour, because the proxies can be gamed. The random orthogonal operator of \S{}1.3 attains the best projection score at $15$ of $42$ sites while its transfer is negative at every one of them and its detector agreement is $F_1=0.000$. Shuffled-pair controls and PCA-aligned subspace controls complete the picture, and the proxies are not worthless: within a family, projection correlates with transfer ($\rho=+0.93$), and the feature-residual ratio correlates with the correct sign --- each instrument is informative inside its scope and insufficient alone. The evaluation rule this paper uses follows directly: feature-target fidelity, composition behaviour, and downstream effect, each against its own baseline, jointly.

Realisation capability varies across layers, and the variation is structured. The next section shows it follows the readout--transformation pairing, and what learning does to it.

\section{How Depth and Learning Reshape Operability}

\subsection{Depth profiles of structure and performance}

What these curves measure, in the theory's terms, is how the piecewise-affine structure of \S{}3.3 shows up layer by layer: a transformation that visits more distinct local rules costs more to realise with one operator, and the unit-level decomposition of \S{}6.6 measures where that cost lands. Operability declines with depth across \textbf{all eight} backbone$\times$family curves measured on real COCO images through four pretrained backbones, and survives the 24/24 re-cut controls and the mild-dissipation ($\sigma=1.0$, single seed) grid. \textbf{Table 10} gives all eight curves. Figure 7 shows the depth profiles and the crossed design together: the curves decline, and the decline follows the pair formed by the transformation and the readout rather than the algebra.

\begin{figure*}[t]
\centering
\includegraphics[width=\textwidth]{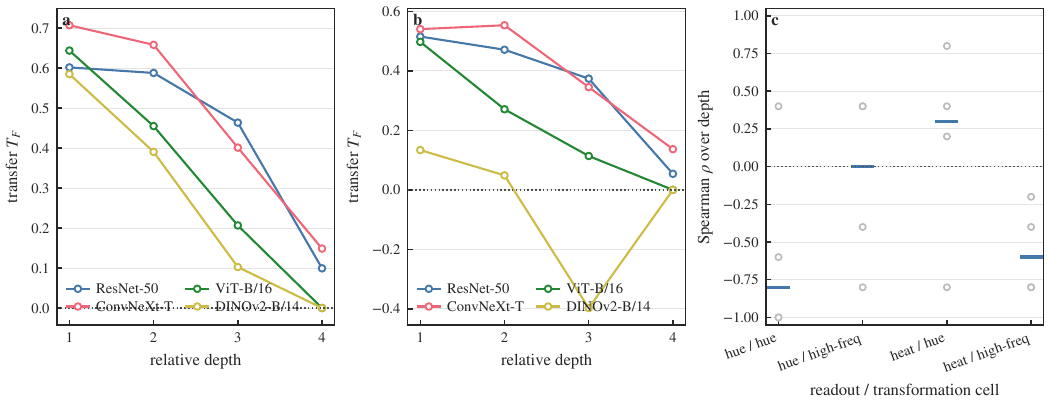}
\caption*{\textbf{Figure 7.} How depth changes operability, and what it changes it to. (a) Transfer of the global linear family against relative depth for four frozen backbones under hue rotation and (b) under heat diffusion; every curve declines, and the final block of the two class-token transformers is causally unreachable, reading zero. (c) The crossed readout $\times$ transformation design summarised by its trend over depth: for each cell, the Spearman correlation between the chance-corrected consumer-visible share and depth, one marker per backbone (open) and the median (bar); the two aligned cells decline and the two off-diagonal cells do not. Seed 0 for all four cells, so the cells are comparable.}
\end{figure*}

\begin{table*}[t]
\centering
\scriptsize
\caption*{\textbf{Table 10.} Downstream transfer of the global linear operator (ridge) against relative depth, real COCO crops, three seeds where noted; $T_F=1$ matches the real route, $0$ is no better than the no-op. The final block of both class-token transformers is causally unreachable and reads exactly $0.000$.}
\setlength{\tabcolsep}{2pt}
\begin{tabularx}{\textwidth}{@{}>{\raggedright\arraybackslash}X >{\raggedright\arraybackslash}X >{\raggedright\arraybackslash}X >{\raggedright\arraybackslash}X >{\raggedright\arraybackslash}X >{\raggedright\arraybackslash}X >{\raggedright\arraybackslash}X@{}}
\toprule
\textbf{backbone} & \textbf{family} & \textbf{d1} & \textbf{d2} & \textbf{d3} & \textbf{d4} & \textbf{trend (ridge)} \\
\midrule
ResNet-50 & hue & 0.602 & 0.588 & 0.464 & 0.099 & $\rho$ = -0.972 \\
ConvNeXt-T & hue & 0.707 & 0.658 & 0.401 & 0.149 & $\rho$ = -0.972 \\
ViT-B/16 & hue & 0.644 & 0.455 & 0.207 & -0.000 & $\rho$ = -0.949 \\
DINOv2-B/14 & hue & 0.585 & 0.391 & 0.103 & -0.000 & $\rho$ = -0.949 \\
ResNet-50 & heat & 0.515 & 0.471 & 0.374 & 0.054 & $\rho$ = -0.972 \\
ConvNeXt-T & heat & 0.540 & 0.553 & 0.345 & 0.137 & $\rho$ = -0.820 \\
ViT-B/16 & heat & 0.497 & 0.271 & 0.114 & 0.000 & $\rho$ = -0.949 \\
DINOv2-B/14 & heat & 0.134 & 0.049 & -0.397 & 0.000 & $\rho$ = -0.949 \\
\bottomrule
\end{tabularx}
\end{table*}

\subsection{The geometry of changing realisability}

A geometric account accompanies the pairing without replacing it. What erodes with depth is the reach of fixed low-parameter families: orbit bending rises to a mean of 729$^\circ$ at the deepest measured layer (up to 760$^\circ$ in a single (shape, saturation, value) cell), closed-loop error worsens (up to 0.24 normalised), and per-shape readouts become switch-like as the continuous rotation field compresses toward discrete sign-like clusters --- three independent measurements agreeing that the coverage a global operator must span grows as the orbit bends, which is also why the local advantage of \S{}5.3 widens with depth. The stronger reading of the same evidence --- a symbolisation transition from continuous fields to discrete clusters --- remains a hypothesis: supported by the geometry, not isolated from the coverage growth that accompanies it.

\subsection{The readout--transformation pairing}

What the depth profiles of \S{}6.1 measure is not the algebra but the \textit{pairing} between the transformation and the readout. The test is a crossed design: two readouts --- a hue-aligned probe and a high-frequency probe --- against the two transformations, on four backbones, with the chance-corrected consumer-visible share as the statistic. If erosion were a property of the algebra, a readout that does not privilege hue would erode under either transformation, and a readout aligned to the dissipative family would be as robust for heat as the hue readout is for hue. Neither holds: each readout erodes only with the transformation it is aligned to, and the off-diagonal cells are flat. \textbf{The erosion is not a numerical-conditioning artefact.} An exactly orthogonal target keeps $T_F=0.9999993$ at depths 1, 4 and 8, at condition number $1.0$, while an ill-conditioned random target at the same depths falls from $0.9993$ to $0.846$ as its condition number rises from $165$ to $2.3\times10^{5}$; what the depth profiles track is the pairing, not the conditioning of the fit.

\begin{table*}[t]
\centering
\small
\caption*{\textbf{Table 11.} The readout $\times$ transformation crossed design: number of backbones (of four) whose trend over depth is a decline, with the median Spearman $\rho$ across sites. ViT-B/16 is a floor case (its chance-corrected visible shares sit at $0.08$--$1.97$) and its flatness carries no information for this judgement; the statistic is the chance-corrected visible share (the visible share divided by the matched-rank random-projector share at the readout's effective rank), not the raw one.}
\begin{tabularx}{\textwidth}{@{}>{\raggedright\arraybackslash}X >{\raggedright\arraybackslash}X >{\raggedright\arraybackslash}X@{}}
\toprule
\textbf{transformation} & \textbf{hue readout} & \textbf{high-frequency readout} \\
\midrule
hue $90^\circ$ (group) & \textbf{3/4 backbones decline}, median $\rho=-0.80$ & 2/4, median $\rho=+0.00$ \\
heat $\sigma=2$ (semigroup) & 1/4, median $\rho=+0.30$ & \textbf{4/4 decline}, median $\rho=-0.60$ \\
\bottomrule
\end{tabularx}
\end{table*}

\textbf{Table 11} reports the design: two readouts (a hue-aligned probe and a high-frequency probe) crossed with the two transformations, four backbones, with the trend of the consumer-visible share over depth as the statistic. Each readout erodes \textbf{only with the transformation it is aligned to} --- the hue readout declines under hue on three of four backbones (median $\rho=-0.80$) and is flat under heat (two of four, median $\rho=+0.00$), while the high-frequency readout declines under heat on \textbf{all four} backbones (median $\rho=-0.60$) and is flat under hue. Depth erosion in this grid therefore follows the readout--transformation pairing, not the group-versus-semigroup distinction. The algebra alone does not predict which representations stay operable; the pair (transformation, consumer) does.

The pairing is not an additional mechanism: it is the piecewise-linear layer of \S{}3.5 read correctly. Proposition D decomposes the effect of a transformation into a within-regime term and a term carried by the units whose gate changes, and a gate change is what makes two cells carry different local rules; whether one operator can serve them both is the joint system of Proposition C. The consumer-visible share of the change is a proxy for that sharing cost, not the cost itself. "A gate change counts only where the consumer reads it" is that statement, and the crossed design tests it. The mechanism numbers are consistent with it and quantitative: on ResNet-50 the consumer-visible gate-change share rises with depth under both algebras --- $0.220\to0.417\to0.627$ for hue and $0.252\to0.302\to0.516$ for heat, layers 2 to 4 --- and across the six (site, algebra) points it is the consumer-visible share of the change that orders the transfer score ($\rho=-0.886$), more tightly than the raw gate-change share ($\rho=-0.83$) and far more tightly than the flip count, which does not order it at all. The ordering, not the algebra, is what the depth trend tracks.

Two consequences follow for how the section, and the paper, should be read. First, the depth trend is \textbf{not} a claim that deep layers represent fewer transformations: it is a claim that the reach of \textit{fixed low-parameter families} shrinks, because the transformation visits more distinct local rules as depth grows, and whether one operator can serve them is the joint system of Proposition C (\S{}3.5). Second, the pragmatic reading of a depth profile is conditional on the consumer: an operator that looks weak under one head can be adequate under another, which is why \S{}5.5 grades capability against the consumer's own baseline and why Table 11's flat cells are evidence rather than noise.

\subsection{Learning in a controlled linear setting}

The $\kappa$ theory generates a learning prediction: with the assumptions of \S{}3.4 held and $\kappa$ fixed, training cannot move the optimal score. Controlled linear networks bear this out --- near-total reorganisation of the retained subspace leaves $\Delta T_F$ at zero without weight decay and raises it only where $\kappa$ moves, and the observed changes track $\Delta\kappa$ ($\rho=-0.833$) rather than the row-space angle ($\rho=-0.146$). Rank and task loss do not describe operability; the induced mixing does. \textbf{Table 12} gives all eight runs with the compression, the loss reduction and the angles.

\begin{table*}[t]
\centering
\scriptsize
\caption*{\textbf{Table 12.} Learning in the controlled linear setting: two geometries, two weight decays, two seeds, byte-identical across invocations. The runs compress the retained subspace by $3.8\times$ and reduce the loss by six orders of magnitude; the row-space angle to initialisation is $80$--$90^\circ$ in every run, the flat ones included.}
\setlength{\tabcolsep}{2pt}
\begin{tabularx}{\textwidth}{@{}>{\raggedright\arraybackslash}X >{\raggedright\arraybackslash}X >{\raggedright\arraybackslash}X >{\raggedright\arraybackslash}X >{\raggedright\arraybackslash}X@{}}
\toprule
\textbf{geometry} & \textbf{weight decay} & \textbf{$\Delta T_F$ (seed 0 / 1)} & \textbf{$\kappa$ before $\to$ after} & \textbf{largest principal angle to init} \\
\midrule
within-block rotation & $0$ & -0.000 / +0.026 & 0.72 $\to$ 0.72 / 0.75 $\to$ 0.75 & 89.7$^\circ$ / 80.3$^\circ$ \\
within-block rotation & $10^{-3}$ & +0.096 / +0.111 & 0.72 $\to$ 0.67 / 0.75 $\to$ 0.68 & 86.2$^\circ$ / 87.8$^\circ$ \\
cross-block shear & $0$ & +0.009 / +0.004 & 0.72 $\to$ 0.71 / 0.70 $\to$ 0.71 & 89.7$^\circ$ / 80.3$^\circ$ \\
cross-block shear & $10^{-3}$ & +0.058 / +0.057 & 0.72 $\to$ 0.58 / 0.70 $\to$ 0.63 & 86.2$^\circ$ / 87.8$^\circ$ \\
\bottomrule
\end{tabularx}
\end{table*}

\subsection{A unified interpretation}

One picture connects the section (Figure 7): the encoder merges and retains, the orbit supplies the realisation form, and the readout decides downstream visibility --- the compatibility principle of \S{}3.2 operating at three places at once. Where the linear law is extrapolated to real nonlinear sites, its closed form deviates systematically. We treat this as a \textit{deviation diagnosis} rather than a failure: the premises fail there independently, and the exact law remains the theory's linear layer, its breakdown pattern is diagnostic rather than predictive --- \S{}3.4 reports that its ordering is no better than the naive baselines --- and the structure--operability links above are measured, not extrapolated.

These regularities say where to extract structure and what form to expect. The next section turns that understanding into design.

\subsection{Where the realisation error goes: a unit-level diagnostic}

The two-source decomposition of \S{}3.3 can be measured on real sites, and the measurement disciplines the mechanism. On a fit split of crops we fit (i) a per-channel affine law for each gate cell, (ii) a pooled per-channel affine law, and (iii) a linear operator in the top-$k$ principal subspace of the features; everything is then evaluated on a \textbf{held-out} split and normalised by the no-op displacement. \textbf{Table 13} gives the held-out realisation error at each site and algebra.

\begin{table*}[t]
\centering
\scriptsize
\caption*{\textbf{Table 13.} Held-out realisation error, ResNet-50, COCO crops, hue $90^\circ$ and heat $\sigma=2$. $R_{\mathrm{diag}}$ is the channel-diagonal operator (equivalently the pooled per-channel law), $S_{\mathrm{cell}}$ the per-cell law, \textit{sharing} their difference, and $R_{\mathrm{full}}(k)$ the linear operator restricted to the top-$k$ principal subspace of the fit features.}
\setlength{\tabcolsep}{2pt}
\begin{tabularx}{\textwidth}{@{}>{\raggedright\arraybackslash}X >{\raggedright\arraybackslash}X >{\raggedright\arraybackslash}X >{\raggedright\arraybackslash}X >{\raggedright\arraybackslash}X >{\raggedright\arraybackslash}X >{\raggedright\arraybackslash}X >{\raggedright\arraybackslash}X@{}}
\toprule
\textbf{site} & \textbf{family} & \textbf{$R_{\mathrm{diag}}$} & \textbf{$S_{\mathrm{cell}}$} & \textbf{sharing} & \textbf{$R_{\mathrm{full}}(8)$} & \textbf{$R_{\mathrm{full}}(16)$} & \textbf{$R_{\mathrm{full}}(32)$} \\
\midrule
layer1 & hue & $0.412$ & $0.412$ & $0.000$ & $0.394$ & $0.213$ & $\mathbf{0.101}$ \\
layer1 & heat & $0.084$ & $0.084$ & $0.000$ & $0.200$ & $0.099$ & $\mathbf{0.042}$ \\
layer2 & hue & $0.347$ & $0.347$ & $0.000$ & $0.921$ & $0.627$ & $0.402$ \\
layer3 & hue & $0.405$ & $0.405$ & $0.000$ & $1.691$ & $1.056$ & $0.846$ \\
layer4 & hue & $0.692$ & $0.650$ & $0.042$ & $1.186$ & $1.114$ & $1.053$ \\
layer4 & heat & $0.750$ & $0.696$ & $0.055$ & $1.365$ & $1.299$ & $1.243$ \\
\bottomrule
\end{tabularx}
\end{table*}

Two statements follow, and neither is the one an in-split measurement suggested. First, \textbf{the cost of sharing one operator across gate cells is negligible out of sample}: the per-cell and pooled channel laws agree to three decimals at the first three sites and differ only at the deepest. A decomposition evaluated inside the fit split reports a large sharing term because small cells are fitted where they are evaluated; held out, the term disappears, and the in-split variant (Table D.20) is retained only as the diagnostic it is. What grows with depth is the \textbf{pooled channel relation itself}. Second, \textbf{the benefit of cross-channel coupling reverses with depth}: a wide operator in the top principal subspace cuts the error substantially at the first layer, while at the last the full operator is worse than the diagonal restriction. Shallow layers let channels help each other; deep layers do not, and at depth the estimate is limited by the sample size available per channel.

That is the mechanism this section can support with one backbone, two algebras and one split: a degrading pooled channel relation and the loss of cross-channel coupling. It is a unit-level diagnostic of \S{}3.3's decomposition, not the full-operator mechanism --- these are per-cell affine and channel-diagonal families fitted at one site, whereas the depth decline of \S{}6.1 is measured for a single global linear operator across layers. A dominant sharing cost is not what the held-out decomposition shows.

\section{From Measured Structure to Operable Interfaces}

\subsection{Design from the measured organisation}

Each measured fact specifies a design choice (Table 5): the low-order spectrum fixes the candidate band K; shared phase fixes a shared action; content-dependent amplitude calls for a content-dependent envelope; the measured degeneracy marks the operating domain; and the complexity of the raw features motivates an optional context branch. K = 2 in what follows is the measured choice, not a theoretical minimum.

\subsection{The interface: learned encoding, fixed action}

The interface is c = E\_$\theta$(z) with action A\_$\Delta$ = diag(I, R($\Delta$), R(2$\Delta$)) --- six dimensions for colour. The action is never fitted --- A\_$\Delta$ is fixed by the measured structure, while the read-in E\_$\theta$ is trained under supervision, so the correspondence is learned but its law is not. The construction's content is that existing features can be read into this structure accurately enough that E\_$\theta$($\psi$($\tau$s)) $\approx$ A\_$\tau$ E\_$\theta$($\psi$(s)) holds off the training manifold; whether a fitted action of equal capacity does better is measured below (Table F.10); the licence for the fixed action is that the two conditions the construction uses --- the retained carrier is closed and its planes align with the measured ones --- hold at the site, and both are measured (\S{}4.3, \S{}7.3). Training and anchoring protocols: method box, Appendix E.

The contribution, stated precisely: the rotation-block form comes from the known structure of the circle group, and the \textit{measurement} decides which blocks are useful, how many are needed, and where the interface is valid; training maps existing features into those coordinates. "The action form is guided by measurement" is what the evidence supports; "the action form was discovered from the network" is not. Two constructions in this paper are also distinct and stay distinct in the wording: the local transporter of \S{}5.3 has its generator and Fourier coefficients fitted and then frozen at inference (Appendix E, Box E1), while the carrier rotation used here is never fitted at all (Box E2).

\textbf{Compared with a learned action.} Whether the fixed action costs anything relative to learning it is now measured rather than left open. Under one read-in, one code supervision and one budget, four arms were trained: the fixed rotation, a learned generator, a learned linear action, and a generic conditioned map carrying more parameters than the fixed action has. The fixed action is exact in the increment it applies and in composition, and its read-out error is not worse than the others'; the learned generator comes closest, since one generator already composes exactly, the learned linear action fits the small increments it saw and fails on an unseen large one, and the generic map is worse than the fixed action on every action metric despite its extra capacity. Learning an action therefore buys nothing here that the measured form does not already give, and it can lose the two properties the construction exists to provide. Table F.10 carries all four arms, three seeds, the parameter counts and the protocol. Figure 8 gives the read-out error, the two never-fitted increments and the composition defect for each arm.

\begin{figure*}[t]
\centering
\includegraphics[width=\textwidth]{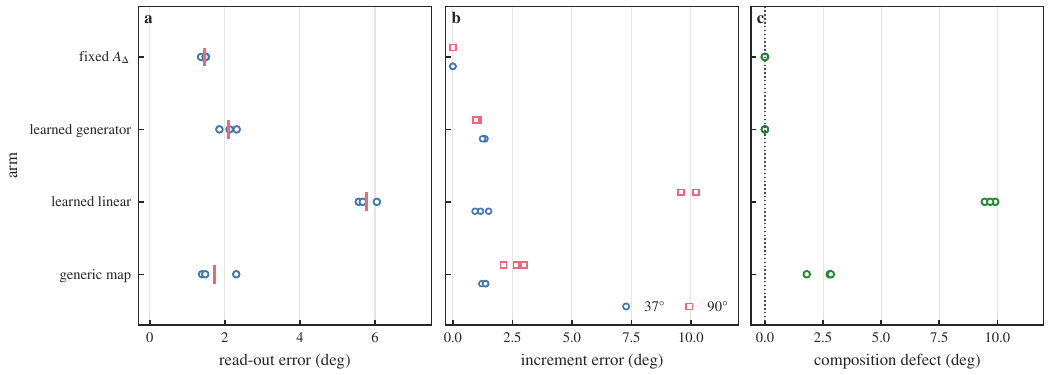}
\caption*{\textbf{Figure 8.} The measured action against learnable actions under one read-in, one supervision and one budget. (a) Zero-shot read-out error. (b) Increment error at the two never-fitted increments, $37^\circ$ (circles) and $90^\circ$ (squares). (c) Defect between a two-step composite and the direct step; the fixed action is exact by construction, and learning the action removes neither the increment error nor the composition defect. Markers are the three seeds.}
\end{figure*}

\subsection{What bounds the interface: the three error terms}

\textbf{Theorem 5 (carrier sufficiency).} Let the carrier be $c$, the synthesis $B$, the target $z'=BAc+w'$ with $\|w'\|_{L^2}\le\delta'$, the read-in $E_\theta$ (any measurable map, in particular the learned MLP), and let the interface apply $A_{\mathrm{int}}$ and decode by $B$. Write $e(z)=E_\theta(z)-c$ for the read-in error. Then the identity
\[\adjustbox{max width=\columnwidth}{$\displaystyle BA_{\mathrm{int}}E_\theta(z)-z'=BA_{\mathrm{int}}\,e(z)+B(A_{\mathrm{int}}-A)c-w'$}\]
holds exactly, and hence, with no cross-term cancellation assumed,
\[\adjustbox{max width=\columnwidth}{$\displaystyle \begin{aligned}
\big\|BA_{\mathrm{int}}E_\theta(z)-z'\big\|_{L^2}\le{}&\|BA_{\mathrm{int}}\|\,\|e\|_{L^2}\\
&+\|B(A_{\mathrm{int}}-A)\|\,\|c\|_{L^2}+\delta'.
\end{aligned}
$}\]
\textit{Proof.} Add and subtract $BA_{\mathrm{int}}c$ and $BAc$; the first term is the read-in error, the second the action mismatch, the third the target's tail. The triangle inequality gives the bound. $\square$

The theorem is an identity, not a conditional bound: it holds for any carrier, action and read-in, and what the measurements supply is the \textit{size} of its three terms rather than a hypothesis of the theorem. Three consequences fix how the bound is to be read. \textit{It applies to the interface as built}: $E_\theta$ need not be linear, and the linear read-in is the special case in which $\|e\|$ is written as an operator norm. \textit{The three terms are read-in error, action mismatch, and target tail} --- only the first is removed by training, and the second vanishes when the interface uses the true carrier action. \textit{Order of the mismatch term:} the bound is first order in $\|A_{\mathrm{int}}-A\|$, and the squared error is therefore second order in the plane misalignment; the measured quantity is the error norm, for which the log-log slope in the tilt angle is $0.99$. If the carrier tail is normalised as an energy fraction, $\mathbb E\|w\|^2\le\varepsilon_{\mathrm{tail}}^2\mathbb E\|z\|^2$, then the measured $12$--$16\%$ energy fraction corresponds to $\varepsilon_{\mathrm{tail}}\approx0.35$--$0.40$, not to $0.12$--$0.16$.

The closure criterion of \S{}4.6 enters separately, and its scope is worth stating precisely: Proposition G bounds what a \textbf{fixed linear map on the measured features} can achieve, and it is that class --- not the interface --- which the measured closure defect limits. The interface escapes the bound exactly because $E_\theta$ is learned and may be nonlinear; what it cannot escape is the \textit{target's} tail and any mismatch between the action it applies and the action the carrier actually has. This is the sense in which the construction is sufficient: the action is fixed by measurement, the read-in is learned, and the two structural conditions --- closure of the retained blocks and alignment of the interface planes --- are checkable in advance on the measured organisation. The three-term inequality is sufficient rather than tight, and the ordering it predicts --- carrier error tracking the tail and the plane misalignment site by site --- is not tested here. \textbf{The framework's \textit{theory} is algebra-general; its \textit{interface} is not.} The hue interface exists because both conditions hold there (\S{}4.3); the dissipative family has no shared decay rate in any basis tested (\S{}8.3, Table 6), which is why it is carried in channel space rather than by the canonical spatial action and why its chains degrade monotonically. A carrier-sufficiency statement whose terms are measurable is a filter rather than a promise: it says which algebras to attempt to compile, and on this instance it returns a negative answer for the semigroup. One boundary of the statement belongs here: the theorem bounds a \textit{feature-space} error, while the interface's read-out figures are measured in degrees, and the link between the two is the decoder's local sensitivity rather than a second theorem. What the decomposition supplies for the capability claims is the attribution: of the three terms, only the read-in error is removed by training, and the few-degree read-out errors of \S{}7.4 are what remains.

\subsection{Generalisation on unseen shapes and parameters}

On shape-held-out grids the interface reads hue zero-shot at a median error of a few degrees --- $3.4^\circ$ by the per-cell-median criterion, $4.4^\circ$ by the per-cell-mean criterion, the two criteria differing in whether two more of the 64 cells pass --- and on real regions it holds wherever the usage rule of \S{}7.7 says it should, at $8.0^\circ$ against $10.0^\circ$ for the non-circular control, far below the $69.5^\circ$ of the unaligned route. The full grid, both criteria, the saturation floor and every route are Table C6. Failures concentrate at low saturation and low value --- the degenerate domain of \S{}4.4 --- and the applicability map they trace is part of the interface's specification, not an afterthought. Figure 9 shows the interface as built: the readout error over the $(s,v)$ domain, the code magnitude that bounds it, the real-region comparison against the pre-registered threshold, and the cost of an edit. \textbf{Figure 10} puts the recovery itself on the same footing, as four separate tests: the operator fitted on a shape it never saw against copying on that shape; the hue phase read out of the code against the hue that was rendered, cell by cell; the same error against how much hue the renderer itself put in the pixel; and the attribute the code reports after a scaling against the attribute the scaling realised.

\begin{figure*}[t]
\centering
\includegraphics[width=\textwidth]{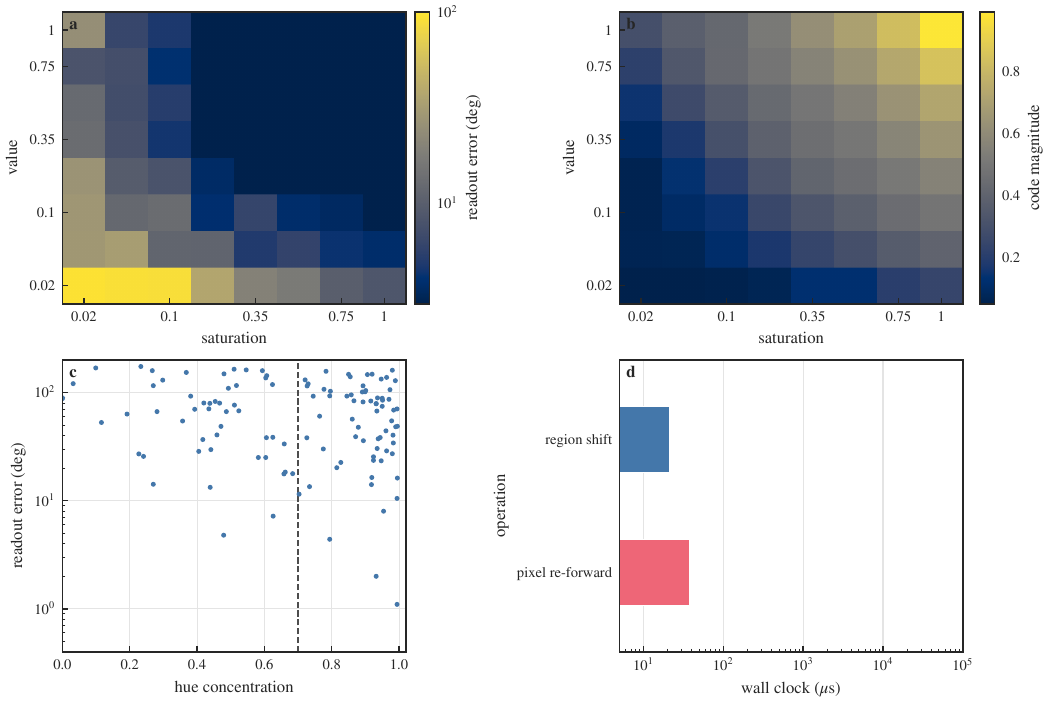}
\caption*{\textbf{Figure 9.} The interface as built, and where it holds. (a) Hue readout error over the (saturation, value) domain on unseen shapes, an $8\times8$ grid on a logarithmic colour scale, and (b) the magnitude of the code's fundamental on the same grid; the readout fails where the code magnitude collapses towards the grey axis. (c) Readout error against the hue-concentration statistic for 120 real COCO regions, with the pre-registered threshold at $0.70$ (dashed); above it the median error is $8.0^\circ$ against $19.6^\circ$ below. (d) Wall clock per operation for a region-level code shift and for a pixel re-colour with a re-forward, logarithmic axis ($20.9$ $\mu$s against $38.1$ ms).}
\end{figure*}

\begin{figure*}[t]
\centering
\includegraphics[width=\textwidth]{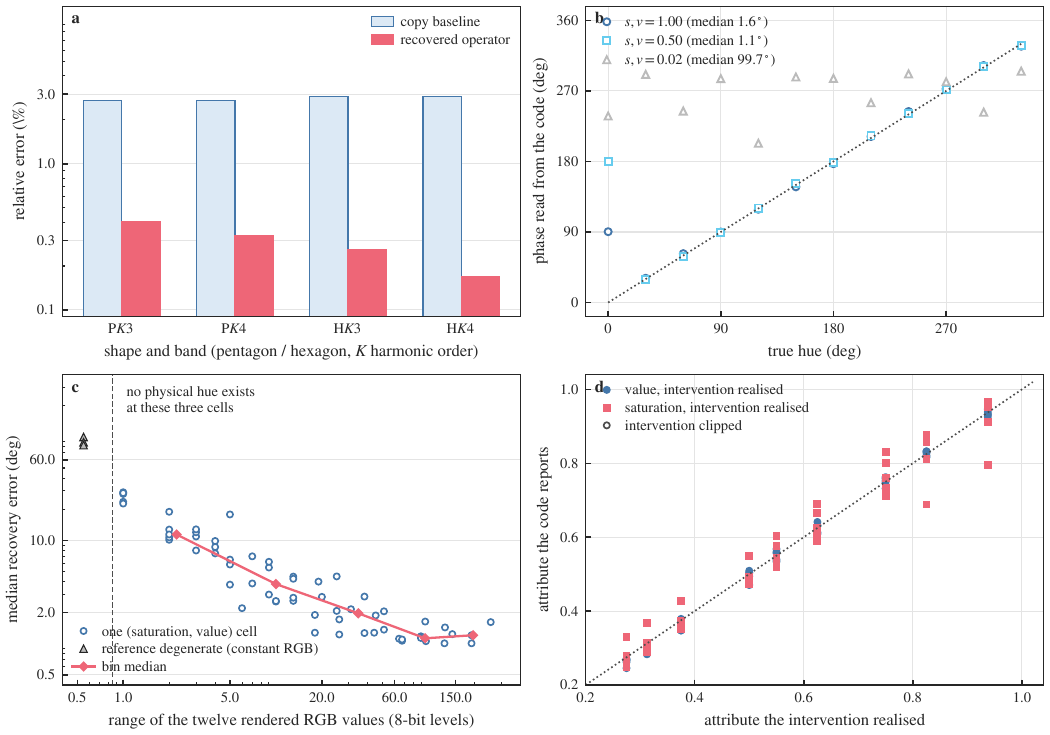}
\caption*{\textbf{Figure 10.} Reading the transformation back out of the features. (a) Relative error of an operator fitted on a shape it was never fitted on, for the pentagon and the hexagon at three and four harmonics, against the copy baseline on the same shape, on a logarithmic axis; the recovered operator is $6.7$--$16.6\times$ closer than copying, at cosine $0.96$--$0.98$. (b) Hue phase read from the code against the hue that was rendered, one point per hue, for three (saturation, value) cells of the dense lattice; the median error is $1.6^\circ$ and $1.1^\circ$ at the two chroma-rich cells, while the third is the degenerate cell whose twelve hues render to a single RGB value and whose phases therefore carry no hue. (c) Median recovery error against the range of the twelve rendered RGB values, logarithmic axes, one point per cell; the error falls with the range the renderer actually provides (Spearman $-0.93$ over the $61$ cells that have a reference) and the three cells with no range at all, the unidentifiable references, carry the largest errors. (d) Attribute the code reports after a multiplicative scaling of the physical attribute against the attribute that scaling realised, one point per trial and both arms, at magnifications $1.1$--$2.0$ with the gamut clip active; the report follows the realised value across the whole range at mean absolute error $0.010$ (value) and $0.031$ (saturation), including the trials whose request saturates the gamut, so the read follows the pixel rather than the request.}
\end{figure*}

\subsection{What the structure buys}

Two comparisons decide what the structure is worth, and they belong to different experiments, so we keep them apart rather than pooling them into one ablation.

\textbf{(a) Usefulness at matched supervision --- the interface.} Under sparse hue supervision (a $40^\circ$ band) the operator-based augmentation reaches $1.3^\circ$ median read-out error against $3.5^\circ$ for input-space interpolation and $25.5^\circ$ for real data only --- the structure substitutes for data the interface never saw, and it even beats the true-midpoint augmentation ceiling. The structured read-out also beats a plain probe on the same features, on the dense grid and on the held-out sheet. The capability table of \S{}7.6 carries both comparisons.

\textbf{(b) Capacity and pairing --- the operator family, not the interface.} The matched-budget capacity controls are run on the plain-CNN stage-1 arm, and we label them as such rather than attaching them to the carrier. At identical architecture, optimiser, epochs and data budget the correct pairing wins and an untrained residual of the same width does not; rank-16 and rank-64 variants land at or below the structured family, so the gain is not width; and shuffling the pairing leaves the projection high while the win rate collapses, which is \S{}5.5's point in its most literal form. \textbf{Table 14} gives the numbers, with the arm named in the table itself.

\begin{table*}[t]
\centering
\small
\caption*{\textbf{Table 14.} Matched-budget capacity and pairing controls, plain-CNN stage-1 arm, hue $\Delta=90^\circ$, three seeds. These rows are an operator-family result; they are not an ablation of the carrier interface, whose matched-supervision comparison is (a) above.}
\begin{tabularx}{\textwidth}{@{}>{\raggedright\arraybackslash}X >{\raggedright\arraybackslash}X >{\raggedright\arraybackslash}X@{}}
\toprule
\textbf{control} & \textbf{projection} & \textbf{paired win rate} \\
\midrule
$1\times3\times3$ residual, correct pairing & $0.816$ & $0.84$ \\
same network, untrained residual & $0.590$ & $0.725$ \\
shuffled pairing & $1.267$ & $0.11$ \\
rank-16 $+$ MLP & $0.716$ & --- \\
rank-16 $+$ $3\times3$ & $0.785$ & --- \\
rank-64 wider family & $0.815$ & --- \\
\bottomrule
\end{tabularx}
\end{table*}

\subsection{Useful operations}

\textbf{Table 15} lists every capability against \textbf{ground truth and its own baseline}, which is the rule the paper applies throughout: a capability claim without a baseline is not a claim. Feature-space augmentation cuts hue read-out error by an order of magnitude against the input-space interpolation baseline; composition reaches parameters never fitted (Table D.11); the detector keeps a third of its seen-shift AP50 at an unseen shift; and region editing costs microseconds against milliseconds because it replaces the re-render it is compared with. \textbf{Figure 11} collects those uses on real images, each against its own baseline.

\begin{figure*}[t]
\centering
\includegraphics[width=\textwidth]{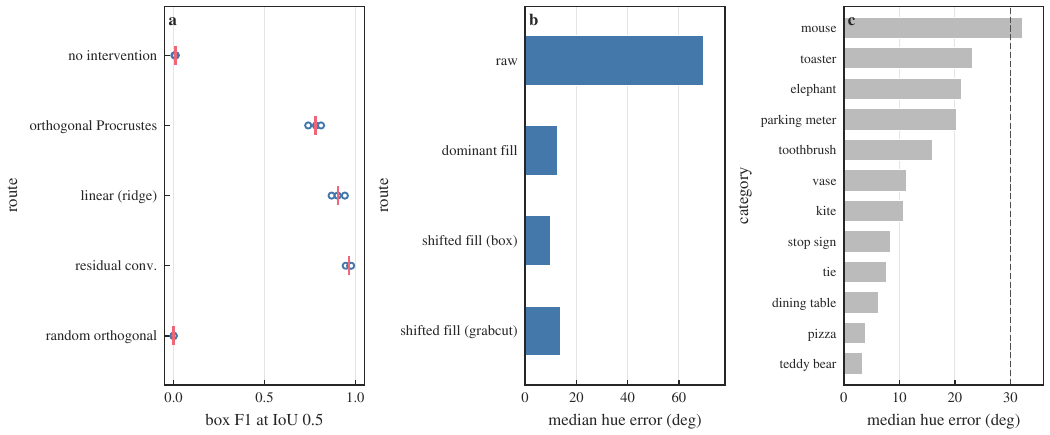}
\caption*{\textbf{Figure 11.} What the interface does on real images. (a) Detector box F1 at IoU $0.5$ for a never-fitted hue shift, one marker per seed (open) and the mean (bar), for five routes; the structured residual family reaches $0.965$, the random orthogonal control $0.000$ and the no-intervention route $0.008$. (b) Median hue error per route on real regions. (c) Median hue error per category on the same regions, sorted, with the $30^\circ$ reference (dashed); eleven of the twelve categories fall below it.}
\end{figure*}

\begin{table*}[t]
\centering
\scriptsize
\caption*{\textbf{Table 15.} Capability claims, each against ground truth and a baseline. Every row belongs to its stated setup; the table reports and the text does not inflate.}
\setlength{\tabcolsep}{2pt}
\begin{tabularx}{\textwidth}{@{}>{\raggedright\arraybackslash}X >{\raggedright\arraybackslash}X >{\raggedright\arraybackslash}X >{\raggedright\arraybackslash}X >{\raggedright\arraybackslash}X@{}}
\toprule
\textbf{capability} & \textbf{metric} & \textbf{baseline} & \textbf{result} & \textbf{setup} \\
\midrule
feature-space augmentation & hue read-out error & input-space interpolation $3.5^\circ$; true-midpoint ceiling $2.2^\circ$ & $25.5^\circ\to 1.3^\circ$ (median) & sparse hue training ($40^\circ$ band), synthetic \\
structured read-out vs plain probe, same features & hue read-out error & plain probe $1.6^\circ$ / $3.7^\circ$ & $0.7^\circ$ dense grid / $0.9^\circ$ held-out sheet & interface, unseen shapes \\
composition to a never-fitted parameter & $T_F$ at the composite & direct fit at that parameter & composed $0.844$ vs direct $0.807$ at $\sigma=1.803$ & plain CNN stage 1, seed 0; never-fitted composite (Table D.11) --- the fixed-total chains of Table 9, three seeds, score $0.451$--$0.488$ against a direct $0.511$ \\
detector at a never-fitted shift & AP50 & seen shift $0.81$ & $0.37$ & YOLO11n, hue-bin classes \\
detector, F1 at IoU $0.5$ & box agreement & null $0.008$ & $0.965$ structured; $0.000$ random & $\Delta=90^\circ$ \\
region editing & wall-clock per operation & pixel re-render $38.1$ ms & $20.9$ $\mu$s ($1825\times$) & measured; the code-shift artefact does not record its device, so the ratio is reported, not a controlled benchmark (Appendix E, Box E4) \\
zero-shot hue read-out & median per-cell error & copy baseline; unaligned route $69.5^\circ$ & $3.4^\circ$ synthetic (median criterion; $4.4^\circ$ mean) / $8.0^\circ$ real high-concentration ($10.0^\circ$ non-circular; $19.6^\circ$ low-concentration) & unseen shapes, 64-cell grid; 120 real regions (69 high-concentration) \\
\bottomrule
\end{tabularx}
\end{table*}

\subsection{Operating domain on real regions}

The operating domain is \textbf{measurable in advance}, which is what makes the interface usable rather than merely demonstrable: on 750 pre-registered COCO regions a hue-concentration threshold separates success from failure on the test half, and the ranking power is moderate rather than sharp (ROC AUC $0.660$), so the rule is published with its calibration curve rather than as a threshold with an implied error bar. On the unaligned 120-region control the concentration statistic alone carries almost no discriminative power and the covariates do the work. \textbf{Table 16} gives every number and route.

\begin{table*}[t]
\centering
\small
\caption*{\textbf{Table 16.} The operating domain on real regions.}
\begin{tabularx}{\textwidth}{@{}>{\raggedright\arraybackslash}X >{\raggedright\arraybackslash}X >{\raggedright\arraybackslash}X@{}}
\toprule
\textbf{sample} & \textbf{statistic} & \textbf{result} \\
\midrule
750 pre-registered COCO regions, test half & concentration threshold, aligned route & $0.770$ vs $0.532$; gap $0.238$, CI $[0.128,\allowbreak{}0.345]$ \\
same & ground-truth-box route & gap $0.188$ \\
same & raw, unaligned route & gap $0.085$ \\
same & ROC AUC & $0.660$ / $0.662$ / $0.579$ (aligned / gtbox / raw) \\
controlled calibration & AUC & $1.0$ \\
120-region control & concentration alone & AUC $0.523$ / $0.480$ \\
same & $+$ mask covariates & $0.754$ / $0.680$ \\
same & $+$ category & $0.651$ / $0.616$ \\
\bottomrule
\end{tabularx}
\end{table*}

The path from the action question through structure to operation is complete. The discussion marks what is general, what is colour, and what comes next.

\section{Discussion}

\textbf{8.1 What transformation structure reveals about representations.} A representation carries more than what can be read from it: it carries what can still be enacted. Encoding merges states, and the mergings that survive transformation define an operable content with explicit conditions --- existence (compatibility), form (the measured organisation), use (the realised interface), and price (what composition costs). Asking what a network's features are becomes, in part, asking what they can still do to themselves. The reframing has a practical edge: an intervention that edits a feature vector is graded here by whether the \textit{remaining} layers produce what the real transformation produces, not by whether the edit looks large in feature space, and the two verdicts come apart routinely --- a random orthogonal operator tops the projection ranking at 15 of 42 sites while its transfer is negative at every one of them (\S{}5.5).

\textbf{8.2 What is general, and what is the colour instance.} The generality of the theory comes from its proofs, and it is worth stating precisely which parts are general. The fibre condition and the inherited composition law of \S{}3.2 hold for any encoder and any transformation family with a composition law; the risk decomposition of \S{}3.3 is an identity for any squared loss and any candidate family; the statement about coordinate-monotone families (Thm 3) is a restricted-class result about closed orbits; the accumulation law (Thm 4) needs only a Lipschitz truth action and a uniform single-step bound; the closure criterion (Props F--G) is stated for any carrier and action, which is why it can be applied to a measured organisation as a design condition; and carrier sufficiency (Thm 5) holds as an identity for any carrier, action and read-in, with the measured conditions sizing its terms, which is why it can return a \textit{negative} answer for an algebra --- as it does for the dissipative family. By contrast the kernel-preservation criterion and the closed form (Thms 1--2) are theorems about the single-region linear layer, and the harmonic concentration, the shared scaffold, the inheritance findings and the interface are colour \textit{measurements}: they are what this instance does, measured where the paper measures, and they are stated with the site and the grid attached. The boundary between the two is the claim--scope table of \S{}3.7, not the reader's inference.

\textbf{8.3 Where the framework stops.} Two limits bound the claims, and both are statements about the object rather than about the evidence for it. The first is the scope of the defect law: it is exact under isotropy, its tightness condition is the vanishing of a cross-covariance term (Proposition B.3), and the score depends on the input covariance (Corollary B.4). Real sites are strongly anisotropic, so the law is not a calibrated predictor there --- on 42 reachable sites the prediction is biased high by $+0.40$ to $+0.65$ and orders no better than the naive baselines --- and what it still supplies is the direction of the effect and a diagnostic use of its own deviation. The second is the dissipative family. The heat semigroup composes to $1.8\times10^{-7}$ at the truth level, yet its per-image decay rate is strongly content-dependent: the rate dispersion is $17.5$ and $8.6$ at layers 2 and 3 against $0.91$ for a synthetic carrier that is shared by construction, and a shared-rate predictor that beats the copy baseline $14.9\times$ on that carrier loses to it on the real sites ($0.83$ and $0.75$). \textbf{Figure 12} collects the out-of-sample tests of this section and of \S{}6.6: the closed form used as a predictor across the reachable sites, the sharing term inside and outside the fit split, the rank-resolved benefit of letting channels correct one another, and the shared-rate extrapolation itself. No basis we tested therefore supplies a common decay rate, the canonical spatial action does not compile, and the family is carried in channel space instead.

\begin{figure*}[t]
\centering
\includegraphics[width=\textwidth]{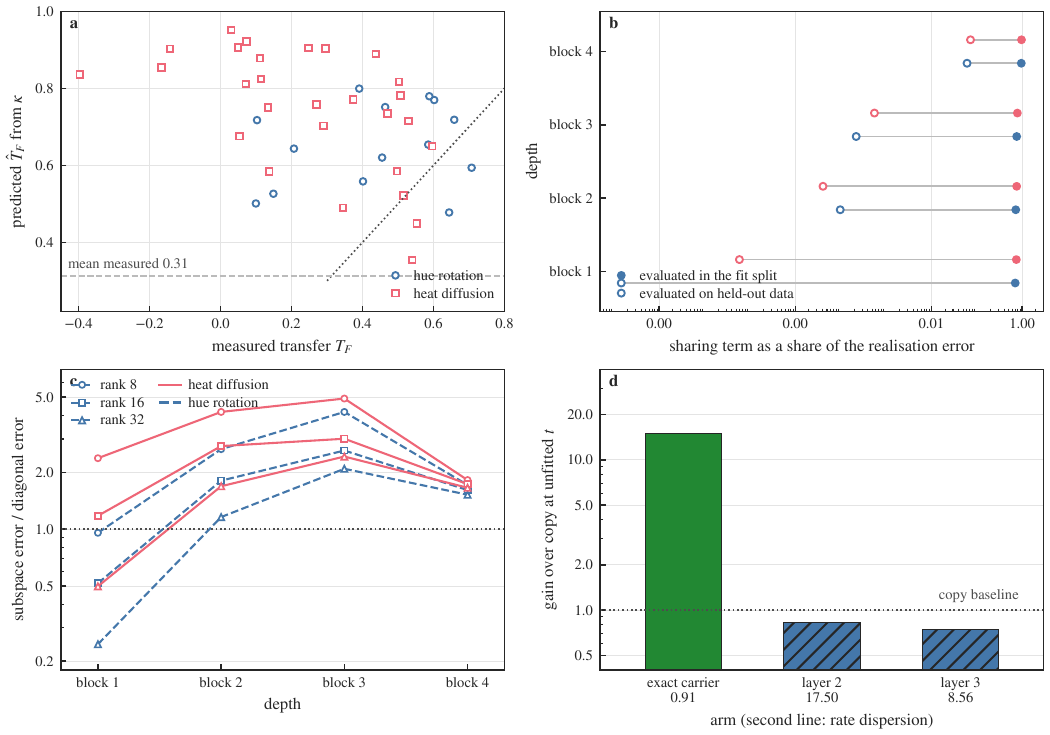}
\caption*{\textbf{Figure 12.} What a fitted law predicts out of sample. (a) Transfer predicted from each site's own mixing ratio by the closed form of Theorem 2 against the transfer measured at that site, one point per reachable site, with the identity line dotted and the measured mean dashed; the prediction is biased high by $0.40$ with mean absolute error $0.43$ against a measured mean of $0.31$, so across real sites the law orders no better than the naive baselines (Spearman $0.48$ between $\kappa$ and transfer, Pearson $-0.44$ between prediction and measurement) --- the negative result of \S{}8.3 shown against the line it would have to lie on. (b) Sharing term as a share of the realisation error, evaluated inside the fit split and on held-out data, per depth and algebra, on a logarithmic axis; the in-split estimate of $0.70$--$0.97$ collapses to $10^{-9}$--$0.073$ out of sample, so a dominant sharing cost is an artefact of fitting and evaluating in the same cells. (c) Error of the rank-$k$ subspace operator relative to the diagonal restriction, against depth, for three ranks and both algebras, with parity dotted; letting channels correct one another pays only at the first block (down to $0.25$ at rank $32$ under hue rotation) and costs at every deeper site, where the diagonal restriction predicts better. (d) Extrapolation to a diffusion time that was never fitted, from the shared mean rate of the fit times, as a gain over the copy baseline; the exact synthetic carrier reaches $14.9\times$ while the two real sites reach $0.83\times$ and $0.75\times$, and the number above each bar is that arm's rate dispersion ($0.91$ for the control against $17.5$ and $8.6$ at the two sites), which is why the shared-rate prediction does not transfer.}
\end{figure*}

\textbf{8.4 The criterion is the instrument, not the object.} The faithfulness score $T_F$ and its companions are properties of a \textit{pair} --- a representation and a consumer. They are not a scalar summary of a representation, and the paper's negative controls show how badly they can mislead when read that way: a random orthogonal operator attains the best projection coefficient at 15 of 42 sites, an unfaithful operator inflates the projection coefficient precisely by being unfaithful, and a consumer that does not read the transformed direction cannot grade any operator at all. The object of study is therefore the transformation's \textit{algebra} and the structure of its representation --- the induced action, its kernel and stabilisers, the four-block split, the gate-change term --- while the score is how we see it. This distinction also explains the paper's most easily misread result: the depth trend is \textit{not} a claim that deep layers represent fewer transformations. What declines with depth is the reach of fixed low-parameter families, and the decline tracks the readout--transformation pairing rather than the algebra (Table 11), which is the same statement as \S{}3.5's: a gate change counts only where the consumer reads it.

\textbf{8.5 Limits and threats to validity.} Six matter. (i) \textit{Residual domain gap.} The reference transformations are exact in rendering-parameter space, and the gap to real images is measured rather than assumed away: $3.4^\circ$ median read-out on synthetic shapes against $8.0^\circ$ on the high-concentration real regions, with the unaligned route at $69.5^\circ$. (ii) \textit{Consumer dependence.} Every capability number belongs to its consumer, which is why \S{}7.5 keeps the interface's matched-supervision comparison separate from the operator-family capacity controls. (iii) \textit{Single-seed cells.} The gate-change study, the semigroup-structure study and the mild-dissipation grid are single-seed, and single-seed cells are never pooled with three-seed cells. (iv) \textit{Estimator-limited fits.} Where the fitted dimension is large against the row count ($d^2/K=428$ at one site), the measured defect over-states the irreducible one, which is why every score carries its rank and budget. (v) \textit{Closure is a design criterion, not a diagnosis.} On measured features the synthesis has full column rank, the exact carrier model is closed automatically, and the residual failure of a global fit is attributable to cross-content inconsistency and tail rather than to non-closure; the criterion's role is to say which coordinate sets a compiled interface may keep. (vi) \textit{Partial algebra coverage.} Colour is carried through the full protocol; heat through composition, pairing and boundary arms; the remaining families are perimeter evidence, and the paper's claims about them are correspondingly narrower.

\textbf{8.6 The next explanatory step.} Three connections are open, and they are ordered by how much they would change the picture. The first is a \textit{nonlinear analogue of the mixing ratio}: where the site is differentiable, the exact intertwining condition implies the local condition $J\psi(\tau s)\,D\tau_s=D\rho_\tau(\psi(s))\,J\psi(s)$ on two Jacobians, so a locally defined $\kappa$, built from the local four-block decomposition, should predict the local defect --- which would turn the crossing decomposition of \S{}3.5 into a quantitative local law rather than a decomposition plus a correlation. The second is a \textit{carrier architecture}: layers that carry transformation laws explicitly, with defect budgets composed across depth the way \S{}5.4 composes them across a chain, would test whether the operability floor of \S{}6.1 is a property of learned representations or of how they are read. The third is the \textit{mechanism-level separation of learning's selection from its reorganisation} (\S{}4.5) --- currently the inheritance result shows training moves the organisation only modestly while the depth profile moves it a great deal, and the two effects are not yet separated by a controlled intervention. A resonance-based readout of shape--colour interaction remains a one-line speculation we do not develop here.

\section{Conclusion}

Reference transformations pass through neural encoders and become operable structure, and that structure can be measured, priced, and built upon. The paper's four contributions are one argument seen from four sides. \textit{Theory:} an induced action exists exactly when the encoder's fibres survive the transformation, existence already delivers the composition law, a linear realisation exists exactly when the kernel is preserved with defect governed by the mixing ratio in the metric the representation induces, on the rectifier layer the failure to realise decomposes into a within-regime term and a term carried by the cells whose rule changes --- the change being the source of rule diversity, not the obstruction --- and a carrier admits a fixed action exactly when its retained blocks are closed, in which case the optimal map is the carrier action seen through the truncation. \textit{Measurement:} four quantities grade an operator, three premises gate the grading, and every capability claim is reported against its own baseline --- a discipline that turns out to be necessary, since the standard proxies can rank an unfaithful operator first. \textit{Empirical law:} colour orbits in frozen features are low-order and cross-shape shared, substantially inherited rather than learned, and their operability erodes with depth along the readout--transformation pairing; on the measured arms a composite fitted only at its parts is not worse than a direct fit (Table D.11), while the three-seed fixed-total chains fall a few points short of a direct refit (Table 9), and the price of a chain is governed by the truth action's Lipschitz constant in the consumer's seminorm, whose regime has to be measured rather than read off the generator. \textit{Construction:} because the measured organisation is prescriptive, it fixes an interface whose action is never fitted and whose premises are checkable --- it reads hue zero-shot at a few degrees on unseen shapes, generalises to never-fitted parameters, survives a real detector at unseen shifts, and edits a region in microseconds against a pixel re-forward measured in milliseconds, with the domain on which it is valid published as a threshold rule rather than left to the reader.

Two things we did not do bound the claims. We did not show the closed form transferring as a calibrated predictor to real backbones --- its magnitude does not, for the measurable reason that real sites violate its isotropy premise --- and we did not build a carrier for the dissipative family, whose truth-level semigroup composes exactly while no basis we tested shares its decay rate. Both are reported as results, and both tell the same story as the positive ones: what survives compression is what does not cross the network's own decision boundaries, and finding out which transformations those are is a measurement, not an assumption. Transformation laws are, in that sense, a concrete object connecting the understanding of neural representations to their design.

\appendix

Appendices A--G collect the material the main text defers: the status of every claim (A), the assumptions and proofs (B), the protocols (C), the full grids (D), the method boxes (E), the capability tables (F) and the positioning (G). Every table states the configuration it was measured on, and every claim carries the scope in which it holds.

The appendices follow in that order.

\textbf{Code and data availability.} The scripts that produce every table and figure of this paper, together with the stored result files they read, are available at \url{https://github.com/hcgt-c/TLNR}. The manuscript states the configuration behind each number; the repository carries the code, the seeds and the raw outputs.

\section{Claim status table}

Every numbered claim and every headline measured claim of the paper in one place, with its status and the evidence that earns it. The labels are fixed: \textbf{Definition} fixes vocabulary; \textbf{Theorem/Proposition} is proved in Appendix B; \textbf{Definitional} is a one-line consequence of a definition, recorded as an observation rather than a theorem; \textbf{Verified} means proved \textit{and} checked in a controlled setting where the answer is known; \textbf{Measured} is an empirical statement with a stated scope, not a theorem; \textbf{Empirical bound} is a measured inequality that is not proved in general; \textbf{Hypothesis} is stated as such in the text and not isolated. No claim in the main text carries a stronger label than it carries here. \textbf{Table A.1}, \textbf{Table A.2} and \textbf{Table A.3} list every claim with the status it carries and the evidence that earns it.

\subsubsection*{A.1 Theory claims}

\begin{table*}[t]
\centering
\scriptsize
\caption*{\textbf{Table A.1.} Claim status: the theory claims, with the evidence that earns each one and the scope in which it holds.}
\setlength{\tabcolsep}{2pt}
\begin{tabularx}{\textwidth}{@{}>{\raggedright\arraybackslash}X >{\raggedright\arraybackslash}X >{\raggedright\arraybackslash}X >{\raggedright\arraybackslash}X >{\raggedright\arraybackslash}X@{}}
\toprule
\textbf{\#} & \textbf{claim} & \textbf{status} & \textbf{evidence} & \textbf{scope} \\
\midrule
Def. 1 & representation of the transformation algebra: $\rho(\tau)\psi(s)=\psi(\tau s)$, with composition and identity inherited & Definition & --- & general encoders and algebras \\
Def. 2 & representation level vs behavioural level & Definition & --- & general \\
Observation & an induced action exists iff fibres are preserved, and existence inherits the composition law & \textbf{Definitional} (one-line, universal property of the quotient) & Appendix B.2 & general \\
Thm. 1 & linear realisation exists iff $g\ker W\subseteq\ker W$ ($C=0$) & \textbf{Verified} & kernel-preserving $g$: $T_F=0.9999992$ (rank 8) to $0.9999943$ (rank 48); residual $2.6\times10^{-5}$ vs displacement $34.3$; Table 3 & linear sites \\
Thm. 2 & $1-T_{RMS}=\kappa_W/\sqrt{1+\kappa_W^2}$ with $\kappa_W=\lVert WC\rVert_F/\lVert W(A-I_{\mathcal R})\rVert_F$; $\kappa$ when $W$ is an isometry on the retained subspace & \textbf{Verified} & $\kappa$ sweep $4.5\times10^{-16}\to9.34$; max deviation $0.0091$; median $0.014$ on 32 trained cells; Table 3 & Assumptions B, C \\
Prop. B.3 & the closed form is tight iff the cross term vanishes & \textbf{Theorem} & Appendix B.3 & Assumption B \\
Cor. B.4 & $T_F$ is not a property of $(W,\allowbreak{}g)$ alone & \textbf{Verified} & anisotropy sweep at fixed $W,\allowbreak{}g$: $T_F$ $0.069\to0.429$ & Assumption B \\
Prop. B.5 & estimation-limited regime: measured $\le$ population $-$ $O(\sqrt{d^2/K})$ & \textbf{Theorem} & two worst cells are those with effective rank $4$ of $32$ & finite samples \\
Prop. C & region-transition criterion: a fixed linear realisation exists iff the joint system over the realised transition cells is consistent (Cor. C.1 for the no-crossing case) & \textbf{Theorem} & Appendix B.4 & piecewise-linear sites \\
Prop. D & $z'-z=G\odot(a'-a)+(G'-G)\odot a'$; gate changes are the source of rule diversity, not the obstruction & \textbf{Theorem} & Appendix B.4; two counterexamples in both directions & rectifier sites \\
Prop. E & $T_F^{\max}=1-\lVert\mathrm{Proj}_F d^\perp\rVert_F/\lVert\mathrm{Proj}_F d\rVert_F$ for the fixed-linear class & \textbf{Theorem} for linear consumers; the gate-change share is an empirical proxy for its slack & evaluated against the measured transfer at six site--algebra points (\S{}6.3) & fixed-linear class \\
Prop. F, G & closure criterion $A\ker B\subseteq\ker B$; optimal linear realisation $K^\star=BAB^\dagger$ with residual $\|BAP_{\ker B}\|_F^2$ (coloured carrier: after whitening) & \textbf{Theorem} + verified & Appendix B.3b; 40 random selections, gap $9\times10^{-16}$ & carrier and action; a \textbf{design} condition \\
Thm. 3 & a closed orbit admits no exact coordinate-monotone realisation; per-coordinate cost bound & \textbf{Theorem} & Appendix B.5.1; best fit measured at $L^2=998$ against $924$ for the best linear map & monotone parameter families in a fixed basis \\
Thm. 4 & accumulation law $E_n\le\varepsilon+LE_{n-1}$; bounded / linear / exponential by $L\lessgtr1$ & \textbf{Theorem}; the regime at a site is measured in the consumer's seminorm (\S{}5.4, Table D.22) & Appendix B.5.2; hue chains flat, heat chains monotone decreasing (Table 9) & any realised family, uniform single-step bound \\
Thm. 5 & carrier sufficiency: three-term $L^2$ bound on read-in error, action mismatch and target tail & \textbf{Theorem} (an identity) & Appendix B.5.3; the measured conditions size its terms: tail energy $12$--$16\%$ ($\varepsilon_{\mathrm{tail}}\approx0.35$--$0.40$), mismatch exponent $0.99$ & any carrier, action and read-in; instantiated for colour \\
\bottomrule
\end{tabularx}
\end{table*}

\subsubsection*{A.2 Measured claims: what the representation carries}

\begin{table*}[t]
\centering
\small
\caption*{\textbf{Table A.2.} Claim status: the measured claims about the organisation of colour orbits and about which realisations work.}
\begin{tabularx}{\textwidth}{@{}>{\raggedright\arraybackslash}X >{\raggedright\arraybackslash}X >{\raggedright\arraybackslash}X >{\raggedright\arraybackslash}X@{}}
\toprule
\textbf{claim} & \textbf{status} & \textbf{key numbers} & \textbf{scope} \\
\midrule
Hue orbits concentrate in low harmonics & Measured & $k\le2$ carries $84$--$88\%$ of orbit energy & measured colour instance \\
Rotation planes are shared across shapes & Measured & $\cos\ge0.96$ between per-shape planes & measured colour instance \\
Amplitude and phase divide the content differently & Measured & shared phase, shape-specific envelope; envelope erodes with depth & measured colour instance \\
The organisation is inherited, not learned & Measured & untrained $0.976$ vs trained $0.868$ under one protocol & colour instance \\
Degenerate states are algebraic, and measured & Measured & grey axis stabilised by every hue rotation; second-harmonic orbits cannot separate $\Delta=180^\circ$ & colour instance \\
Global realisation is possible for one family only & Measured & band-limited $K=3$--$4$: $0.17$--$0.40\%$ relative MSE, $6.7$--$16.6\times$ over copy; other five families rejected & colour sites \\
Local state-conditioned transport succeeds where global fails & Measured & local vs copy $1.55$--$15.94\times$; global vs copy $0.08$--$0.84\times$ at the same sites; loop closure $0.4\%$ & colour sites \\
Chains come within a few points of a direct fit; single-step substitutes do not & Measured & hue $45{+}45$ $0.472$ vs direct $0.511$ (conv-res); single $22.5^\circ$ $0.207$; heat monotone $0.760\to0.481$ & plain-CNN stage 1, three seeds \\
Path consistency is necessary but not sufficient & Measured & most consistent path is not the best transporting one & same \\
Proxies can certify a decoy & Measured & random orthogonal: best projection at $15/42$ sites, negative transfer at $42/42$, detector F1 $0.000$ & cross-architecture grid \\
Within a family, projection tracks transfer & Measured & $\rho=+0.93$ & same \\
\bottomrule
\end{tabularx}
\end{table*}

\subsubsection*{A.3 Measured claims: depth, learning and the interface}

\begin{table*}[t]
\centering
\small
\caption*{\textbf{Table A.3.} Claim status: the measured claims about depth, learning, the two negative results and the interface.}
\begin{tabularx}{\textwidth}{@{}>{\raggedright\arraybackslash}X >{\raggedright\arraybackslash}X >{\raggedright\arraybackslash}X >{\raggedright\arraybackslash}X@{}}
\toprule
\textbf{claim} & \textbf{status} & \textbf{key numbers} & \textbf{scope} \\
\midrule
Operability declines with depth & Measured & $8/8$ curves, family-wise $\rho=-0.82$ to $-0.97$; re-cut $24/24$ negative & four backbones, real images \\
The decline follows the read-out--transformation pairing & Measured & $2\times2$ crossed design; the off-diagonal cells are flat (Table 11) & four backbones \\
The depth ceiling is not low power & Measured & $28/28$ matched pairs lower at depth, sign test $p=7.5\times10^{-9}$; partial $\rho=-0.662$ & 98 points \\
The gate-change concentration, not the flip count, orders the measured transfer & Measured & gate-change share $\rho=-0.83$ ($p=0.042$); consumer-visible share $\rho=-0.886$ ($p=0.019$); flip count $\rho=+0.49$ n.s. & ResNet-50, $n=6$ \\
Learning moves operability only through $\Delta\kappa$ & Measured & $\Delta T_F\in\{-0.000,\allowbreak{}+0.026\}$ flat without weight decay, $+0.096$/$+0.111$ with it; $\rho(\Delta\kappa,\allowbreak{}\Delta T_F)=-0.833$ & controlled linear setting \\
The $\kappa$ law does not transfer as a calibrated predictor & Measured (negative) & bias $+0.40$ to $+0.65$; pooled $\rho\in[-0.48,\allowbreak{}+0.04]$ & 42 real sites \\
The dissipative family needs a shared decay rate and fails & Measured (negative) & truth-level defect $1.8\times10^{-7}$ vs truncated Gaussian $0.2215$ ($\sigma=0.5$ effect $0.1993$) & heat family \\
The interface reads hue zero-shot & Measured & $3.4^\circ$ median on synthetic shapes (median criterion); $8.0^\circ$ on high-concentration real regions, $10.0^\circ$ non-circular control & colour instance \\
The interface transfers to a detector & Measured & AP50 $0.81$ seen vs $0.37$ at an unseen shift & YOLO11n, hue-bin classes \\
Real-region editing is cheap & Measured & $20.9$ $\mu$s region-level shift vs $38.1$ ms pixel re-forward ($1825\times$) & measured hardware \\
The operating domain is knowable in advance & Measured & $0.770$ vs $0.532$ on the pre-registered test half; gap $0.238$, CI $[0.128,\allowbreak{}0.345]$; AUC $0.660$ & 750 real regions \\
The sharing term of the two-source decomposition is an in-split artefact & Measured (negative, held-out) & per-cell and pooled channel laws agree to $0.000$ at the first three sites and differ by $0.042$/$0.055$ at the last; what degrades is the pooled relation ($0.412\to0.347\to0.405\to0.692$ for hue) & ResNet-50, held-out split \\
Input-space metric distortion does not transfer to the consumer & Measured (negative) & site metric $0.80$--$0.99$ (heat) and $0.97$--$1.07$ (hue); consumer response $1.02$--$1.06$ for both & ResNet-50, four depths \\
The fixed action is not worse than learnable actions under identical supervision & Measured & fixed action exact in the increment and in composition; learned generator $1.0$--$1.3^\circ$; learned linear action $9.8^\circ$ at an unseen increment and $9.7^\circ$ composition defect; generic conditioned map worse on every action metric at $1.8\times$ the parameters & synthetic orbits, unseen shapes, three seeds \\
\bottomrule
\end{tabularx}
\end{table*}

\section{Proofs, assumptions, and the status of every statement}

The main text states each result next to the experiment it motivates. This appendix collects the statements, the assumptions under which they hold, and their proofs; where a statement is not proved, its status is said in the same place rather than left to be inferred. The summary of statuses is Table B.4.

\subsubsection*{B.1 Standing assumptions}

\textbf{(A1) Exactness of the reference.} The transformed input is produced by the transformation itself --- a re-render from the transformed parameter, or an exact spectral multiplication --- never by an approximation carried out in a different colour space. Everywhere a reference route $g_\tau z$ appears, it is the image of this exact construction.

\textbf{(A2) Split disjointness.} Every fit and its evaluation use disjoint samples; where a state is said to be \textit{never fitted}, no sample from its neighbourhood was used in fitting.

\textbf{(A3) Power is reported.} Wherever $T_F$ is reported, the displacement $\lVert F(z)-F(g_\tau z)\rVert$ (the no-op error, $T_F$'s denominator) is reported or recorded with it.

\textbf{(A4) Well-definedness of $g_\tau$.} Where $g_\tau$ is claimed to be a function on activations, the fibre condition (F) below holds.

\textbf{Assumption B (linear site, invertible transformation).} On a neighbourhood of the data the site map is linear, $z=Wx$ with $W\in\mathbb R^{d\times D}$ of full row rank $d$, and $\tau$ acts through an invertible linear $g\in GL(D)$. Write $\mathcal R=\operatorname{row}(W)$ (retained), $\mathcal K=\ker(W)=\mathcal R^\perp$ (discarded), and decompose $\mathbb R^D=\mathcal R\oplus\mathcal K$ with
\[\adjustbox{max width=\columnwidth}{$\displaystyle \begin{aligned}
g&=\begin{pmatrix}A&C\\ D&B\end{pmatrix},\qquad A=P_{\mathcal R}gP_{\mathcal R},\ \ C=P_{\mathcal R}gP_{\mathcal K},\\
D&=P_{\mathcal K}gP_{\mathcal R},\ \ B=P_{\mathcal K}gP_{\mathcal K}.
\end{aligned}
$}\]

\textbf{Assumption C (isotropy).} $\mathbb E[x]=0$ and $\mathbb E[xx^\top]=\sigma^2 I_D$. This is the assumption under which the closed form of Theorem 2 is exact; the general case is Proposition B.3.

\subsubsection*{B.2 The general layer (the fibre condition)}

\textbf{Observation (fibre compatibility, and the inherited composition law).} \textit{An induced action of $\tau$ on $Z_\psi$ exists if and only if} (F) \textit{holds, and when it does the induced family $\rho:M\to\operatorname{End}(Z_\psi)$ satisfies $\rho(\tau_2\tau_1)=\rho(\tau_2)\circ\rho(\tau_1)$ and $\rho(e)=I$.} This is the universal property of the quotient $Z_\psi$ written for a transformation family. \textit{Well-definedness:} if (F) holds, $\rho_\tau(\psi(s)):=\psi(\tau s)$ depends only on the class of $s$; conversely, if $\rho_\tau\psi=\psi\circ\tau$ then $\psi(s_1)=\psi(s_2)$ gives $\psi(\tau s_1)=\rho_\tau\psi(s_1)=\rho_\tau\psi(s_2)=\psi(\tau s_2)$. \textit{Inheritance:} for $z=\psi(s)$, $\rho(\tau_2)\rho(\tau_1)z=\rho(\tau_2)\psi(\tau_1s)=\psi(\tau_2\tau_1s)=\rho(\tau_2\tau_1)z$, and $\rho(e)\psi(s)=\psi(es)=\psi(s)$. $\square$

\textit{Two consequences carry the weight; the statement itself is definitional.} First, existence is a single burden rather than two --- satisfying (F) secures well-definedness and the composition law at once, so a representation either exists and composes or does not exist. Second, the composition law therefore \textbf{cannot be falsified by a network-level composition experiment}: what such an experiment measures is the defect of a fitted family, which is why \S{}5.4 grades chains against the truth-level composite and against a direct refit rather than treating them as a test of the group axiom.

\textbf{Approximate form.} If (F) holds only up to $\eta$ --- $\lVert\psi(\tau s_1)-\psi(\tau s_2)\rVert\le\eta$ whenever $\psi(s_1)=\psi(s_2)$ --- then $\rho_\tau$ is well defined on $Z_\psi$ only after quotienting by the equivalence generated by the $\eta$-ball identification, and the induced map carries a defect bounded by $\eta$. All colour and heat measurements of this paper are in the exact regime (A1); the approximate form is what the local transporter of \S{}5.3 accepts deliberately, and the composition price it pays is the one \S{}5.4 measures.

\subsubsection*{B.3 The single-region linear layer (Theorems 1 and 2)}

\textbf{Lemma B.1 (the four blocks, and which one obstructs).} \textit{Write $x=x_R+x_K$ with $x_R=P_{\mathcal R}x$, $x_K=P_{\mathcal K}x$. Then}
\[\adjustbox{max width=\columnwidth}{$\displaystyle z(x)=Wx_R,\qquad z(gx)=WA\,x_R+WC\,x_K .$}\]
\textit{Proof.} $W|_{\mathcal K}=0$ gives the first identity; $gx=Ax_R+Cx_K+Dx_R+Bx_K$ and the last two terms lie in $\mathcal K$, hence are annihilated by $W$. $\square$

The blocks are not interchangeable. $A$ is the part of the action the representation can express; $D$ moves retained information into the discarded subspace and \textit{loses} information without creating a contradiction; $B$ acts inside the discarded subspace and is invisible; $C$ moves discarded information \textbf{into} the retained subspace and is, by Lemma B.1, the only term contributing to $z(gx)$ while being unavailable from $z(x)$. \textbf{$D$ costs information; $C$ costs consistency.}

\textbf{Theorem 1 (existence is kernel preservation).} \textit{Under Assumption B there exists $\rho\in\mathbb R^{d\times d}$ with $z(gx)=\rho\,z(x)$ for all $x$ if and only if $g\mathcal K\subseteq\mathcal K$, equivalently $C=0$, equivalently $\operatorname{row}(Wg)\subseteq\operatorname{row}(W)$.} \textit{Proof.} $z(gx)=Wgx$ and $\rho z(x)=\rho Wx$, so we need $\rho W=Wg$, which has a solution iff $\operatorname{row}(Wg)\subseteq\operatorname{row}(W)$. Since $\operatorname{row}(M)^\perp=\ker M$, that is $\ker W\subseteq\ker(Wg)$, i.e. $Wgx_K=0$ for every $x_K\in\mathcal K$, i.e. $g\mathcal K\subseteq\mathcal K$. By Lemma B.1 the condition is $C=0$. $\square$

\textit{Remark.} The condition constrains a \textbf{direction}, not an amount: $\lVert C\rVert$ says nothing about $\dim\mathcal K$. A rank-1 and a rank-$k$ site are equally editable when $g\mathcal K\subseteq\mathcal K$ and equally broken when $\lVert C\rVert$ is large; compression enters only through where the split is drawn.

\textbf{Theorem 2 (defect law, in the induced metric).} \textit{Under Assumptions B and C let $\rho^\star$ minimise $\mathbb E\lVert\rho z(x)-z(gx)\rVert^2$ over the population, and let $T_{RMS}$ be the RMS-criterion transfer of \S{}3.3. Then}
\[\adjustbox{max width=\columnwidth}{$\displaystyle \begin{aligned}
1-T_{RMS}&=\frac{\lVert WC\rVert_F}{\sqrt{\lVert W(A-I_{\mathcal R})\rVert_F^2+\lVert WC\rVert_F^2}}
=\frac{\kappa_W}{\sqrt{1+\kappa_W^2}},\\
\kappa_W&:=\frac{\lVert WC\rVert_F}{\lVert W(A-I_{\mathcal R})\rVert_F},
\end{aligned}
$}\]
\textit{and when $W$ is an isometry on the retained subspace this reduces to $\kappa=\lVert C\rVert_F/\lVert A-I\rVert_F$ with the optimum at $\rho^\star=A$; otherwise the optimum is $W$-conjugate.} \textit{Proof.} By Lemma B.1 the objective is $\mathbb E\lVert(\rho W-WA)x_R-WCx_K\rVert^2$. On $\mathcal R$ the operator $\rho W-WA$ acts and $x_R$ ranges over a dense subset, so the minimiser restricts to $\rho W|_{\mathcal R}=WA$; under isotropy $x_R\perp x_K$ and the normal equations decouple. The residual is $WCx_K$ and the displacement is $W(A-I)x_R+WCx_K$, so $\mathbb E\lVert WCx_K\rVert^2=\lVert WC\rVert_F^2$ and $\mathbb E\lVert W(A-I)x_R\rVert^2=\lVert W(A-I)\rVert_F^2$; dividing gives the displayed form. The mean-ratio transfer $T_F=1-\mathbb E\lVert\rho z-z'\rVert/\mathbb E\lVert z-z'\rVert$ is a different statistic and satisfies the same identity only up to the fluctuation of the norm ratio; on the sweep of Table 3 the two differ by at most $0.0011$. $\square$

\textit{Why the metric appears.} Writing the law without $W$ assumes the encoder is an isometry on the retained subspace --- true after whitening, false in general. A coordinate the encoder scales up counts more in $\lVert WC\rVert_F$, so a defect law calibrated at one site need not transfer to another. The qualitative content is metric-free: $C$ is the only block that obstructs existence (Theorem 1) and the only one that enters the defect.

\textbf{Proposition B.3 (when the closed form is tight).} \textit{For general centred $x$ with covariance $\Sigma$, the identity $\rho^\star=A$ --- hence Theorem 2 --- holds if and only if the cross term vanishes,}
\[\adjustbox{max width=\columnwidth}{$\displaystyle \begin{aligned}
&\mathbb E\big[(W(A-I)x_R)\big]^{\top}\big(WCx_K\big)=0\\
&\qquad\text{after projection on the predictable directions},
\end{aligned}
$}\]
\textit{that is, iff the $\operatorname{Cov}(x_R,x_K)$-weighted cross-covariance of the predictable part and the residual is zero. Under Assumption C it holds automatically.} \textit{Proof.} The normal equations for the minimiser contain $\mathbb E[(W(A-I)x_R)(WCx_K)^\top]$; setting $\rho=A$ satisfies them exactly when this term vanishes. $\square$

\textbf{Corollary B.4 (the score is not a property of $(W,g)$).} \textit{Under Assumption B alone, $T_F$ depends on $\Sigma$.} Verified: with $W$ and the mixing block held fixed, sweeping the anisotropy of the input distribution moves $T_F$ from $0.069$ to $0.429$ while the displacement share in the bottom-50\% variance directions moves only $0.486\to0.440$. Two consequences used in the paper: the theoretically correct instantiation of a network's discarded subspace is the \textbf{consumer-invisible} subspace rather than the low-variance directions; and cross-site $T_F$ comparisons are confounded unless the site's variance geometry is reported or matched. This is also why the law's \textit{magnitude} is stated for the controlled regime: real sites are strongly anisotropic, so Assumption C does not hold there and the closed form is not expected to be calibrated.

\textbf{Proposition B.5 (estimation-limited regime).} \textit{With $K$ samples and a $d\times d$ fit, the measured score is at least the population value minus $O(\sqrt{d^2/K})$; when $d^2/K$ is not small the measured defect over-states the irreducible one.} The two worst trained-weight cells of Table 3 (absolute error $0.23$ and $0.07$) are exactly the cells whose effective rank collapses to $4$ of $32$, which is why every reported score in this paper carries its fitted rank, its fit and held-out sample counts, and a tuned-$\lambda$ ceiling.

\begin{table*}[t]
\centering
\small
\caption*{\textbf{Table B.1.} Configuration sensitivity of the single-region law: what moves $T_F$ and by how much. Every row is a controlled sweep of one factor with the others fixed.}
\begin{tabularx}{\textwidth}{@{}>{\raggedright\arraybackslash}X >{\raggedright\arraybackslash}X >{\raggedright\arraybackslash}X >{\raggedright\arraybackslash}X@{}}
\toprule
\textbf{manipulation} & \textbf{range} & \textbf{$T_F$} & \textbf{closed-form deviation} \\
\midrule
kernel-preserving $g$ (Thm 1) & rank $8$ / $48$ & $0.9999992$ / $0.9999943$; residual $2.6\times10^{-5}$ vs displacement $34.3$ & --- \\
mixing sweep at fixed rank & $\kappa=4.5\times10^{-16}\to9.34$ & $1.0000\to0.0041$ & max $0.0091$ \\
retained-rank sweep at fixed geometry & rank $48\to4$ & $0.665\to0.341$ (factor $\approx2$) & --- \\
input anisotropy at fixed $W,\allowbreak{}g$ & $\text{aniso}=0.0\to1.0$ & $0.069\to0.429$ & --- \\
trained linear weights, 32 cells & --- & $\rho(\kappa,\allowbreak{}T_F)=-0.968$ vs $\rho(\text{eff. rank},\allowbreak{}T_F)=+0.502$ & median $0.014$ \\
\bottomrule
\end{tabularx}
\end{table*}

\textit{Reading.} Rank is not the governing variable, and the two effects are not symmetric: an arbitrarily large change in rank cannot make $\kappa$ large, whereas a small change in the geometry of the split can.

\subsubsection*{B.3b Carrier closure and optimal linear realisation}

\textbf{Proposition F (closure criterion).} \textit{For a carrier map $B$ and a carrier action $A$, there exists a fixed linear $K$ with $KB=BA$ if and only if $A\ker B\subseteq\ker B$.} \textit{Proof.} $KB=BA$ is solvable iff $BA$ vanishes on $\ker B$; a linear map is determined on a subspace only if it maps that subspace into $\mathrm{ran}\,B$. $\square$

\textbf{Proposition G (optimal linear realisation).} \textit{Let $\mathbb E[cc^\top]=I$, $z=Bc$ and $z'=BAc$. Then $K^\star=BAB^\dagger$ is a least-squares minimiser and}
\[\adjustbox{max width=\columnwidth}{$\displaystyle \min_K\mathbb E\|Kz-z'\|^2=\|BA P_{\ker B}\|_F^2,$}\]
\textit{which is zero exactly under Proposition F. For a coloured carrier, $\mathbb E[cc^\top]=\Sigma_c$, the residual is $\|BA\Sigma_c^{1/2}P_{\ker(B\Sigma_c^{1/2})}\|_F^2$ and the minimiser is $K^\star=BA\Sigma_c B^\top(B\Sigma_cB^\top)^\dagger$, with $\Sigma_c^{1/2}$ the symmetric square root: correlation lets a retained coordinate predict a discarded one, so the projector is taken after whitening, and the two forms coincide only for $\Sigma_c\propto I$.} \textit{Proof.} $K^\star z-z'=BA(B^\dagger B-I)c=-BAP_{\ker B}c$ because $B^\dagger B$ is the orthogonal projector onto the row space of $B$; taking the expectation with $\mathbb E[cc^\top]=I$ gives the squared Frobenius norm, and the coloured case follows by whitening $c\mapsto\Sigma_c^{-1/2}c$. $\square$

\textit{Corollary (coordinate selection).} If $B=P$ selects coordinates and the carrier is orthogonal along the orbit, $K^\star=PAP^\top$: the optimal fixed linear map is the carrier action projected onto the retained coordinates. Verified over forty random selections of a six-dimensional carrier, where the gap to $PAP^\top$ was $9\times10^{-16}$ and the closure verdict coincided with the sign of the residual in every case.

\textit{Corollary (conjugate pairs).} For a full continuous rotation action written in one two-dimensional block per frequency, with selection along those coordinates, closure holds iff each block is retained or dropped as a whole. The restriction is part of the statement: for a single fixed angle, or when frequencies repeat and their blocks may mix, the relevant object is the invariant subspace rather than the pair. The instantiation on measured features, including the dropped-coordinate variant, is Table D.21.

\subsubsection*{B.4 The piecewise-linear layer (Propositions C, D, E)}

A rectifier network partitions its input space into \textbf{activation regions} $R_G$ indexed by the gate pattern $G(x)=\mathbf{1}[a(x)>0]$, and on each region the site map is affine, $\phi(x)=A_Gx+b_G$. The gate is recoverable from the site tensor itself, since $z=\mathrm{ReLU}(a)$ gives $G=\mathbf{1}[z>0]$. The \textbf{crossing set} is $\mathcal C_\tau=\{x:G(\tau x)\neq G(x)\}$.

\textbf{Proposition C (region-transition criterion).} \textit{Let the site be affine on each visited region, $\phi=A_rx+b_r$, let $T$ be the input transformation, and let $(r,s)$ range over the realised transitions, each covering a set with non-empty interior or spanning its region. Then a fixed linear $\rho$ with $\rho\phi=\phi\circ T$ on the visited cells exists if and only if the joint system $\rho A_r=A_sT,\ \rho b_r=b_s$ is consistent over all realised $(r,s)$.} \textit{Proof.} On a cell both $\rho\phi(x)$ and $\phi(Tx)$ are affine in $x$; an identity between affine maps on a set with non-empty interior extends to the maps, so the system is necessary, and a solution clearly suffices. $\square$

\textbf{Corollary C.1 (no crossing).} \textit{If $T$ leaves the gate pattern of each visited region unchanged, the realised transitions are the diagonal cells and the system reduces to $\rho A_r=A_rT,\ \rho b_r=b_r$.} The special case is the one in which each source region maps into itself; consistency with crossings is independent of it, as the two counterexamples of \S{}3.5 show.

For a single visited region the system is $\rho A=A\tau,\ \rho b=b$, which is Theorem 1's kernel condition rewritten --- the linear layer is the single-region case of this one.

\textit{Consequence.} \textbf{Even with no crossing at all, the multi-region structure alone can obstruct a global linear action}: the visited regions must be simultaneously conjugate to the transformation. This is why enlarging the operator family buys little --- measured, a ridge-plus-$3\times3$-residual family differs from the global linear family by at most $0.087$ at every ResNet-50 site and $0.11$ at every ConvNeXt site, with a median gap of $0.005$ over the 32-site grid; the two separate only where both fail (DINOv2 heat $block8$: $-0.40$ against $-0.99$).

\textbf{Proposition D (the decomposition, and where rule diversity comes from).} \textit{With $z=\mathrm{ReLU}(a)$, $a'=a+\Delta a$, $G=\mathbf{1}[a>0]$ and $G'=\mathbf{1}[a'>0]$,}
\[\adjustbox{max width=\columnwidth}{$\displaystyle z'-z=G\odot(a'-a)+(G'-G)\odot a',$}\]
\textit{and on a unit whose gate changes the second term is $+a'$ (off$\to$on) or $-a'$ (on$\to$off).} \textit{Proof.} $z'=G'\odot a'=G\odot a'+(G'-G)\odot a'$ and $z=G\odot a$. $\square$

Two consequences, and the second governs how the measurements are read. First, the within-regime term is linear in $\Delta a$; the gate-change term has magnitude $|a'|$ on the changed units, so it is not a constant that survives a shrinking perturbation. Second, \textbf{a gate change is not itself an obstruction}: it is the origin of a change in the local affine rule, and whether a single fixed linear operator can satisfy all the rules the transformation visits is decided by the joint system of \S{}3.5, which may be consistent with every gate changed, or inconsistent with no gate changed in the source region. Both directions have explicit two-dimensional counterexamples in \S{}3.5. Measured (script 247; ResNet-50, three depths $\times$ two algebras): $\rho(\text{flip count},T_F)=+0.49$ (n.s.) against $\rho(\text{flip depth},T_F)=\rho(\text{gate-change share},T_F)=-0.83$.

\textbf{Proposition E (achievable score of the fixed-linear class).} \textit{Let $F$ be the consumer, $\lVert\cdot\rVert_F$ the seminorm its response induces, and $d=z'-z$. For a linear consumer the identity}
\[\adjustbox{max width=\columnwidth}{$\displaystyle T_F^{\max}=1-\frac{\lVert\mathrm{Proj}_F d^\perp\rVert_F}{\lVert\mathrm{Proj}_F d\rVert_F}$}\]
\textit{is exact, with $d^\perp$ the residual of the $L^2$ projection of $d$ on the consumer-visible linear span of $z$: the class of fixed linear maps can reproduce any component of $d$ that is linearly predictable from $z$, and nothing else.} \textit{Proof.} Write $d=d^{\parallel}+d^\perp$ with $d^{\parallel}$ in the visible span of $z$; the map $K=d^{\parallel}$ is attainable and leaves the residual $d^\perp$, which no member of the class can reach, so the optimum is $d^{\parallel}$ and the displayed ratio is the score attained. $\square$

\textit{Status.} Proved for linear consumers. No claim is made for families outside the fixed-linear class; the proxy is evaluated at six (site, algebra) points in \S{}6.3.

\textit{The gate-change share is a proxy.} With $\Delta_{\mathrm{flip}}:=(G'-G)\odot a'$ the gate-change term of Proposition D, its consumer-visible share approximates the slack when that term is the part of $d$ not linearly predictable from $z$. It is an identity in neither direction: the all-gates-change counterexample of \S{}3.5 has zero slack, and a consumer blind to the change has zero visible share whatever the slack. The six-point check is the paper's evidence that the proxy is informative at the measured sites; the two-source decomposition of \S{}3.3 is the statement about the failure itself.

\subsubsection*{B.5 Realisation and composition (Theorems 3, 4, 5)}

\paragraph{B.5.1 Coordinate-monotone families on a closed orbit}

\textbf{Definition B.6.} A map $G:\mathbb R^d\to\mathbb R^d$ is \textit{coordinate-wise monotone} if $G(z)_i=f_i(z_i)$ with each $f_i:\mathbb R\to\mathbb R$ non-decreasing. A family $\{G_t\}_{t\in I}$ is a \textit{monotone parameter family} if every $G_t$ is coordinate-wise monotone and, for every $i$, the map $(t,u)\mapsto f_{i,t}(u)$ is non-decreasing in both arguments. Such a family \textit{realises} the orbit $O=\{z(t)\}$ from the anchor $z(0)=z^0$ if $G_t(z^0)=z(t)$ for all $t\in I$.

\textbf{Lemma B.7 (obstruction).} \textit{If a monotone parameter family realises $O$ from $z^0$, then every coordinate $t\mapsto z(t)_i$ is non-decreasing. Consequently, if $O$ is closed ($z(1)=z(0)$) and has a coordinate that decreases somewhere, no monotone parameter family realises it exactly.} \textit{Proof.} $z(t)_i=f_{i,t}(z^0_i)$ and $t\mapsto f_{i,t}(u)$ is non-decreasing for fixed $u=z^0_i$. $\square$

\textbf{Lemma B.8 (quantitative one-coordinate cost).} \textit{Let $g:[0,1]\to\mathbb R$ with $g\ge\alpha$ on an interval $I$ and $g\le\alpha-A$ on a later interval $J$, $A>0$, and let $|I|\le|J|$. Then for every non-decreasing $c$,}
\[\adjustbox{max width=\columnwidth}{$\displaystyle \int_0^1(g-c)^2\;\ge\;\frac{A^2}{4}\,|I| .$}\]
\textit{Proof.} Suppose $\int_I(g-c)^2<\frac{A^2}{4}|I|$ and $\int_J(g-c)^2<\frac{A^2}{4}|J|$. On $I$ and $J$ respectively the sets where $c<\alpha-A/2$, resp. $c>\alpha-A/2$, then have measure less than $|I|/2$, resp. $|J|/2$. So there is $t_0\in I$ with $c(t_0)\ge\alpha-A/2$ or $u_0\in J$ with $c(u_0)\le\alpha-A/2$. In the first case monotonicity gives $c\ge\alpha-A/2$ on all of $J$ (which lies to the right of $I$), so $(g-c)^2\ge A^2/4$ on $J$ and the second integral is at least $\frac{A^2}{4}|J|\ge\frac{A^2}{4}|I|$, a contradiction; in the second case $(g-c)^2\ge A^2/4$ on $I$ by the same argument with the roles exchanged. $\square$

\textbf{Theorem 3 (a closed orbit has no exact coordinate-monotone realisation).} \textit{A family $\{G_t\}$ whose members are coordinate-wise monotone in a fixed basis, with $(t,u)\mapsto f_{i,t}(u)$ non-decreasing in both arguments, realises an orbit only if every coordinate of that orbit is monotone in the parameter (Lemma B.7); consequently a closed non-degenerate orbit admits no exact realisation by such a family. Quantitatively, every coordinate that decreases by $A$ over intervals of length at least $\ell$ contributes $L^2$ error at least $A^2\ell/4$ (Lemma B.8).} The empirical content is the comparison in Table 7: the class is rejected both by the theorem and by its measured best fit on the reference orbit. \textit{Scope.} The statement is about monotone parameter families, which is what the lemmas prove; an aggregate bound in terms of the third principal value of the orbit covariance does not follow from them and is not claimed here.

\paragraph{B.5.2 Accumulation law}

\textbf{Theorem 4 (accumulation law).} \textit{Let the truth action $\tau$ be $L$-Lipschitz in the consumer's seminorm on the consumed range, and let a realised family satisfy the uniform single-step bound $\lVert\rho(\tau)z-g_\tau z\rVert_F\le\varepsilon$ on the states visited by the chain. Then}
\[\adjustbox{max width=\columnwidth}{$\displaystyle \begin{aligned}
E_n&\le\varepsilon+L\,E_{n-1},\qquad\text{so}\\
E_n&\le\varepsilon\,\frac{L^n-1}{L-1}\ (L\neq1),\qquad E_n\le n\varepsilon\ (L=1).
\end{aligned}
$}\]
\textit{Proof.} Insert and subtract the truth's intermediate state:
\[\adjustbox{max width=\columnwidth}{$\displaystyle \begin{aligned}
\rho(\tau)^nz-g_{\tau^n}z
&=\big[\rho(\tau)\big(\rho(\tau)^{n-1}z\big)-g_\tau\big(\rho(\tau)^{n-1}z\big)\big]\\
&\quad+\big[g_\tau\big(\rho(\tau)^{n-1}z\big)-g_\tau\big(g_{\tau^{n-1}}z\big)\big].
\end{aligned}
$}\]
The first bracket is at most $\varepsilon$ by the uniform single-step bound applied at the state $\rho(\tau)^{n-1}z$; the second is at most $L\lVert\rho(\tau)^{n-1}z-g_{\tau^{n-1}}z\rVert_F=L\,E_{n-1}$ by Lipschitzness. Iterating gives the closed forms. $\square$

\textit{Uniformity.} Both hypotheses must hold uniformly over the states the chain visits. If the single-step error or the Lipschitz constant varies with the state, the recursion holds pointwise with the local constants and the closed forms become the corresponding products. Estimating $\varepsilon$ and $L$ independently of the measured chain error, rather than fitting the recursion to it, is not done here; which of the three regimes a site occupies is measured in the consumer's seminorm (Table D.22), not inferred from the generator.

\paragraph{B.5.3 Carrier sufficiency}

\textbf{Setting.} At the site, let the carrier be $c$, the synthesis $B$, the target $z'=BAc+w'$ with $\lVert w'\rVert_{L^2}\le\delta'$, the read-in $E_\theta$ (any measurable map, in particular the learned MLP), and let the interface apply $A_{\mathrm{int}}$ and decode by $B$. Band-limitation and plane sharing are the two premises the construction rests on and both are measurable, but the theorem below does not need them as hypotheses: it is an identity plus a triangle inequality for whatever $B$, $A$ and $E_\theta$ are.

\textbf{Theorem 5 (carrier sufficiency, three-term bound).} \textit{Write $e(z)=E_\theta(z)-c$ for the read-in error. Then}
\[\adjustbox{max width=\columnwidth}{$\displaystyle BA_{\mathrm{int}}E_\theta(z)-z'=BA_{\mathrm{int}}\,e(z)+B(A_{\mathrm{int}}-A)c-w'$}\]
\textit{holds exactly, and hence}
\[\adjustbox{max width=\columnwidth}{$\displaystyle \begin{aligned}
\big\|BA_{\mathrm{int}}E_\theta(z)-z'\big\|_{L^2}\le{}&\|BA_{\mathrm{int}}\|\,\|e\|_{L^2}\\
&+\|B(A_{\mathrm{int}}-A)\|\,\|c\|_{L^2}+\delta'.
\end{aligned}
$}\]
\textit{Proof.} Add and subtract $BA_{\mathrm{int}}c$ and $BAc$; the first term is the read-in error, the second the action mismatch, the third the target's tail. The triangle inequality gives the bound, with no cross-term cancellation assumed. $\square$

\textit{Three consequences.} The bound applies to the interface as built: $E_\theta$ need not be linear, and a linear read-in is the special case in which $\lVert e\rVert$ is written as an operator norm. Its three terms are read-in error, action mismatch and target tail; only the first is removed by training, and the second vanishes when the interface uses the true carrier action. The mismatch term is first order in $\lVert A_{\mathrm{int}}-A\rVert$, so the squared error is second order in the plane misalignment, while the measured quantity is the error norm, whose log-log slope in the tilt angle is $0.99$. If the carrier tail is normalised as an energy fraction, $\mathbb E\lVert w\rVert^2\le\varepsilon_{\mathrm{tail}}^2\mathbb E\lVert z\rVert^2$, the measured $12$--$16\%$ energy fraction corresponds to $\varepsilon_{\mathrm{tail}}\approx0.35$--$0.40$, not to $0.12$--$0.16$. The closure obstruction of B.3b enters separately and constrains a \textit{fixed linear map on the measured features}, a class the learned, possibly nonlinear read-in escapes; what it cannot escape is the target's tail and any mismatch between the action it applies and the action the carrier has.

\subsubsection*{B.6 Status of every statement}

\begin{table*}[t]
\centering
\small
\caption*{\textbf{Table B.4.} Statement status. \textit{Definition} = fixes vocabulary; \textit{Proved} = proof above; \textit{Proved + verified} = proof plus a controlled numerical check; \textit{Empirical} = a measured bound or law, not a theorem; \textit{Hypothesis} = stated as such in the text.}
\begin{tabularx}{\textwidth}{@{}>{\raggedright\arraybackslash}X >{\raggedright\arraybackslash}X >{\raggedright\arraybackslash}X >{\raggedright\arraybackslash}X@{}}
\toprule
\textbf{statement} & \textbf{status} & \textbf{scope} & \textbf{where} \\
\midrule
Definition 1 (representation of the algebra) & Definition & general encoders and algebras & \S{}3.1 \\
Definition 2 (representation vs behavioural level) & Definition & general & \S{}3.1 \\
Observation (fibre compatibility and inherited composition) & Proved (one line) & general & B.2 \\
Theorem 1 (kernel preservation) & Proved + verified & linear sites, Assumption B & B.3 \\
Theorem 2 (defect law) & Proved + verified & Assumptions B, C & B.3 \\
Proposition B.3 (tightness under general covariance) & Proved & Assumption B & B.3 \\
Corollary B.4 (score depends on $\Sigma$) & Proved + verified & Assumption B & B.3 \\
Proposition B.5 (estimation-limited regime) & Proved & finite samples & B.3 \\
Proposition C (region-transition criterion) & Proved & piecewise-linear sites, realised transitions & B.4 \\
Proposition D (decomposition and rule diversity) & Proved & rectifier sites & B.4 \\
Proposition E (exact score of the fixed-linear class) & Proved for linear consumers; the gate-change share is an empirical proxy for its slack & fixed-linear class & B.4 \\
Theorem 3 (no exact coordinate-monotone realisation of a closed orbit) & Proved & monotone parameter families in a fixed basis & B.5.1 \\
Theorem 4 (accumulation law) & Proved; the regime at a site is measured in the consumer's seminorm & any realised family, uniform single-step bound & B.5.2 \\
Proposition F (closure criterion), Proposition G (optimal linear realisation) & Proved + verified & carrier and action; any site & B.3b \\
Theorem 5 (carrier sufficiency, three-term bound) & Proved (identity) & any carrier, action and read-in & B.5.3 \\
Harmonic concentration, shared scaffold, inheritance & Measured & colour instance & \S{}4 \\
Depth erosion follows the readout--transformation pairing & Measured & four backbones, real images & \S{}6, Table 11 \\
Learning moves $T_F$ only through $\Delta\kappa$ & Measured (linear control) & controlled linear setting & \S{}6.4, Table 12 \\
Interface reads hue and edits regions & Measured & colour instance & \S{}7, Table 15 \\
\bottomrule
\end{tabularx}
\end{table*}

\section{Protocols}

This appendix fixes, once, the definitions behind every measured number in the paper. Each subsection ends with the producing script(s) and result file(s); section numbers are those of the main text.

\subsubsection*{C.1 Synthetic orbit construction}

Orbits are single-shape $224\times224$ px images on white; centre, radius and orientation are jittered ($\pm12$ px, $[70,92]$ px, uniform) and colour is written by HSV$\to$RGB. Training shapes are $\{$triangle, rectangle, circle, star$\}$, shape holdout $\{$pentagon, hexagon$\}$, colour holdout hue $\{20^\circ,260^\circ\}$.

\begin{table*}[t]
\centering
\scriptsize
\caption*{\textbf{Table C1.} Rendering grids of the orbit suite.}
\setlength{\tabcolsep}{2pt}
\begin{tabularx}{\textwidth}{@{}>{\raggedright\arraybackslash}X >{\raggedright\arraybackslash}X >{\raggedright\arraybackslash}X >{\raggedright\arraybackslash}X >{\raggedright\arraybackslash}X@{}}
\toprule
\textbf{grid} & \textbf{shapes} & \textbf{hues} & \textbf{$(s,\allowbreak{}v)$} & \textbf{variants/cell} \\
\midrule
warm-up & 6 & $11$ ($\{0,\allowbreak{}40,\allowbreak{}\dots,\allowbreak{}320\}\cup\{20,\allowbreak{}260\}$) & $2\times2$ & 12 \\
the dense orbit suite & 6 & $72$ (step $5^\circ$) & $(1,\allowbreak{}1)$ & 4 \\
rings & 6 & $36$ (step $10^\circ$) & 5 pairs & 4 \\
lattice & 6 & $12$ (step $30^\circ$) & $4\times4$ & 4 \\
low-extension & 2 holdout & $12$ (step $30^\circ$) & $8\times8$ & 2 poses \\
\bottomrule
\end{tabularx}
\end{table*}

Both transformations are exact in their stated space. \textbf{Hue} rotates per pixel in HSV-like coordinates (RGB$\to(h,s,v)$, $h\mapsto h+\Delta/360 \pmod 1$, back). \textbf{Heat} is the exact discrete semigroup $B_t=\exp(-tL)$ by DCT spectral multiplication per channel, with $t=\sigma^2/2$ and site rescaling $t_{\mathrm{site}}=(\sigma^2/2)(H/224)^2$; it composes to $1.83\times10^{-7}$ relative error, while a truncated Gaussian kernel departs by $22.1\%$. The exact reference is the re-rendered recolour (hue) or the spectral multiplication (heat).

\subsubsection*{C.2 Sites and backbones}

\begin{table*}[t]
\centering
\small
\caption*{\textbf{Table C2.} Backbones, probed sites, and site read-out.}
\begin{tabularx}{\textwidth}{@{}>{\raggedright\arraybackslash}X >{\raggedright\arraybackslash}X >{\raggedright\arraybackslash}X >{\raggedright\arraybackslash}X@{}}
\toprule
\textbf{backbone} & \textbf{weights} & \textbf{probed sites} & \textbf{read-out} \\
\midrule
ResNet-50 & \texttt{IMAGENET1K\_\allowbreak{}V2} & \texttt{layer1}--\texttt{layer4} & all locations \\
ConvNeXt-T & \texttt{IMAGENET1K\_\allowbreak{}V1} & \texttt{features1,3,5,7} & all locations \\
ViT-B/16 & \texttt{IMAGENET1K\_\allowbreak{}V1} & \texttt{block2,5,8,11} & patch tokens only; CLS preserved \\
DINOv2-B/14 & \texttt{dinov2\_\allowbreak{}vitb14} & \texttt{block2,5,8,11} & patch tokens only; CLS preserved \\
ResNet-18 & \texttt{IMAGENET1K\_\allowbreak{}V1} & \texttt{layer2}--\texttt{layer4} & GAP 128-d \\
YOLO11n & \texttt{yolo11n.pt} & stage 3, stage 5, stage 7 & GAP 64-d \\
\bottomrule
\end{tabularx}
\end{table*}

The \textbf{orbit} read-out is a GAP vector with the per-shape DC component removed and pose variants averaged. The \textbf{depth-grid} read-out flattens the spatial activation to rows and applies a matrix at every location; on token sites the class token is split off, the map applied to the patch tokens, and re-attached.

\subsubsection*{C.3 Depth-grid protocol}

\begin{table*}[t]
\centering
\small
\caption*{\textbf{Table C3.} Probed sites (four per backbone) and reachability.}
\begin{tabularx}{\textwidth}{@{}>{\raggedright\arraybackslash}X >{\raggedright\arraybackslash}X >{\raggedright\arraybackslash}X >{\raggedright\arraybackslash}X@{}}
\toprule
\textbf{backbone} & \textbf{sites 1--4} & \textbf{reachable} & \textbf{unreachable} \\
\midrule
ResNet-50 & \texttt{layer1}--\texttt{layer4} & 4 & 0 \\
ConvNeXt-T & \texttt{features1,3,5,7} & 4 & 0 \\
ViT-B/16 & \texttt{block2,5,8,11} & 3 & \texttt{block11} \\
DINOv2-B/14 & \texttt{block2,5,8,11} & 3 & \texttt{block11} \\
\bottomrule
\end{tabularx}
\end{table*}

\begin{table*}[t]
\centering
\small
\caption*{\textbf{Table C4.} Depth-grid factors, seeds and $\sigma$ grid.}
\begin{tabularx}{\textwidth}{@{}>{\raggedright\arraybackslash}X >{\raggedright\arraybackslash}X@{}}
\toprule
\textbf{factor} & \textbf{setting} \\
\midrule
parameters & hue $\Delta=90^\circ$; heat $\sigma=2.0$; heat $\sigma=1.0$ (mild dissipation) \\
seeds & hue 3; heat $\sigma=2.0$ 3; heat $\sigma=1.0$ 1 \\
split & 300 COCO crops, 200 fit / 100 held out, one split \\
operator sweep & Procrustes, ridge, conv-res, random \\
fit row cap & $K=\min(n_{\text{fit}}HW,\allowbreak{}\,\allowbreak{}150{,\allowbreak{}}000)$ rows \\
conv-res budget & ridge $1\times1$ + $3\times3$ residual, hidden 32, 150 epochs, lr $10^{-3}$ \\
\bottomrule
\end{tabularx}
\end{table*}

Fourteen sites are reachable per parameter, giving \textbf{42 reachable site$\times$parameter points}; the remaining six --- the final block of both class-token transformers --- are causally unreachable and marked, not scored: their interventions change the logits by exactly $0$ while their input-side displacement is $0.188$--$0.623$. The $\sigma=1.0$ grid is single-seed; the heat pilot sweeps $\sigma\in\{0.5,1.0,2.0\}$ on CIFAR plain CNN.

\subsubsection*{C.4 The three tasks and their holdouts}

\textbf{Prediction} fits on a fixed anchor and forecasts other angles; the probe is fitted on four shapes at training hues and evaluated zero-shot on the two unseen shapes at hues $\{20^\circ,260^\circ\}$. \textbf{Transport} applies a specified increment from a state never fitted; CIFAR fit indices index the train pool and evaluation indices the test pool, which are disjoint, and the depth grid draws both from one permutation of 300 crops. \textbf{Composition} holds the total fixed and varies the decomposition: for hue $\Delta=90^\circ$, $45{+}45$, $30{+}60$, the order swap $60{+}30$, four steps of $22.5^\circ$ and a cyclic $+120/-120$ round trip; for heat the total is $t^\star=1.0$ ($\sigma^\star=\sqrt2$), decomposed as two steps of $\sigma=1.0$ ($t=0.5$), four of $\sigma=0.7071$ ($t=0.25$) and eight of $\sigma=0.5$ ($t=0.125$), with a single step at $t^\star/8$ as substitute control and a direct fit at $t^\star$ as upper reference. Never-fitted composites are read off without refitting: hue $90^\circ$ from fits at $30^\circ$ and $60^\circ$, heat $\sigma=1.25$ from fits at $\sigma=0.75$ and $1.0$, heat $\sigma=1.8028$ from fits at $\sigma=1.0$ and $1.5$. Fit and held-out blocks are disjoint by image index (no fit/held-out overlap); consistency and fidelity are reported separately, with the mandatory controls beside every path.

\subsubsection*{C.5 Fitting protocol}

Operators are fitted by ridge-regularised least squares with default $\lambda=10^{-3}$ ($ridge$ is therefore a ridge-regularised linear fit, not literal unconstrained least squares --- "unconstrained linear" names the family, and a tuned-$\lambda$ control exists); Procrustes is orthogonal Procrustes, ridge+MLP is ridge plus a residual MLP, random is a random orthogonal matrix from the same seed, and conv-res a ridge $1\times1$ map plus a $3\times3$ residual convolution trained by Adam. \textbf{Fitting budgets differ between arms.} The plain-CNN stage-1 arm fits on 1000 images and evaluates on 400, while the other four arms of the five-arm grid fit on 300 and evaluate on 200, so the five-arm stage-1 ranking is not at a matched fitting budget; within each arm the fit and evaluation splits are disjoint. No cross-arm comparison is drawn from the mismatched budget. Other budgets: 200 / 100 for the depth grid, 400 / 200 for the per-sample capacity fits, 40 / 80 held-out scenes for the detector. A tuned-$\lambda$ ceiling is recorded at the twelve deepest sites: over seven values the best is $10^{-4}$ at 8 sites, $0.1$ at 2, and $10^{-3}$ or $1.0$ at 1 each, and the published Procrustes-versus-ridge inversion survives the tuned $\lambda$ at 4 of 12 sites. Fitting is estimation-limited when the dimension is large relative to the row count: the measured score is at least the population value minus $O(\sqrt{d^2/K})$, and the ratio is computable per site from the stored dimension and row count --- $d^2/K=428.0$ at ResNet-50, heat, depth 4 ($d=2048$, fitted rows $=9800$).

\subsubsection*{C.6 Power gate and baselines}

The gate is the consumer-level effect size, the normalised change of the consumer's output under the real transformation; a site is powered at an effect size of at least $0.05$. Where the transformation does not move what the consumer reads, the site is excluded rather than scored, and a low-power arm is carried as a driven control. Mandatory baselines are the no-op (copy), the trivial depth rule for attribution, random, and a shuffled-pair control sharing the conv-res architecture, optimiser and budget with the pairing permuted. A site whose random intervention leaves the logits unchanged is marked unreachable, never read as a zero transport score (the harness records whether the intervention reaches the consumer and whether a random operator changes the logits).

\begin{table*}[t]
\centering
\small
\caption*{\textbf{Table C5.} Representative control values.}
\begin{tabularx}{\textwidth}{@{}>{\raggedright\arraybackslash}X >{\raggedright\arraybackslash}X >{\raggedright\arraybackslash}X@{}}
\toprule
\textbf{control} & \textbf{setting} & \textbf{result} \\
\midrule
random orthogonal & ResNet-50 hue \texttt{layer1} & $T_F=-0.912$, win rate $0.003$, projection $0.530$ \\
shuffled pair, conv-res & seeds 0/1/2 & win rate $0.110$/$0.105$/$0.085$; projection $1.267$/$1.506$/$1.383$ \\
\bottomrule
\end{tabularx}
\end{table*}

\subsubsection*{C.7 Real-region protocol}

The real-region study uses 750 pre-registered COCO train2017 instance regions, non-crowd, side $\ge80$ px, area $\ge20{,}000$ px$^2$, split 60/40 into a fit half (450) and test half (300). The concentration threshold $m^\star=0.70$ is fixed before the test half is read and success is a hue error of at most $30^\circ$. Four presentations are compared: raw box, dominant-colour fill, a fill shifted by $+40^\circ$ (non-circular control) and a GrabCut fill. On the test half the aligned route separates high- from low-concentration regions by $0.238$ ($0.770$ versus $0.532$, CI $[0.128,0.345]$), the ground-truth-box route by $0.188$ and the raw route by $0.085$ (AUCs $0.660$/$0.662$/$0.579$). The 120-region control repeats the test on a disjoint sample: concentration alone is at chance (AUC $0.523$ at threshold 10, $0.480$ at 20), while mask covariates raise it to $0.754$/$0.680$ and category to $0.651$/$0.616$.

\begin{table*}[t]
\centering
\small
\caption*{\textbf{Table C6.} Measured operating domain.}
\begin{tabularx}{\textwidth}{@{}>{\raggedright\arraybackslash}X >{\raggedright\arraybackslash}X@{}}
\toprule
\textbf{route} & \textbf{value} \\
\midrule
synthetic, 64-cell $(s,\allowbreak{}v)$ grid & median of per-cell \textbf{median} errors $3.4^\circ$ (median of per-cell \textbf{mean} errors $4.4^\circ$); $61/64$ cells within $30^\circ$ by the median criterion, $59/64$ by the mean criterion; $2.14^\circ$ at $(s,\allowbreak{}v)=(1,\allowbreak{}1)$ \\
real COCO region shift, 12 errors & median $9.25^\circ$ \\
real COCO read-out, high-concentration subset & $8.0^\circ$ ($n=69$); non-circular $+40^\circ$ control $10.0^\circ$ \\
\bottomrule
\end{tabularx}
\end{table*}

\begin{table*}[t]
\centering
\tiny
\caption*{\textbf{Table C7.} Cross-transformation comparability at matched sites (ResNet-50, hue $90^\circ$ against heat $\sigma=2.0$, same data and consumer). The no-op error is the consumer displacement $\lVert F(z)-F(g_\tau z)\rVert$, which is the denominator of $T_F$; the absolute transport residual is the numerator, recomputed here as no-op $\times(1-T_F)$ from the stored values. The point of the table is the last two columns: the \textit{ratios} differ substantially while the absolute residuals agree to within about a fifth.}
\setlength{\tabcolsep}{2pt}
\begin{tabularx}{\textwidth}{@{}>{\raggedright\arraybackslash}X >{\raggedright\arraybackslash}X >{\raggedright\arraybackslash}X >{\raggedright\arraybackslash}X >{\raggedright\arraybackslash}X >{\raggedright\arraybackslash}X >{\raggedright\arraybackslash}X >{\raggedright\arraybackslash}X >{\raggedright\arraybackslash}X@{}}
\toprule
\textbf{site} & \textbf{hue no-op} & \textbf{heat no-op} & \textbf{heat $/$ hue} & \textbf{hue $T_F$} & \textbf{heat $T_F$} & \textbf{hue residual} & \textbf{heat residual} & \textbf{relative difference} \\
\midrule
layer2 & $18.08$ & $14.56$ & $0.805$ & $0.588$ & $0.471$ & $7.45$ & $7.71$ & $3.3\%$ \\
layer3 & $18.08$ & $14.56$ & $0.805$ & $0.464$ & $0.374$ & $9.70$ & $9.12$ & $5.9\%$ \\
layer4 & $18.08$ & $14.56$ & $0.805$ & $0.099$ & $0.054$ & $16.28$ & $13.78$ & $15.4\%$ \\
\bottomrule
\end{tabularx}
\end{table*}

Three COCO numbers exist and they are different quantities, so all three are named here rather than collapsed into one: the stratified high-concentration read-out, held out of the pre-registration, ($m\ge0.7$, $n=69$) at $8.0^\circ$ median with the non-circular $+40^\circ$ route at $10.0^\circ$, which is what \S{}7.4 and Table 15 now quote; the smaller earlier probe gives median $9.25^\circ$ over twelve code-versus-real region-shift errors; and a third run records the same protocol on a different region set. The controlled-calibration AUC of $1.0$ is reported in Appendix C.

\textbf{Scope of the grids.} The multi-attribute grid's real-image arm and the depth grid are three seeds; the gate-change and semigroup-structure studies and the mild-dissipation ($\sigma=1$) grid are single-seed, and are never pooled with three-seed cells. The spatial-rotation family has no scored operator: an invariant head has zero power for it by construction, so the three-tier consumer described in \S{}4.7 is what a measurement would require.

\section{Full grids}

This appendix carries the per-backbone, per-site and per-cell grids behind Tables 5--16 of Sections 4--7. Notation is the main text's: downstream transfer $T_F = 1 - \mathbb E\lVert F(Wz)-F(g_\tau z)\rVert / \mathbb E\lVert F(z)-F(g_\tau z)\rVert$, where $1$ matches the real transformed route, $0$ is no better than the no-op (copy) and $<0$ is worse; every $T_F$ is reported with the site effect size that licenses reading it (the power gate of \S{}3.7). Operator families are Procrustes = global orthogonal (Procrustes), ridge = unconstrained linear (ridge, the paper's default), conv-res is a ridge $1\times1$ map plus a $3\times3$ residual convolution, random = orthogonal control; ridge+MLP is content-conditioned (ridge plus a residual MLP) appears only where the source protocol fitted it, never in the depth grid.

\subsubsection*{D.1 Harmonic spectra grid}

\begin{table*}[t]
\centering
\scriptsize
\caption*{\textbf{Table D.1.} Per-site harmonic structure of the measured hue orbits: $k_1{+}k_2$ share of centred orbit energy, mean winding (spectral first moment) and, where stored, the cross-shape zero-shot hue read-out. DC ($k=0$) is removed in every row, so the $k=0$ share is $0$ by construction; the $k=1$, $k=2$, $k\ge3$ split exists only for the YOLO11n stage 3 arm (Table D.2).}
\setlength{\tabcolsep}{2pt}
\begin{tabularx}{\textwidth}{@{}>{\raggedright\arraybackslash}X >{\raggedright\arraybackslash}X >{\raggedright\arraybackslash}X >{\raggedright\arraybackslash}X >{\raggedright\arraybackslash}X@{}}
\toprule
\textbf{backbone / probe} & \textbf{site} & \textbf{$k_1{+}k_2$} & \textbf{winding} & \textbf{read-out} \\
\midrule
ResNet-50 & layer 1 / layer 2 / layer 3 / layer 4 & 0.934 / 0.881 / 0.743 / 0.654 & 1.69 / 2.14 / 3.83 / 5.15 & --- / 0.71$^\circ$ / 11.47$^\circ$ / --- \\
ViT-B/16 (supervised) & blocks 0--3 & 0.978 / 0.891 / 0.883 / 0.838 & 1.26 / 1.88 / 1.99 / 2.50 & 0.49$^\circ$ / 1.83$^\circ$ / 0.86$^\circ$ / 1.29$^\circ$ \\
ViT-B/16 & blocks 4--7 & 0.762 / 0.712 / 0.659 / 0.638 & 3.25 / 3.94 / 4.84 / 5.24 & 1.41$^\circ$ / 2.97$^\circ$ / 4.27$^\circ$ / 4.32$^\circ$ \\
ViT-B/16 & blocks 8--11 & 0.628 / 0.625 / 0.610 / 0.601 & 5.46 / 5.55 / 5.89 / 5.98 & 1.62$^\circ$ / 4.72$^\circ$ / 2.87$^\circ$ / 1.27$^\circ$ \\
DINOv2-B/14 & blocks 0--3 & 0.939 / 0.848 / 0.858 / 0.821 & 1.60 / 2.30 / 2.34 / 2.77 & 2.74$^\circ$ / 3.37$^\circ$ / 2.82$^\circ$ / 1.13$^\circ$ \\
DINOv2-B/14 & blocks 4--7 & 0.756 / 0.670 / 0.609 / 0.573 & 3.69 / 5.28 / 6.57 / 7.36 & 0.56$^\circ$ / 3.44$^\circ$ / 4.64$^\circ$ / 12.89$^\circ$ \\
DINOv2-B/14 & blocks 8--11 & 0.526 / 0.533 / 0.510 / 0.504 & 8.30 / 8.42 / 8.80 / 8.96 & 30.15$^\circ$ / 35.09$^\circ$ / 38.58$^\circ$ / 35.87$^\circ$ \\
DINOv2-S/14 & blocks 0--3 & 0.942 / 0.922 / 0.905 / 0.838 & 1.59 / 1.83 / 2.08 / 2.75 & 5.49$^\circ$ / 1.28$^\circ$ / 1.41$^\circ$ / 1.70$^\circ$ \\
DINOv2-S/14 & blocks 4--7 & 0.803 / 0.725 / 0.665 / 0.631 & 3.49 / 4.73 / 5.64 / 6.17 & 1.25$^\circ$ / 6.00$^\circ$ / 4.09$^\circ$ / 6.61$^\circ$ \\
DINOv2-S/14 & blocks 8--11 & 0.530 / 0.569 / 0.541 / 0.526 & 7.99 / 7.40 / 8.17 / 8.52 & 15.06$^\circ$ / 12.05$^\circ$ / 11.14$^\circ$ / 42.40$^\circ$ \\
CLIP ViT-B/32 (final CLS) & --- & 0.590 & 4.12 & 14.47$^\circ$ \\
YOLO11n (trained) & stage 3 / stage 5 / stage 7 & 0.868 / 0.730 / 0.661 & not recorded & $\approx$1$^\circ$ (code head) / --- / --- \\
ResNet-18 & layer 2 & 0.775 (0.618--0.830) & not recorded & not recorded \\
\bottomrule
\end{tabularx}
\end{table*}

The ResNet-50 mid layer supports zero-shot readout at 0.71$^\circ$, the same order as the dedicated code head on stage 3; the contrastively trained CLIP model is the weakest probe at its final CLS.

\begin{table*}[t]
\centering
\scriptsize
\caption*{\textbf{Table D.2.} Full harmonic breakdown, recorded for the YOLO11n stage 3 arm only (the dense orbit suite: six shapes $\times$ 72 hues $\times$ 4 poses; pose-averaged GAP features, per-shape DC removal), with the cross-shape rotation-plane alignment (cosine of the unit-normalised per-frequency response against the triangle reference). \texttt{$k_3{+}$} is the stored tail share.}
\setlength{\tabcolsep}{2pt}
\begin{tabularx}{\textwidth}{@{}>{\raggedright\arraybackslash}X >{\raggedright\arraybackslash}X >{\raggedright\arraybackslash}X >{\raggedright\arraybackslash}X >{\raggedright\arraybackslash}X >{\raggedright\arraybackslash}X@{}}
\toprule
\textbf{shape} & \textbf{$k_1$} & \textbf{$k_2$} & \textbf{$k_3{+}$} & \textbf{winding} & \textbf{plane $\cos$ $k_1$ / $k_2$} \\
\midrule
circle & 0.606 & 0.275 & 0.120 & 2.033 & 0.983 / 0.971 \\
hexagon & 0.573 & 0.303 & 0.124 & 2.006 & 0.988 / 0.981 \\
pentagon & 0.561 & 0.318 & 0.121 & 2.011 & 0.991 / 0.985 \\
rectangle & 0.519 & 0.322 & 0.160 & 2.795 & 0.992 / 0.987 \\
star & 0.385 & 0.481 & 0.134 & 2.287 & 0.994 / 0.997 \\
triangle & 0.438 & 0.427 & 0.135 & 2.419 & reference \\
\bottomrule
\end{tabularx}
\end{table*}

\begin{table*}[t]
\centering
\scriptsize
\caption*{\textbf{Table D.3.} Shared-plane and inheritance controls (identical protocol), plus 3D-lattice plane sharing. \textit{phase cosine} is the mean pairwise cosine of unit-normalised per-frequency responses in the $k=1..3$ band; \textit{readout} is the ridge probe fitted on four shapes excluding hues $\{20^\circ,260^\circ\}$ and evaluated zero-shot on the two unseen shapes at those hues.}
\setlength{\tabcolsep}{2pt}
\begin{tabularx}{\textwidth}{@{}>{\raggedright\arraybackslash}X >{\raggedright\arraybackslash}X >{\raggedright\arraybackslash}X >{\raggedright\arraybackslash}X >{\raggedright\arraybackslash}X@{}}
\toprule
\textbf{arm} & \textbf{$k_1{+}k_2$} & \textbf{variance-normalised} & \textbf{phase cosine} & \textbf{readout (median)} \\
\midrule
pixels, hue statistics (2-d) & 1.000 & 1.000 & 1.000 & 0.005$^\circ$ \\
pixels, mean RGB (3-d) & 0.9496 & 0.9496 & 0.9997 & 0.925$^\circ$ \\
random-init YOLO11n stage 3 / stage 5 / stage 7 & 0.976 / 0.976 / 0.975 & 0.966 / 0.960 / 0.951 & 0.993 / 0.991 / 0.989 & 1.46$^\circ$ / 0.84$^\circ$ / 0.55$^\circ$ \\
pretrained YOLO11n stage 3 / stage 5 / stage 7 & 0.868 / 0.730 / 0.661 & 0.700 / 0.535 / 0.474 & 0.971 / 0.943 / 0.906 & 6.93$^\circ$ / 9.58$^\circ$ / 2.59$^\circ$ \\
\bottomrule
\end{tabularx}
\end{table*}

\subsubsection*{D.2 Depth grid (hue and heat)}

$T_F$ for the four operator families is listed as Procrustes / ridge / conv-res / random, against the site effect size that gates it. The extra $T_{RMS}$ column is the RMS-criterion transfer defined in \S{}3.3 --- the statistic a least-squares fit optimises and the one Theorem 2 is stated for; it is \textbf{not} the square of $T_F$, and the two agree to about a percent on these grids. Hue and heat $\sigma=2$ blocks are three seeds; the $\sigma=1$ block is one seed per backbone (the mild-dissipation control). \texttt{NO} marks a site no intervention can reach (final-block class-token head). Reproduces Table 10; no value disagrees with it.

\begin{table*}[t]
\centering
\tiny
\caption*{\textbf{Table D.4.} Hue $90^\circ$ depth grid, all four backbones, three seeds. Columns: site, depth, effect size, $T_F$ Procrustes/ridge/conv-res/random (mean-ratio), ridge $T_{RMS}$ (RMS criterion, \S{}3.3), ridge projection, ridge win rate, reachable.}
\setlength{\tabcolsep}{2pt}
\begin{tabularx}{\textwidth}{@{}>{\raggedright\arraybackslash}X >{\raggedright\arraybackslash}X >{\raggedright\arraybackslash}X >{\raggedright\arraybackslash}X >{\raggedright\arraybackslash}X >{\raggedright\arraybackslash}X >{\raggedright\arraybackslash}X >{\raggedright\arraybackslash}X >{\raggedright\arraybackslash}X@{}}
\toprule
\textbf{backbone} & \textbf{site} & \textbf{d} & \textbf{effect} & \textbf{$T_F$ Procrustes/ridge/conv-res/random} & \textbf{ridge $T_{RMS}$} & \textbf{ridge proj} & \textbf{ridge win} & \textbf{reach.} \\
\midrule
ResNet-50 & layer1 & 1 & 0.372 & 0.36/0.60/0.64/-0.91 & 0.603 & 0.951 $\pm$ 0.006 & 0.96 & yes \\
ResNet-50 & layer2 & 2 & 0.436 & 0.50/0.59/0.59/-0.82 & 0.592 & 0.884 $\pm$ 0.016 & 0.97 & yes \\
ResNet-50 & layer3 & 3 & 0.550 & 0.39/0.46/0.46/-1.28 & 0.472 & 0.760 $\pm$ 0.010 & 0.96 & yes \\
ResNet-50 & layer4 & 4 & 0.961 & 0.08/0.10/0.09/-0.81 & 0.116 & 0.555 $\pm$ 0.019 & 0.65 & yes \\
ConvNeXt-T & features1 & 1 & 0.272 & 0.30/0.71/0.72/-0.94 & 0.704 & 0.930 $\pm$ 0.005 & 0.97 & yes \\
ConvNeXt-T & features3 & 2 & 0.235 & 0.50/0.66/0.67/-0.97 & 0.655 & 0.902 $\pm$ 0.008 & 0.97 & yes \\
ConvNeXt-T & features5 & 3 & 0.184 & 0.27/0.40/0.39/-1.21 & 0.408 & 0.676 $\pm$ 0.009 & 0.94 & yes \\
ConvNeXt-T & features7 & 4 & 0.863 & 0.05/0.15/0.08/-1.16 & 0.159 & 0.518 $\pm$ 0.007 & 0.74 & yes \\
ViT-B/16 & block2 & 1 & 0.608 & 0.47/0.64/0.64/-0.54 & 0.636 & 0.887 $\pm$ 0.008 & 0.96 & yes \\
ViT-B/16 & block5 & 2 & 0.578 & 0.36/0.46/0.45/-0.48 & 0.458 & 0.690 $\pm$ 0.017 & 0.96 & yes \\
ViT-B/16 & block8 & 3 & 0.644 & 0.13/0.21/0.21/-0.37 & 0.219 & 0.372 $\pm$ 0.025 & 0.93 & yes \\
ViT-B/16 & block11 & 4 & 0.623 & -0.00/-0.00/-0.00/-0.00 & 0.000 & -0.000 $\pm$ 0.000 & 0.04 & NO \\
DINOv2-B/14 & block2 & 1 & 0.454 & 0.49/0.58/0.59/-1.33 & 0.586 & 0.842 $\pm$ 0.005 & 0.96 & yes \\
DINOv2-B/14 & block5 & 2 & 0.453 & 0.33/0.39/0.39/-1.32 & 0.397 & 0.635 $\pm$ 0.007 & 0.95 & yes \\
DINOv2-B/14 & block8 & 3 & 0.367 & 0.15/0.10/-0.12/-1.27 & 0.140 & 0.451 $\pm$ 0.015 & 0.73 & yes \\
DINOv2-B/14 & block11 & 4 & 0.535 & -0.00/-0.00/-0.00/-0.00 & -0.000 & -0.000 $\pm$ 0.000 & 0.05 & NO \\
\bottomrule
\end{tabularx}
\end{table*}

Depth trend (Spearman over sites $\times$ seeds): ResNet-50 Procrustes $-0.39$ ($p=0.21$, n=12), ridge $=-0.97$ ($p=1.4\times10^{-7}$), conv-res $=-0.97$, random $=0.00$; ConvNeXt-T Procrustes $-0.78$ ($p=0.003$), ridge $-0.97$, conv-res $-0.97$, random $-0.73$ ($p=0.007$); ViT-B/16 Procrustes $-0.95$ ($p=9.6\times10^{-5}$, n=9), ridge $-0.95$, conv-res $-0.95$, random $+0.90$ ($p=0.001$); DINOv2-B/14 Procrustes $-0.95$, ridge $-0.95$, conv-res $-0.95$, random $+0.26$ ($p=0.49$).

\begin{table*}[t]
\centering
\tiny
\caption*{\textbf{Table D.5.} Heat $\sigma=2.0$ depth grid, all four backbones, three seeds. Same columns as Table D.4.}
\setlength{\tabcolsep}{2pt}
\begin{tabularx}{\textwidth}{@{}>{\raggedright\arraybackslash}X >{\raggedright\arraybackslash}X >{\raggedright\arraybackslash}X >{\raggedright\arraybackslash}X >{\raggedright\arraybackslash}X >{\raggedright\arraybackslash}X >{\raggedright\arraybackslash}X >{\raggedright\arraybackslash}X >{\raggedright\arraybackslash}X@{}}
\toprule
\textbf{backbone} & \textbf{site} & \textbf{d} & \textbf{effect} & \textbf{$T_F$ Procrustes/ridge/conv-res/random} & \textbf{ridge $T_{RMS}$} & \textbf{ridge proj} & \textbf{ridge win} & \textbf{reach.} \\
\midrule
ResNet-50 & layer1 & 1 & 0.315 & 0.15/0.51/0.60/-1.33 & 0.503 & 0.778 $\pm$ 0.019 & 1.00 & yes \\
ResNet-50 & layer2 & 2 & 0.370 & 0.25/0.47/0.47/-1.17 & 0.471 & 0.791 $\pm$ 0.017 & 1.00 & yes \\
ResNet-50 & layer3 & 3 & 0.466 & 0.24/0.37/0.37/-1.77 & 0.379 & 0.673 $\pm$ 0.014 & 0.96 & yes \\
ResNet-50 & layer4 & 4 & 0.789 & 0.08/0.05/0.05/-1.15 & 0.073 & 0.501 $\pm$ 0.011 & 0.61 & yes \\
ConvNeXt-T & features1 & 1 & 0.300 & 0.20/0.54/0.65/-1.62 & 0.518 & 0.828 $\pm$ 0.024 & 0.99 & yes \\
ConvNeXt-T & features3 & 2 & 0.272 & 0.32/0.55/0.58/-1.61 & 0.549 & 0.807 $\pm$ 0.021 & 1.00 & yes \\
ConvNeXt-T & features5 & 3 & 0.179 & 0.24/0.34/0.33/-1.94 & 0.351 & 0.579 $\pm$ 0.020 & 0.98 & yes \\
ConvNeXt-T & features7 & 4 & 0.647 & 0.04/0.14/0.08/-1.91 & 0.151 & 0.420 $\pm$ 0.013 & 0.77 & yes \\
ViT-B/16 & block2 & 1 & 0.333 & 0.11/0.50/0.50/-1.82 & 0.489 & 0.749 $\pm$ 0.011 & 0.98 & yes \\
ViT-B/16 & block5 & 2 & 0.430 & 0.16/0.27/0.28/-1.75 & 0.265 & 0.445 $\pm$ 0.033 & 0.94 & yes \\
ViT-B/16 & block8 & 3 & 0.435 & 0.09/0.11/0.11/-1.46 & 0.113 & 0.208 $\pm$ 0.024 & 0.84 & yes \\
ViT-B/16 & block11 & 4 & 0.441 & 0.00/0.00/0.00/0.00 & 0.000 & 0.000 $\pm$ 0.000 & 0.06 & NO \\
DINOv2-B/14 & block2 & 1 & 0.358 & -0.14/0.13/0.14/-4.64 & 0.135 & 0.546 $\pm$ 0.049 & 0.73 & yes \\
DINOv2-B/14 & block5 & 2 & 0.293 & 0.06/0.05/0.04/-4.63 & 0.055 & 0.205 $\pm$ 0.003 & 0.67 & yes \\
DINOv2-B/14 & block8 & 3 & 0.317 & 0.02/-0.40/-0.99/-4.47 & -0.369 & 0.214 $\pm$ 0.047 & 0.11 & yes \\
DINOv2-B/14 & block11 & 4 & 0.303 & 0.00/0.00/0.00/0.00 & 0.000 & 0.000 $\pm$ 0.000 & 0.06 & NO \\
\bottomrule
\end{tabularx}
\end{table*}

Depth trend: ResNet-50 Procrustes $-0.32$ ($p=0.30$, n=12), ridge $-0.97$, conv-res $-0.97$, random $+0.13$; ConvNeXt-T Procrustes $-0.39$ ($p=0.21$), ridge $-0.82$ ($p=0.001$), conv-res $-0.97$, random $-0.76$ ($p=0.004$); ViT-B/16 Procrustes $-0.26$ ($p=0.49$, n=9), ridge $-0.95$, conv-res $-0.95$, random $+0.84$ ($p=0.004$); DINOv2-B/14 Procrustes $+0.47$ ($p=0.20$), ridge $-0.95$, conv-res $-0.95$, random $+0.58$ ($p=0.10$).

\begin{table*}[t]
\centering
\tiny
\caption*{\textbf{Table D.6.} $\sigma=1.0$ single-seed heat grid (the mild-dissipation control of \S{}6.1). Same columns as Table D.4.}
\setlength{\tabcolsep}{2pt}
\begin{tabularx}{\textwidth}{@{}>{\raggedright\arraybackslash}X >{\raggedright\arraybackslash}X >{\raggedright\arraybackslash}X >{\raggedright\arraybackslash}X >{\raggedright\arraybackslash}X >{\raggedright\arraybackslash}X >{\raggedright\arraybackslash}X >{\raggedright\arraybackslash}X >{\raggedright\arraybackslash}X@{}}
\toprule
\textbf{backbone} & \textbf{site} & \textbf{d} & \textbf{effect} & \textbf{$T_F$ Procrustes/ridge/conv-res/random} & \textbf{ridge $T_{RMS}$} & \textbf{ridge proj} & \textbf{ridge win} & \textbf{reach.} \\
\midrule
ResNet-50 & layer1 & 1 & 0.167 & 0.21/0.51/0.55/-3.93 & 0.494 & 0.743 & 1.00 & yes \\
ResNet-50 & layer2 & 2 & 0.200 & 0.26/0.44/0.44/-3.47 & 0.442 & 0.692 & 0.98 & yes \\
ResNet-50 & layer3 & 3 & 0.254 & 0.19/0.30/0.30/-5.24 & 0.306 & 0.537 & 0.96 & yes \\
ResNet-50 & layer4 & 4 & 0.437 & -0.02/-0.17/-0.20/-3.54 & -0.137 & 0.339 & 0.21 & yes \\
ConvNeXt-T & features1 & 1 & 0.170 & 0.20/0.60/0.63/-3.98 & 0.567 & 0.812 & 1.00 & yes \\
ConvNeXt-T & features3 & 2 & 0.153 & 0.32/0.53/0.55/-4.30 & 0.499 & 0.709 & 1.00 & yes \\
ConvNeXt-T & features5 & 3 & 0.104 & 0.20/0.29/0.27/-4.16 & 0.281 & 0.454 & 0.99 & yes \\
ConvNeXt-T & features7 & 4 & 0.362 & 0.04/0.07/-0.03/-4.31 & 0.078 & 0.296 & 0.69 & yes \\
ViT-B/16 & block2 & 1 & 0.154 & 0.16/0.50/0.50/-4.94 & 0.493 & 0.669 & 0.98 & yes \\
ViT-B/16 & block5 & 2 & 0.201 & 0.17/0.25/0.26/-4.88 & 0.237 & 0.352 & 0.89 & yes \\
ViT-B/16 & block8 & 3 & 0.203 & 0.07/0.07/0.08/-4.11 & 0.073 & 0.132 & 0.76 & yes \\
ViT-B/16 & block11 & 4 & 0.216 & 0.00/0.00/0.00/0.00 & -0.000 & 0.000 & 0.05 & NO \\
DINOv2-B/14 & block2 & 1 & 0.196 & 0.01/0.11/0.15/-8.63 & 0.135 & 0.463 & 0.64 & yes \\
DINOv2-B/14 & block5 & 2 & 0.168 & 0.02/0.03/0.01/-8.59 & 0.034 & 0.117 & 0.50 & yes \\
DINOv2-B/14 & block8 & 3 & 0.180 & -0.00/-0.14/-0.39/-8.34 & -0.114 & 0.042 & 0.16 & yes \\
DINOv2-B/14 & block11 & 4 & 0.188 & -0.00/-0.00/-0.00/-0.00 & 0.000 & 0.000 & 0.05 & NO \\
\bottomrule
\end{tabularx}
\end{table*}

Trends are stored for ResNet-50 (ridge $\rho=-1.00$, conv-res $\rho=-1.00$, random $\rho=0.00$) and ConvNeXt-T (ridge $\rho=-1.00$, conv-res $\rho=-1.00$, random $\rho=-0.80$) over four points; ViT-B/16 and DINOv2-B/14 trends are \textit{not recorded}.

\subsubsection*{D.3 Readout $\times$ transformation crossed grid}

\begin{table*}[t]
\centering
\scriptsize
\caption*{\textbf{Table D.7.} Raw consumer-visible share of the transformation displacement, per backbone $\times$ site, for the four crossed arms (transformation $\to$ readout; \S{}6.3, Table 11).}
\setlength{\tabcolsep}{2pt}
\begin{tabularx}{\textwidth}{@{}>{\raggedright\arraybackslash}X >{\raggedright\arraybackslash}X >{\raggedright\arraybackslash}X >{\raggedright\arraybackslash}X >{\raggedright\arraybackslash}X >{\raggedright\arraybackslash}X@{}}
\toprule
\textbf{backbone} & \textbf{site} & \textbf{hue$\to$hue} & \textbf{hue$\to$hf} & \textbf{heat$\to$hue} & \textbf{heat$\to$hf} \\
\midrule
ResNet-50 & layer1--4 & 0.127 / 0.062 / 0.025 / 0.003 & 0.029 / 0.020 / 0.011 / 0.004 & 0.008 / 0.008 / 0.008 / 0.004 & 0.046 / 0.032 / 0.014 / 0.004 \\
ConvNeXt-T & features1/3/5/7 & 0.144 / 0.053 / 0.010 / 0.008 & 0.089 / 0.029 / 0.007 / 0.007 & 0.020 / 0.010 / 0.003 / 0.007 & 0.033 / 0.020 / 0.004 / 0.007 \\
ViT-B/16 & blocks 2/5/8/11 & 0.005 / 0.011 / 0.012 / 0.006 & 0.001 / 0.004 / 0.007 / 0.004 & 0.001 / 0.002 / 0.005 / 0.003 & 0.007 / 0.007 / 0.008 / 0.004 \\
DINOv2-B/14 & blocks 2/5/8/11 & 0.025 / 0.021 / 0.001 / 0.011 & 0.006 / 0.009 / 0.000 / 0.009 & 0.014 / 0.009 / 0.000 / 0.007 & 0.026 / 0.011 / 0.000 / 0.007 \\
\bottomrule
\end{tabularx}
\end{table*}

\begin{table*}[t]
\centering
\scriptsize
\caption*{\textbf{Table D.8.} Chance-corrected visible share (visible share divided by the matched-rank random-projector share at the readout's effective rank --- the control that reproduces the Table 11 ordering) and the over-isotropic share, same arms and sites, each cell as corrected / isotropic.}
\setlength{\tabcolsep}{2pt}
\begin{tabularx}{\textwidth}{@{}>{\raggedright\arraybackslash}X >{\raggedright\arraybackslash}X >{\raggedright\arraybackslash}X >{\raggedright\arraybackslash}X >{\raggedright\arraybackslash}X@{}}
\toprule
\textbf{backbone} & \textbf{hue$\to$hue} & \textbf{hue$\to$hf} & \textbf{heat$\to$hue} & \textbf{heat$\to$hf} \\
\midrule
ResNet-50 & 9.01/2.51, 8.35/2.46, 4.26/1.95, 0.99/0.53 & 0.94/0.57, 1.29/0.80, 1.21/0.83, 0.60/0.56 & 0.52/0.15, 1.04/0.31, 1.29/0.60, 1.04/0.56 & 1.43/0.90, 2.03/1.28, 1.60/1.11, 0.68/0.63 \\
ConvNeXt-T & 4.13/1.06, 2.76/0.79, 1.08/0.30, 0.89/0.48 & 2.56/0.66, 1.50/0.43, 0.34/0.20, 0.46/0.42 & 0.62/0.15, 0.48/0.14, 0.36/0.10, 0.71/0.39 & 1.02/0.25, 1.01/0.30, 0.22/0.13, 0.47/0.43 \\
ViT-B/16 & 0.64/0.31, 1.44/0.65, 1.52/0.70, 0.94/0.35 & 0.08/0.07, 0.27/0.22, 0.51/0.42, 0.25/0.23 & 0.12/0.05, 0.31/0.14, 0.69/0.31, 0.51/0.19 & 0.48/0.40, 0.50/0.40, 0.59/0.48, 0.26/0.24 \\
DINOv2-B/14 & 3.29/1.50, 3.48/1.26, 0.10/0.04, 1.47/0.67 & 0.39/0.33, 0.59/0.51, 0.03/0.02, 0.60/0.55 & 1.71/0.81, 1.36/0.52, 0.03/0.01, 0.87/0.40 & 1.84/1.52, 0.77/0.65, 0.01/0.01, 0.47/0.42 \\
\bottomrule
\end{tabularx}
\end{table*}

Trend summary over the four sites (Spearman of the corrected share against depth, per backbone): hue$\to$hue 3/4 backbones decline, median $\rho=-0.80$; hue$\to$hf 2/4, median $+0.00$; heat$\to$hue 1/4, median $+0.30$; heat$\to$hf 4/4, median $\rho=-0.60$ --- Table 11 exactly. Per-arm power gate (effect size of the readout) and full-rank ceiling $T_F$: hue arms $0.739/0.603$, $0.723/0.708$, $0.837/0.639$, $0.618/0.587$; heat arms $0.581/0.528$, $0.523/0.575$, $0.465/0.505$, $0.251/0.160$, in the ResNet-50 / ConvNeXt-T / ViT-B/16 / DINOv2-B/14 order.

\subsubsection*{D.4 Composition grids}

\begin{table*}[t]
\centering
\small
\caption*{\textbf{Table D.9.} Composition at a fixed total hue $90^\circ$, plain CNN stage 1, three seeds.}
\begin{tabularx}{\textwidth}{@{}>{\raggedright\arraybackslash}X >{\raggedright\arraybackslash}X >{\raggedright\arraybackslash}X@{}}
\toprule
\textbf{path} & \textbf{ridge $T_F$} & \textbf{conv-res $T_F$} \\
\midrule
direct fit at $90^\circ$ & 0.285 $\pm$ 0.008 & 0.511 $\pm$ 0.019 \\
$45+45$ & 0.237 $\pm$ 0.013 & 0.472 $\pm$ 0.023 \\
$30+60$ & 0.249 $\pm$ 0.012 & 0.488 $\pm$ 0.027 \\
$60+30$ (order swap) & 0.245 $\pm$ 0.014 & 0.484 $\pm$ 0.028 \\
$22.5\times4$ & 0.253 $\pm$ 0.024 & 0.451 $\pm$ 0.032 \\
single $22.5^\circ$ ($1/4$ of total) & 0.131 $\pm$ 0.019 & 0.207 $\pm$ 0.010 \\
cyclic $+120/-120$ (identity) & 0.067 $\pm$ 0.029 & 0.254 $\pm$ 0.021 \\
\bottomrule
\end{tabularx}
\end{table*}

\begin{table*}[t]
\centering
\small
\caption*{\textbf{Table D.10.} Composition at a fixed total heat $t^\star=1.0$, plain CNN stage 1, three seeds.}
\begin{tabularx}{\textwidth}{@{}>{\raggedright\arraybackslash}X >{\raggedright\arraybackslash}X >{\raggedright\arraybackslash}X@{}}
\toprule
\textbf{path} & \textbf{ridge $T_F$} & \textbf{conv-res $T_F$} \\
\midrule
direct fit at $t^\star$ & 0.610 $\pm$ 0.107 & 0.760 $\pm$ 0.030 \\
two steps & 0.514 $\pm$ 0.132 & 0.679 $\pm$ 0.019 \\
four steps & 0.437 $\pm$ 0.112 & 0.625 $\pm$ 0.063 \\
eight steps & 0.372 $\pm$ 0.114 & 0.481 $\pm$ 0.025 \\
single step ($t^\star/8$) & 0.572 $\pm$ 0.025 & 0.597 $\pm$ 0.025 \\
\bottomrule
\end{tabularx}
\end{table*}

\begin{table*}[t]
\centering
\scriptsize
\caption*{\textbf{Table D.11.} The never-fitted composite under the dissipative semigroup: operators fitted at $t_1,t_2$ only, evaluated at the composed parameter that was never fitted. \textit{direct} is fitted at the composed parameter (upper reference); \textit{single} is the $t_1$ operator evaluated on the composite.}
\setlength{\tabcolsep}{2pt}
\begin{tabularx}{\textwidth}{@{}>{\raggedright\arraybackslash}X >{\raggedright\arraybackslash}X >{\raggedright\arraybackslash}X >{\raggedright\arraybackslash}X >{\raggedright\arraybackslash}X >{\raggedright\arraybackslash}X@{}}
\toprule
\textbf{arm} & \textbf{composite} & \textbf{operator} & \textbf{composed} & \textbf{direct} & \textbf{single} \\
\midrule
plain CNN (3 seeds) & $\sigma$ 0.75+1.0 $\to$ 1.250 & ridge & 0.711 $\pm$ 0.032 & 0.699 $\pm$ 0.033 & 0.296 $\pm$ 0.011 \\
plain CNN (3 seeds) & $\sigma$ 0.75+1.0 $\to$ 1.250 & conv-res & 0.834 $\pm$ 0.023 & 0.829 $\pm$ 0.022 & 0.315 $\pm$ 0.024 \\
plain CNN (seed 0) & $\sigma$ 1.0+1.5 $\to$ 1.803 & ridge & 0.607 & 0.570 & 0.441 \\
plain CNN (seed 0) & $\sigma$ 1.0+1.5 $\to$ 1.803 & conv-res & 0.844 & 0.807 & 0.422 \\
colour-equivariant CNN (seed 0) & $\sigma$ 0.75+1.0 $\to$ 1.250 & ridge & 0.705 & 0.663 & 0.261 \\
colour-equivariant CNN (seed 0) & $\sigma$ 0.75+1.0 $\to$ 1.250 & conv-res & 0.859 & 0.845 & 0.282 \\
\bottomrule
\end{tabularx}
\end{table*}

Power gate of the composite (plain CNN seed 0, $\sigma$ 1.0+1.5): effect size 0.881, top-1 flip 0.795, accuracy 0.915 $\to$ 0.225. Random-orthogonal control at these points: $T_F$ from $-0.76$ to $-0.17$.

\begin{table*}[t]
\centering
\small
\caption*{\textbf{Table D.12.} Path agreement (feature-space consistency between operators of the same family under different totals), pooled over three seeds.}
\begin{tabularx}{\textwidth}{@{}>{\raggedright\arraybackslash}X >{\raggedright\arraybackslash}X >{\raggedright\arraybackslash}X >{\raggedright\arraybackslash}X@{}}
\toprule
\textbf{family / operator} & \textbf{same total} & \textbf{different total} & \textbf{ratio} \\
\midrule
hue, ridge & 0.047 & 0.338 & 7.13 \\
hue, conv-res & 0.085 & 0.316 & 3.72 \\
heat, ridge & 0.246 & 0.859 & 3.49 \\
heat, conv-res & 0.342 & 0.766 & 2.24 \\
\bottomrule
\end{tabularx}
\end{table*}

Stored per-pair examples: hue $45+45$ against $30+60$ $0.020 \pm 0.002$ (ridge); heat two-step against four-step $0.173 \pm 0.070$ (ridge). The global linear family agrees most tightly between paths and transports worst --- the consistency--fidelity split of \S{}5.4.

\subsubsection*{D.5 Attribute and perimeter grids}

\begin{table*}[t]
\centering
\scriptsize
\caption*{\textbf{Table D.13.} Multi-attribute intervention on frozen ViT-B/16 (block 4), three seeds, 180 synthetic scenes each (100 fit / 80 held-out): transfer and projection per operator family, with the attribute's own input power. \textit{transfer} is the fraction of the no-op $\to$ real read-out error removed.}
\setlength{\tabcolsep}{2pt}
\begin{tabularx}{\textwidth}{@{}>{\raggedright\arraybackslash}X >{\raggedright\arraybackslash}X >{\raggedright\arraybackslash}X >{\raggedright\arraybackslash}X >{\raggedright\arraybackslash}X@{}}
\toprule
\textbf{attribute (power)} & \textbf{Procrustes} & \textbf{ridge+MLP} & \textbf{conv-res} & \textbf{random} \\
\midrule
hue (0.288 $\pm$ 0.021) & 0.867 $\pm$ 0.048 / 0.843 & 0.989 $\pm$ 0.016 / 0.939 & 0.989 $\pm$ 0.014 / 0.940 & -0.119 $\pm$ 0.087 / 0.460 \\
saturation (0.213 $\pm$ 0.016) & 0.946 $\pm$ 0.034 / 0.910 & 0.966 $\pm$ 0.010 / 0.982 & 0.971 $\pm$ 0.007 / 0.997 & -2.178 $\pm$ 1.103 / -1.756 \\
value (0.510 $\pm$ 0.036) & 0.947 $\pm$ 0.013 / 0.928 & 0.947 $\pm$ 0.015 / 0.935 & 0.946 $\pm$ 0.013 / 0.934 & -0.137 $\pm$ 0.623 / -0.115 \\
quantity (0.366 $\pm$ 0.002) & 0.606 $\pm$ 0.022 / 0.456 & 0.645 $\pm$ 0.055 / 0.501 & 0.634 $\pm$ 0.055 / 0.500 & -3.630 $\pm$ 2.103 / 4.788 \\
\bottomrule
\end{tabularx}
\end{table*}

Each cell is transfer / projection (projection sd omitted for width; hue Procrustes 0.054, ridge+MLP 0.018, conv-res 0.022, random 0.028). Null probe error $\to$ real probe error: hue $88.10\to18.89$, saturation $0.370\to0.070$, value $0.368\to0.044$, quantity $0.980\to0.256$.

\begin{table*}[t]
\centering
\scriptsize
\caption*{\textbf{Table D.14.} The same protocol on real COCO instance crops (240 regions; both backbones three seeds, fitting on 150 and evaluating on 80, all under the image-grouped split so that no two crops of one image can straddle fit and evaluation). Errors are degrees for hue and attribute units otherwise.}
\setlength{\tabcolsep}{2pt}
\begin{tabularx}{\textwidth}{@{}>{\raggedright\arraybackslash}X >{\raggedright\arraybackslash}X >{\raggedright\arraybackslash}X >{\raggedright\arraybackslash}X >{\raggedright\arraybackslash}X >{\raggedright\arraybackslash}X >{\raggedright\arraybackslash}X >{\raggedright\arraybackslash}X@{}}
\toprule
\textbf{backbone} & \textbf{attribute (power)} & \textbf{err null $\to$ real} & \textbf{Procrustes} & \textbf{ridge+MLP} & \textbf{conv-res} & \textbf{random} & \textbf{random win} \\
\midrule
ViT-B/16 (3) & hue (0.687 $\pm$ 0.009) & 81.8 $\to$ 36.4 & 0.650 $\pm$ 0.061 & 0.561 $\pm$ 0.067 & 0.517 $\pm$ 0.097 & 0.023 $\pm$ 0.472 & 0.482 \\
ViT-B/16 (3) & saturation (0.350 $\pm$ 0.007) & 0.177 $\to$ 0.108 & 0.587 $\pm$ 0.043 & 0.720 $\pm$ 0.046 & 0.719 $\pm$ 0.014 & -0.871 $\pm$ 1.165 & 0.300 \\
ViT-B/16 (3) & value (0.499 $\pm$ 0.005) & 0.170 $\to$ 0.081 & 0.707 $\pm$ 0.107 & 0.857 $\pm$ 0.090 & 0.850 $\pm$ 0.060 & -1.995 $\pm$ 1.327 & 0.154 \\
DINOv2-B/14 (3) & hue (0.404 $\pm$ 0.004) & 83.7 $\to$ 68.6 & 0.576 $\pm$ 0.129 & 0.424 $\pm$ 0.066 & 0.507 $\pm$ 0.138 & -0.411 $\pm$ 0.145 & 0.145 \\
DINOv2-B/14 (3) & saturation (0.238 $\pm$ 0.006) & 0.284 $\to$ 0.233 & 0.592 $\pm$ 0.103 & 0.802 $\pm$ 0.136 & 0.763 $\pm$ 0.145 & -3.160 $\pm$ 2.136 & 0.163 \\
DINOv2-B/14 (3) & value (0.338 $\pm$ 0.005) & 0.217 $\to$ 0.181 & 0.528 $\pm$ 0.050 & 0.623 $\pm$ 0.072 & 0.667 $\pm$ 0.133 & -8.240 $\pm$ 3.346 & 0.137 \\
\bottomrule
\end{tabularx}
\end{table*}

\begin{table*}[t]
\centering
\small
\caption*{\textbf{Table D.15.} Second-attribute chain: identification, geometry and the spatial fallback. The code head reads value and saturation as monotone scalars, so unlike hue neither has an identifiability floor.}
\begin{tabularx}{\textwidth}{@{}>{\raggedright\arraybackslash}X >{\raggedright\arraybackslash}X >{\raggedright\arraybackslash}X@{}}
\toprule
\textbf{block} & \textbf{quantity} & \textbf{value} \\
\midrule
identify & value MAE / max ($\hat l$ vs $v$) & 0.054 / 0.172 \\
identify & saturation MAE / max ($\hat s$ vs $s$) & 0.067 / 0.251 \\
geometry & cross-shape cosine, saturation / value & 0.732 (min 0.553) / 0.718 (min 0.614) \\
geometry & read-out MAE $s$/$v$ at stage 3, stage 5, stage 7 & 0.066/0.035, 0.069/0.038, 0.057/0.033 \\
spatial fallback & global pooled scalar (two objects, gap 0.5) & 0.250 (= the mathematical floor) \\
spatial fallback & object-mask pool / raw local window & 0.146 (0.03--0.35) / 0.454 \\
\bottomrule
\end{tabularx}
\end{table*}

\begin{table*}[t]
\centering
\small
\caption*{\textbf{Table D.16.} Clipped-scaling family (monoid boundary, Table 6 row 3): realised report change / read-out error / clipped fraction, by headroom to the gamut clip.}
\begin{tabularx}{\textwidth}{@{}>{\raggedright\arraybackslash}X >{\raggedright\arraybackslash}X >{\raggedright\arraybackslash}X >{\raggedright\arraybackslash}X@{}}
\toprule
\textbf{attribute} & \textbf{headroom $\ge1$ (n=48)} & \textbf{$0.33$--$1$ (n=24)} & \textbf{$<0.33$ (n=24)} \\
\midrule
saturation & 0.173 / 0.035 / 0.125 & 0.191 / 0.054 / 0.500 & 0.000 / 0.036 / 1.000 \\
value & 0.174 / 0.014 / 0.125 & 0.191 / 0.009 / 0.500 & 0.000 / 0.012 / 1.000 \\
\bottomrule
\end{tabularx}
\end{table*}

The false-confidence control records a mean absolute report change of exactly $0.0$ (n=24) in the fully clipped bin for both attributes. The Table 6 entry (realised report change $0.174$, read-out error $0.0143$) is the value row of the headroom $\ge1$ column.

\begin{table*}[t]
\centering
\scriptsize
\caption*{\textbf{Table D.17.} Spatial-rotation control ($\mathrm{rot}90$ on the YOLO11n feature site, stride 2, 200 images, one seed). The transformation is exact at the truth level ($0.0$ max abs difference on a double rotation); it is \textbf{zero-power} for an invariant head, so no operator is scored: measuring it requires a consumer that reads rotation, which an invariant head does not.}
\setlength{\tabcolsep}{2pt}
\begin{tabularx}{\textwidth}{@{}>{\raggedright\arraybackslash}X >{\raggedright\arraybackslash}X >{\raggedright\arraybackslash}X >{\raggedright\arraybackslash}X >{\raggedright\arraybackslash}X >{\raggedright\arraybackslash}X@{}}
\toprule
\textbf{arm} & \textbf{logit effect} & \textbf{top-1 flip} & \textbf{$\mathrm{rot}90$ residual} & \textbf{aligned shift-4 residual} & \textbf{exactness} \\
\midrule
plain CNN & 0.762 & 0.640 & 1.031 & 0.405 & 0.0 \\
colour-equivariant CNN (hue-equivariant) & 0.669 & 0.610 & 0.951 & 0.364 & 0.0 \\
\bottomrule
\end{tabularx}
\end{table*}

\subsubsection*{D.6 Boundary grids}

\begin{table*}[t]
\centering
\small
\caption*{\textbf{Table D.18.} The 750-region pre-registered operating domain (COCO instance regions, statistic $m$, window $\pm30^\circ$, success at $30^\circ$; fit half 450, test half 300, threshold $m^\star=0.70$ fixed in advance).}
\begin{tabularx}{\textwidth}{@{}>{\raggedright\arraybackslash}X >{\raggedright\arraybackslash}X >{\raggedright\arraybackslash}X@{}}
\toprule
\textbf{route / statistic} & \textbf{value} & \textbf{95\% CI} \\
\midrule
$m^\star$; median $m$; fraction $m\ge m^\star$ & 0.70; 0.824; 0.644 & --- \\
test half, aligned: success $m\ge m^\star$ vs $m<m^\star$ & 0.770 vs 0.532; gap 0.238 & [0.128, 0.345] \\
test half, ground-truth-box / raw unaligned route & gap 0.188 / 0.085 & not recorded \\
ROC AUC aligned / gtbox / raw & 0.660 / 0.662 / 0.579 & aligned [0.588, 0.725] \\
fit half, aligned & 0.805 vs 0.551; gap 0.254 & --- \\
controlled calibration (fit on controlled domain) & AUC 1.0 & [1.0, 1.0] \\
\bottomrule
\end{tabularx}
\end{table*}

\begin{table*}[t]
\centering
\small
\caption*{\textbf{Table D.19.} The unaligned 120-region covariate control (5-fold CV; AUC for success classification at thresholds 10 and 20). Concentration alone carries almost no power; the covariates do the work.}
\begin{tabularx}{\textwidth}{@{}>{\raggedright\arraybackslash}X >{\raggedright\arraybackslash}X >{\raggedright\arraybackslash}X >{\raggedright\arraybackslash}X@{}}
\toprule
\textbf{covariate set} & \textbf{AUC (thr 10)} & \textbf{AUC (thr 20)} & \textbf{gain over concentration} \\
\midrule
concentration $m$ alone & 0.523 & 0.480 & --- \\
$+$ mask covariates & 0.754 & 0.680 & +0.231 / +0.199 \\
$+$ category & 0.651 & 0.616 & +0.127 / +0.135 \\
\bottomrule
\end{tabularx}
\end{table*}

Regression targets are near zero for all three sets (CV $R^2$: $-0.018$, $+0.010$, $-0.095$), so the control bounds ranking power, not calibrated error.

\subsubsection*{D.7 Cross-cutting note}

The depth grid and the gate-change diagnosis agree. On ResNet-50 the consumer-visible crossing share rises with depth from 0.220 (hue) and 0.252 (heat) at layer 2 to 0.627 and 0.516 at layer 4, while the gate flip fraction falls over the same range (0.091 and 0.094 to 0.064 and 0.051). Over the six (site, algebra) points, Spearman against $T_F$ is $-0.829$ for the raw gate-change share, $-0.886$ for the consumer-visible gate-change share, and $+0.486$ ($p=0.33$, n.s.) for the flip fraction.

\subsubsection*{D.8 Realisation-error and metric diagnostics}

\begin{table*}[t]
\centering
\scriptsize
\caption*{\textbf{Table D.20.} The two sources of realisation error measured \textit{inside the fit split} --- the in-split diagnostic whose held-out version is Table 13 of the main text: the within-cell term, the cost of sharing one operator across the realised transition cells, and the share of the within-cell term contributed by the off$\to$on cell (where the source activation is zero while the target varies).}
\setlength{\tabcolsep}{2pt}
\begin{tabularx}{\textwidth}{@{}>{\raggedright\arraybackslash}X >{\raggedright\arraybackslash}X >{\raggedright\arraybackslash}X >{\raggedright\arraybackslash}X >{\raggedright\arraybackslash}X >{\raggedright\arraybackslash}X >{\raggedright\arraybackslash}X@{}}
\toprule
\textbf{site} & \textbf{hue within-cell} & \textbf{hue sharing (share of total)} & \textbf{heat within-cell} & \textbf{heat sharing (share of total)} & \textbf{off$\to$on share of within (hue / heat)} & \textbf{units} \\
\midrule
\texttt{layer1} & 0.0836 & 0.2000 (71\%) & 0.1583 & 0.4655 (75\%) & 6\% / 17\% & 1135.0 \\
\texttt{layer2} & 0.0789 & 0.2085 (73\%) & 0.1104 & 0.3485 (76\%) & 8\% / 15\% & 2615.0 \\
\texttt{layer3} & 0.0535 & 0.1697 (76\%) & 0.0669 & 0.2458 (79\%) & 13\% / 20\% & 5640.0 \\
\texttt{layer4} & 0.0131 & 0.3133 (96\%) & 0.0112 & 0.3622 (97\%) & 59\% / 63\% & 8834.0 \\
\bottomrule
\end{tabularx}
\end{table*}

\begin{table*}[t]
\centering
\scriptsize
\caption*{\textbf{Table D.21.} The closure criterion instantiated on measured features: the carrier of the known input hue against the estimated synthesis $B$, with the relative reconstruction tail, the closure defect of $B$ against $A_\Delta$, and the held-out relative error of a global linear map fitted on half the hues; the last three columns repeat the analysis with the $\sin2h$ coordinate dropped.}
\setlength{\tabcolsep}{2pt}
\begin{tabularx}{\textwidth}{@{}>{\raggedright\arraybackslash}X >{\raggedright\arraybackslash}X >{\raggedright\arraybackslash}X >{\raggedright\arraybackslash}X >{\raggedright\arraybackslash}X >{\raggedright\arraybackslash}X >{\raggedright\arraybackslash}X@{}}
\toprule
\textbf{site} & \textbf{tail} & \textbf{closure defect} & \textbf{held-out global fit} & \textbf{tail (no $\sin2h$)} & \textbf{defect (no $\sin2h$)} & \textbf{held-out (no $\sin2h$)} \\
\midrule
YOLO11n stage 3 & 0.046 & 3.2e-15 & 0.282 & 0.060 & 2.0e-15 & 0.334 \\
ResNet-18 layer 2 & 0.064 & 3.5e-15 & 0.413 & 0.087 & 2.3e-15 & 0.395 \\
\bottomrule
\end{tabularx}
\end{table*}

\begin{table*}[t]
\centering
\scriptsize
\caption*{\textbf{Table D.22.} Metric distortion of the truth action in the site metric and in the consumer's response, ResNet-50, four depths, both algebras: median and p90 of the pairwise ratio of feature distances after and before the transformation, over 120 COCO crops with a 60-crop subsample; the consumer column is the same at every site because the consumer is the network's own output.}
\setlength{\tabcolsep}{2pt}
\begin{tabularx}{\textwidth}{@{}>{\raggedright\arraybackslash}X >{\raggedright\arraybackslash}X >{\raggedright\arraybackslash}X >{\raggedright\arraybackslash}X >{\raggedright\arraybackslash}X@{}}
\toprule
\textbf{site} & \textbf{hue: site metric (median / p90)} & \textbf{heat: site metric (median / p90)} & \textbf{hue: consumer response} & \textbf{heat: consumer response} \\
\midrule
layer1 & $1.067$ / $1.376$ & $0.847$ / $0.989$ & $1.056$ & $1.015$ \\
layer2 & $0.995$ / $1.094$ & $0.800$ / $0.900$ & $1.056$ & $1.015$ \\
layer3 & $0.974$ / $1.035$ & $0.877$ / $0.946$ & $1.056$ & $1.015$ \\
layer4 & $1.027$ / $1.180$ & $0.990$ / $1.140$ & $1.056$ & $1.015$ \\
\bottomrule
\end{tabularx}
\end{table*}

\subsubsection*{D.9 Direct test of the fibre condition}

\begin{table*}[t]
\centering
\scriptsize
\caption*{\textbf{Table D.23.} The fibre condition tested directly at the consumer and at each site: the fraction of base-equal pairs that remain equal after the transformation, at tolerance quantiles $q=0.01$ and $q=0.05$ of the base pair-distance distribution (the random-pair baseline is $q$ by construction), with the transformation's effect size on the consumer and the per-site rates at $q=0.01$. Shallow to deep, left to right.}
\setlength{\tabcolsep}{2pt}
\begin{tabularx}{\textwidth}{@{}>{\raggedright\arraybackslash}X >{\raggedright\arraybackslash}X >{\raggedright\arraybackslash}X >{\raggedright\arraybackslash}X >{\raggedright\arraybackslash}X >{\raggedright\arraybackslash}X@{}}
\toprule
\textbf{backbone} & \textbf{family} & \textbf{effect size} & \textbf{consumer $q=0.01$} & \textbf{consumer $q=0.05$} & \textbf{per-site $q=0.01$ (four sites, shallow$\to$deep)} \\
\midrule
ResNet-50 & hue & 0.724 & 0.327 & 0.499 & 0.59 / 0.67 / 0.63 / 0.31 \\
ResNet-50 & heat & 0.594 & 0.595 & 0.655 & 0.61 / 0.57 / 0.51 / 0.61 \\
ConvNeXt-T & hue & 0.708 & 0.454 & 0.599 & 0.55 / 0.56 / 0.31 / 0.45 \\
ConvNeXt-T & heat & 0.517 & 0.577 & 0.612 & 0.55 / 0.51 / 0.35 / 0.18 \\
ViT-B/16 & hue & 0.812 & 0.249 & 0.332 & 0.67 / 0.70 / 0.59 / 0.38 \\
ViT-B/16 & heat & 0.465 & 0.710 & 0.727 & 0.62 / 0.62 / 0.70 / 0.56 \\
DINOv2-B/14 & hue & 0.659 & 0.463 & 0.609 & 0.74 / 0.76 / 0.69 / 0.57 \\
DINOv2-B/14 & heat & 0.262 & 0.815 & 0.879 & 0.75 / 0.80 / 0.75 / 0.82 \\
\bottomrule
\end{tabularx}
\end{table*}

\section{Method boxes}

Implementation-level specifications for the four components behind the paper's construction and measurement claims. Each box states its hyperparameters, what is fitted and what is not. Section numbers are those of the main text. Every value is copied from a file read while writing this appendix.

\subsubsection*{Box E1. Local phase-conditioned transporter (\S{}5.3)}

\textbf{Partition.} The hue circle is divided into nine segments of $40^\circ$, with starts $h\in\{0^\circ,40^\circ,\dots,320^\circ\}$; a segment operator is trained on the paired endpoints $z(h)\to z(h+40^\circ)$.

\textbf{Per-segment generator.} In a $D=24$ PCA subspace of the site's features, each segment carries an antisymmetric generator $A_k\in\mathfrak{so}(24)$, fitted so that $\exp(A_k\cdot40^\circ)\,z_1\approx z_2$ on the paired endpoints, using the four training shapes and all four $(s,v)$ pairs.

\textbf{Smoothing and read-out at inference.} The nine per-segment generators are fitted by a third-order Fourier field, $\Omega(\theta)=\Omega_0+\sum_{k=1}^{3}\big(C_k\cos(k\omega\theta)+S_k\sin(k\omega\theta)\big)$ with $\omega=2\pi/360$, one coefficient matrix per basis element by least squares on the nine sampled phases; the fit reaches $R^2=0.8672$. Phases are shared across shapes, amplitudes are not, so the phase is read from the representation at inference and the segment generator is applied as a fixed function of the read phase: a step of size $\delta\le40^\circ$ is taken with the midpoint value $\Omega(\theta+\delta/2)$, $z\mapsto\exp(\Omega(\theta+\delta/2)\,\delta)\,z$. No operator is refitted on the test shape or the test starting state.

\textbf{Fitting budget.} Adam, 800 epochs, learning rate $5\times10^{-3}$, minibatch 64, nine segments; the $D=24$ subspace and the Fourier coefficients are fixed after fitting. The locality comparison uses a separate per-segment rank-32 log-domain operator $z+U\big((e^{\lambda}-1)\odot(U^{\top}z)\big)$ with $U\in\mathbb{R}^{d\times32}$ (the site width; $d=64$ at stage 3), Adam, 150 epochs, learning rate $8\times10^{-3}$, trained on the four training shapes and tested on the two holdout shapes; across five sites it beats the copy baseline by $1.55$--$15.94\times$ while the global family reaches only $0.08$--$0.84\times$. The local construction closes a full $360^\circ$ loop to a relative error of $0.0037$ on stage 3 and $0.0062$ on ResNet-18 layer 2.

\textbf{Parameter count, and what is not fitted.} Each generator has $D(D-1)/2=276$ free parameters in the $24$-dimensional subspace, and the smoothed field stores $1+2K=7$ coefficient matrices; this is the entire fitted object. The backbone is frozen; the action applied at inference is a fixed function of the read phase, not a fitted operator; no test-sample operator, template or per-shape calibration is fitted.

\subsubsection*{Box E2. The carrier interface: learned encoding, fixed action (\S{}7.2)}

\textbf{Read-in.} $E_\theta:\mathbb{R}^{d}\to\mathbb{R}^6$ is a three-layer MLP of widths $64\to256\to256\to6$ with SiLU activations, applied to the stage 3 GAP feature divided by the training-pool standard deviation ($\texttt{feat\_std}=0.7240$). Its six outputs are read as $l=\mathrm{softplus}(o_0)$, $\hat s=\mathrm{sigmoid}(o_1)$ and $(z_{1x},z_{1y},z_{2x},z_{2y})=o_{2:6}$, so the code is $(l,\hat s,z_1,z_2)$ with $z_1$ the fundamental and $z_2$ the second harmonic.

\textbf{Fixed action.} $A_\Delta=\mathrm{diag}(I_2,R(\Delta),R(2\Delta))$: the two scalar channels are held; the fundamental rotates by $\Delta$ and the second harmonic by $2\Delta$. $R$ is the standard plane rotation; the action carries no parameters and is never fitted, so only $E_\theta$ is learned.

\textbf{Objective and anchoring.} The head is trained on rendered $(s,v,h)$ cells against the target $(v,\,s,\,m^\star\cos\theta,\,m^\star\sin\theta,\,m^\star\cos2\theta,\,m^\star\sin2\theta)$, with $\theta$ the nominal hue and $m^\star$ the measured per-$(s,v)$ orbit amplitude; the loss is the sum of mean-squared errors on the two scalar channels and the four harmonic coordinates. $m^\star$ is the amplitude anchor: it fixes the phase gauge by tying the code's radial scale to the measured orbit, which is what makes the read-out well posed. Where no semantic label is available the same six channels are additionally anchored on label-free pixel statistics --- the saturation-weighted dominant hue, saturation and value --- with only the code head trained and the backbone frozen. Hyperparameters: AdamW, learning rate $2\times10^{-3}$, weight decay $10^{-4}$, 1500 epochs, 24 minibatches per epoch, batch 96; trained on four shapes, evaluated zero-shot on the two unseen shapes.

\textbf{Scope.} The action is fixed by the measured structure, so the correspondence is learned but its law is not. Whether a fitted action of equal capacity does better is measured in \S{}7.2 (Table F.10); the licence for the fixed action is that the two conditions the construction uses are measured to hold at the site (\S{}4.3, \S{}7.3). The two constructions differ in kind: the local transporter of Box E1 fits a generator and its Fourier coefficients and freezes them at inference, whereas the action here is never fitted and only the read-in is learned. One caveat belongs to the runtime figure of Box E4: the region-level timing artefact does not record its device, and the later verification runs are CPU-only, so the ratio is the reported measurement rather than a controlled head-to-head benchmark.

\subsubsection*{Box E3. Consumer heads}

\textbf{ImageNet head.} For the depth grid and the transformer probes the consumer is the backbone's own frozen pretrained classifier head (torchvision \texttt{IMAGENET1K\_\allowbreak{}V2}/\texttt{V1}, DINOv2 with a frozen linear head on final pooled features). The criterion is label-free: it compares $F(Wz)$ with $F(g_\tau z)$, so no label enters, and the effect size --- the relative change of these logits under the real transformation --- licenses reading any operator result.

\textbf{Attribute read-outs.} Hue, saturation and value are read by ridge probes fitted on the pooled final features of the base and variant images for that attribute; one probe is shared by every route so no operator is advantaged. The YOLO detector arm reads a 64-dimensional stage 3 GAP feature. The hue read-out reported in \S{}7.4 is the code head of Box E2, not a ridge probe.

\textbf{YOLO hue-bin detector.} YOLO11n is trained on a synthetic set whose class \textit{is} the hue bin: eight bins of $45^\circ$ with the shape randomised across classes. Recipe: 60 epochs at $224$ px, batch 16, with every augmentation (mosaic, mixup, hue, saturation, value, horizontal and vertical flip) disabled, seed $0$; validation mAP50 $0.9871$ (mAP50-95 $0.9094$). Reading convention: box-level F1 at IoU $0.5$ against the detection obtained after a real recolour of the same scene, with a mid-layer transport at stage 3, 80 held-out scenes and three seeds. At $\Delta=90^\circ$ the null (no intervention) reaches F1 $0.008$, Procrustes $0.778$, ridge $0.905$, conv-res $0.965$ and the random orthogonal control random $0.000$; AP50 tracks F1 throughout.

\subsubsection*{Box E4. Hardware and runtime}

Two configurations were used. The original training and measurement runs used a single NVIDIA GeForce RTX 2060 (6 GB); the device string is recorded with the timing artefact. The later verification runs are CPU-only, launched on four threads at low priority, which is the configuration recorded in the usage lines of the composition, split-control, and figure-composition scripts.

\begin{table*}[t]
\centering
\small
\caption*{\textbf{Table E1.} Runtime figures already reported in the main text.}
\begin{tabularx}{\textwidth}{@{}>{\raggedright\arraybackslash}X >{\raggedright\arraybackslash}X >{\raggedright\arraybackslash}X@{}}
\toprule
\textbf{operation} & \textbf{time} & \textbf{ratio} \\
\midrule
region-level code shift (both regions) & $20.89$ $\mu$s & $1825\times$ versus pixel route \\
pixel re-colour + re-forward (same regions) & $38.13$ ms & reference \\
backbone full forward, $224$ px, batch 1 & $22.92$ ms & --- \\
pixel-side HSV shift, $224$ px & $15.21$ ms & --- \\
code rotation, batched, per object & $27.6$ ns & $1.38\times10^{6}\times$ versus pixel re-colour \\
\bottomrule
\end{tabularx}
\end{table*}

Two caveats are part of the figure. First, the ratio is the cost of the \textit{operation}, not a whole-system speed-up: the region-level code shift edits an already-extracted code, while the pixel route re-renders and re-forwards. Second, the pixel-route device is recorded as the RTX 2060; the region-level artefact does not record its device, and the later verification runs are CPU-only, so the two columns of the first row are not guaranteed to be the same machine and the ratio should be read as the reported measurement, not as a controlled benchmark.

\section{Capability tables per setup}

One small table per experimental setup, giving the setup name, what was fitted, what was held out, and the primary capability statistic with its baseline. Values are those of the main text's Tables 14 and 15 unless the row states otherwise.

\begin{table*}[t]
\centering
\scriptsize
\caption*{\textbf{Table F.1.} CIFAR-10 ResNet-44 causal arms (five arms; stage 1; hue $\Delta=90^\circ$; three seeds). \textit{Fitted:} one global or content-conditioned channel-mixing operator per shift at the mid-layer site. \textit{Held out:} images disjoint from the fit --- \textbf{marked cell:} the raws store 1000 fit / 400 held-out for the plain-CNN arm at stage 1 (all three seeds) and for seed 0 at stage 2, but 300 fit / 200 held-out for the other four arms and for plain-CNN seeds 1--2 at stage 2, so the five-arm stage-1 comparison is not at a matched fitting budget for the plain-CNN row (the stage-2 part of the same disclosure is in the per-arm files named below). The primary statistic is the conv-res paired win rate against the no-op; the no-op is the baseline by construction, and the random orthogonal control is given for comparison.}
\setlength{\tabcolsep}{2pt}
\begin{tabularx}{\textwidth}{@{}>{\raggedright\arraybackslash}X >{\raggedright\arraybackslash}X >{\raggedright\arraybackslash}X >{\raggedright\arraybackslash}X >{\raggedright\arraybackslash}X@{}}
\toprule
\textbf{arm} & \textbf{effect size} & \textbf{conv-res win (Procrustes / ridge / ridge+MLP)} & \textbf{conv-res mean gain} & \textbf{random win} \\
\midrule
plain CNN & 0.389 & \textbf{0.871} (0.798 / 0.723 / 0.789) & +0.274 & 0.005 \\
$+$ hue augmentation & 0.141 & --- (0.532 / 0.527 / ---) & --- & 0.000 \\
CEConv & 0.335 & \textbf{0.975} (0.957 / 0.965 / ---) & +0.316 & 0.007 \\
LCER & 0.013 & --- (0.273 / 0.312 / ---) & --- & 0.000 \\
code-auxiliary arm & 0.369 & --- (0.812 / 0.692 / ---) & --- & 0.027 \\
\bottomrule
\end{tabularx}
\end{table*}

The two low-effect arms (0.141 and 0.013) are the effect-size collapse of \S{}3.7: the test is unpowered there and the correct reading is that the question has no content, not that the representation is deficient.

\begin{table*}[t]
\centering
\scriptsize
\caption*{\textbf{Table F.2.} Five-arm fair benchmark, CIFAR-10, 20 epochs $\times$ three seeds; every arm at an identical budget. \textit{Fitted:} end-to-end classification only. \textit{Held out:} the standard CIFAR-10 test split and the out-of-distribution colour sweep. The baseline is the plain CNN. This is a calibration table, not an accuracy claim.}
\setlength{\tabcolsep}{2pt}
\begin{tabularx}{\textwidth}{@{}>{\raggedright\arraybackslash}X >{\raggedright\arraybackslash}X >{\raggedright\arraybackslash}X >{\raggedright\arraybackslash}X >{\raggedright\arraybackslash}X >{\raggedright\arraybackslash}X@{}}
\toprule
\textbf{arm} & \textbf{params} & \textbf{FLOPs/img} & \textbf{clean acc} & \textbf{OOD sweep mean} & \textbf{min acc over sweep} \\
\midrule
plain CNN & 2.64M & 0.78G & 0.9050 $\pm$ 0.0014 & 0.8195 $\pm$ 0.0008 & 0.778 \\
$+$ hue augmentation & 2.64M & 0.78G & 0.9035 $\pm$ 0.0024 & \textbf{0.8904 $\pm$ 0.0034} & 0.870 \\
CEConv (rot = 8) & 2.58M & 5.96G & 0.8907 $\pm$ 0.0006 & 0.8227 $\pm$ 0.0056 & 0.786 \\
LCER ($G=8$) & 2.57M & 5.99G & 0.8822 $\pm$ 0.0051 & 0.8815 $\pm$ 0.0047 & \textbf{0.876} \\
code-auxiliary (ours) & 2.65M & 0.78G & \textbf{0.9082 $\pm$ 0.0016} & 0.8258 $\pm$ 0.0017 & 0.784 \\
\bottomrule
\end{tabularx}
\end{table*}

The code auxiliary's clean-accuracy edge is within seed noise and is not an accuracy claim; the OOD ordering is an ordering of invariance built at training time, not of the interface.

\begin{table*}[t]
\centering
\scriptsize
\caption*{\textbf{Table F.3.} Frozen transformer grid (ViT-B/16 supervised and DINOv2-B/14 self-supervised; split after block 4; hue $\Delta=60^\circ$; 200 held-out images; three seeds). \textit{Fitted:} the operator family at block 4 with a frozen linear head on the final pooled features. \textit{Held out:} the 200 evaluation images; the split and head are fixed, so only the conv-res residual varies across seeds. Baseline: the no-op; the contrast between the two rows is a power contrast, not an architectural one.}
\setlength{\tabcolsep}{2pt}
\begin{tabularx}{\textwidth}{@{}>{\raggedright\arraybackslash}X >{\raggedright\arraybackslash}X >{\raggedright\arraybackslash}X >{\raggedright\arraybackslash}X >{\raggedright\arraybackslash}X >{\raggedright\arraybackslash}X@{}}
\toprule
\textbf{backbone} & \textbf{effect size} & \textbf{top-1 flip} & \textbf{Procrustes proj / win} & \textbf{ridge+MLP proj / win} & \textbf{conv-res proj / win} \\
\midrule
ViT-B/16 (supervised) & 0.79 & 0.51 & 0.757 / 0.780 & \textbf{0.890} / 0.773 & 0.849 / 0.777 \\
DINOv2-B/14 (self-supervised) & 0.13 & 0.075 & 0.366 / 0.660 & 0.459 / 0.665 & 0.301 (0.261--0.354) / 0.677 \\
\bottomrule
\end{tabularx}
\end{table*}

\begin{table*}[t]
\centering
\small
\caption*{\textbf{Table F.4.} Colour-sensitive detector (YOLO11n trained on eight hue-bin classes, shapes randomised across classes, mAP50 0.987; split at stage 3; 40 fit / 80 held-out scenes; three seeds). \textit{Fitted:} a mid-layer transport operator per hue shift. \textit{Held out:} the 80 scenes, and --- in the second block --- the shift itself, with $90^\circ$ never fitted. Primary statistic: box AP50, baseline the no-intervention (null) route.}
\begin{tabularx}{\textwidth}{@{}>{\raggedright\arraybackslash}X >{\raggedright\arraybackslash}X >{\raggedright\arraybackslash}X >{\raggedright\arraybackslash}X@{}}
\toprule
\textbf{route} & \textbf{AP50 ($\Delta=90^\circ$)} & \textbf{F1@IoU0.5 ($\Delta=90^\circ$)} & \textbf{baseline (null)} \\
\midrule
Procrustes (Procrustes) & 0.785 $\pm$ 0.027 & 0.778 $\pm$ 0.029 & 0.003 / 0.008 \\
least-squares (ridge) & 0.906 $\pm$ 0.031 & 0.905 $\pm$ 0.030 & 0.003 / 0.008 \\
$+$ $3\times3$ residual (conv-res) & \textbf{0.969 $\pm$ 0.013} & \textbf{0.965 $\pm$ 0.013} & 0.003 / 0.008 \\
random orthogonal (random) & 0.000 $\pm$ 0.000 & 0.000 $\pm$ 0.000 & --- \\
\bottomrule
\end{tabularx}
\end{table*}

Never-fitted shift ($90^\circ$): a single-parameter generator extrapolation reaches AP50 0.3654 and a fitted-$30^\circ/60^\circ$ chain 0.511, against 0.8052 for an operator fitted directly at $90^\circ$ (the seen-shift reference). Here a random orthogonal operator attains projection coefficients 0.32--0.46 while its detection F1 is 0.000 at every shift.

\begin{table*}[t]
\centering
\small
\caption*{\textbf{Table F.5.} Real-image task consumer (small CNN task head trained end-to-end on real COCO crops to identify the value level $0.60/1.00/1.45$, an exact transform; 150 fit / 90 held-out regions). \textit{Fitted:} the task head and the stage-2 feature intervention. \textit{Held out:} 90 regions; the label is the task's own. Primary statistic: mean paired gain over the no-op, with its 95\% CI; baseline gain $0$.}
\begin{tabularx}{\textwidth}{@{}>{\raggedright\arraybackslash}X >{\raggedright\arraybackslash}X >{\raggedright\arraybackslash}X >{\raggedright\arraybackslash}X@{}}
\toprule
\textbf{route} & \textbf{mean gain (95\% CI)} & \textbf{projection} & \textbf{win rate} \\
\midrule
ridge+MLP (per-cell MLP) & \textbf{0.579} [0.509, 0.653] & 0.860 & 0.967 \\
conv-res (content $+$ neighbourhood) & 0.576 [0.512, 0.643] & 0.838 & 1.000 \\
random (random orthogonal) & -0.180 [-0.468, 0.047] & 0.745 & 0.622 \\
\bottomrule
\end{tabularx}
\end{table*}

Power: task accuracy over the three value levels 0.796; relative displacement of the real transform 0.956; top-1 flip 0.700.

\begin{table*}[t]
\centering
\scriptsize
\caption*{\textbf{Table F.6.} Real-region usage rule (750 pre-registered COCO instance regions, 737 images; fit half 450 / test half 300; statistic $m$ = saturated-colour share within $\pm30^\circ$ of the circular-mean hue; threshold $m^\star=0.70$ fixed in advance; success at $30^\circ$). \textit{Fitted:} the threshold on the fit half only. \textit{Held out:} the test half. Primary statistic: test-half success rate above versus below threshold; baseline is the below-threshold half.}
\setlength{\tabcolsep}{2pt}
\begin{tabularx}{\textwidth}{@{}>{\raggedright\arraybackslash}X >{\raggedright\arraybackslash}X >{\raggedright\arraybackslash}X >{\raggedright\arraybackslash}X >{\raggedright\arraybackslash}X@{}}
\toprule
\textbf{route} & \textbf{success $m\ge m^\star$} & \textbf{success $m<m^\star$} & \textbf{gap (95\% CI)} & \textbf{ROC AUC} \\
\midrule
aligned route & \textbf{0.770} & 0.532 & 0.238 [0.128, 0.345] & 0.660 \\
ground-truth-box & 0.848 & 0.661 & 0.188 & 0.662 \\
raw, unaligned & 0.277 & 0.193 & 0.085 & 0.579 \\
\bottomrule
\end{tabularx}
\end{table*}

Unaligned 120-region control: concentration alone AUC 0.523 / 0.480 (thresholds 10 / 20), $+$ mask covariates 0.754 / 0.680, $+$ category 0.651 / 0.616.

\begin{table*}[t]
\centering
\small
\caption*{\textbf{Table F.7.} Carrier interface: generalisation and ablation. \textit{Fitted:} the read-in $E_\theta$ only; the action $A_\Delta=\mathrm{diag}(I,R(\Delta),R(2\Delta))$ (six dimensions for colour) is fixed by the measured structure and never fitted. \textit{Held out:} the two unseen shapes (pentagon, hexagon) and the hues $\{20^\circ,260^\circ\}$ excluded from the read-in fit; on real regions, the region. Primary statistic: zero-shot hue read-out error, baseline the copy route (a 60$^\circ$ error) and the input-space interpolation baseline.}
\begin{tabularx}{\textwidth}{@{}>{\raggedright\arraybackslash}X >{\raggedright\arraybackslash}X >{\raggedright\arraybackslash}X@{}}
\toprule
\textbf{capability} & \textbf{primary} & \textbf{baseline} \\
\midrule
zero-shot hue read-out, 64-cell $s,\allowbreak{}v$ grid & 59 of 64 cells $\le30^\circ$; median 3.4$^\circ$; saturation floor 2.14$^\circ$ & copy 60$^\circ$; random phase 90$^\circ$ \\
$\Delta=40^\circ$ equivariance residual & 2.36$^\circ$ & --- \\
feature-space augmentation (sparse hue training) & 1.3$^\circ$ & input-space interpolation 3.5$^\circ$; real-only 25.5$^\circ$ \\
real COCO region read-out under the usage rule & 10.0$^\circ$ median (non-circular control) & raw composite 69.5$^\circ$ \\
\bottomrule
\end{tabularx}
\end{table*}

\begin{table*}[t]
\centering
\small
\caption*{\textbf{Table F.8.} The two strongest applications, each against ground truth and its own baseline, with the measured cost. \textit{Fitted:} the fixed action on the learned code; no re-forward. \textit{Held out:} the region-level shifts are verified against a real pixel recolour re-forwarded on the same regions.}
\begin{tabularx}{\textwidth}{@{}>{\raggedright\arraybackslash}X >{\raggedright\arraybackslash}X >{\raggedright\arraybackslash}X >{\raggedright\arraybackslash}X@{}}
\toprule
\textbf{application} & \textbf{metric} & \textbf{baseline} & \textbf{result} \\
\midrule
region-level shift, two regions in one pass & wall-clock per operation & pixel re-render $38.1$ ms & \textbf{20.9 $\mu$s ($1825\times$)} \\
detector box agreement at $\Delta=90^\circ$ & F1@IoU0.5 & null $0.008$ & \textbf{0.965 structured; 0.000 random} \\
composition to a never-fitted parameter & $T_F$ at the composite & direct fit at that parameter & composed $0.844$ vs direct $0.807$ at $\sigma=1.803$ (Table D.11) \\
zero-shot hue read-out & median per-cell error & copy baseline & 3.4$^\circ$ synthetic / 8.0$^\circ$ real high-concentration (10.0$^\circ$ non-circular) \\
\bottomrule
\end{tabularx}
\end{table*}

\begin{table*}[t]
\centering
\small
\caption*{\textbf{Table F.9.} Hardware configuration.}
\begin{tabularx}{\textwidth}{@{}>{\raggedright\arraybackslash}X >{\raggedright\arraybackslash}X@{}}
\toprule
\textbf{phase} & \textbf{configuration} \\
\midrule
original training runs (CIFAR arms, depth grid, detector, interface) & single GPU, NVIDIA GeForce RTX 2060, 6 GB \\
later verification runs (existence test and controls) & CPU-only, four threads at low priority, GPU deliberately untouched \\
\bottomrule
\end{tabularx}
\end{table*}

\begin{table*}[t]
\centering
\scriptsize
\caption*{\textbf{Table F.10.} The fixed, measured action against learnable actions under identical supervision. One read-in ($64\to256\to256\to6$), one code supervision, 1500 epochs $\times$ 24 minibatches $\times$ 96 code samples, train shapes 0--3 and unseen shapes 4--5; the three learnable arms additionally see paired increments drawn from the same hue labels. Errors are medians over the 226 unseen-shape cells, averaged over three seeds. \textit{Increment} is the error of the increment the arm applies against the true increment, at $37^\circ$ and $90^\circ$, neither of them fitted; \textit{composition defect} compares the two-step composite with the direct step.}
\setlength{\tabcolsep}{2pt}
\begin{tabularx}{\textwidth}{@{}>{\raggedright\arraybackslash}X >{\raggedright\arraybackslash}X >{\raggedright\arraybackslash}X >{\raggedright\arraybackslash}X >{\raggedright\arraybackslash}X >{\raggedright\arraybackslash}X >{\raggedright\arraybackslash}X >{\raggedright\arraybackslash}X@{}}
\toprule
\textbf{arm} & \textbf{action} & \textbf{action params} & \textbf{read-out} & \textbf{increment $37^\circ$} & \textbf{increment $90^\circ$} & \textbf{composition defect} & \textbf{composite vs truth} \\
\midrule
fixed (ours) & $A_\Delta=\operatorname{diag}(I_2,\allowbreak{}R(\Delta),\allowbreak{}R(2\Delta))$, never fitted & 0 & \textbf{1.45$^\circ$} & \textbf{0.00$^\circ$} & \textbf{0.00$^\circ$} & \textbf{0.00$^\circ$} & \textbf{0.00$^\circ$} \\
learned generator & $\exp(\Delta G)$, $G\in\mathbb R^{6\times6}$ free & 36 & 2.10$^\circ$ & 1.31$^\circ$ & 1.02$^\circ$ & 0.00$^\circ$ & 1.02$^\circ$ \\
learned linear action & $I+\Delta B$, $B\in\mathbb R^{6\times6}$ free & 36 & 5.76$^\circ$ & 1.20$^\circ$ & 9.80$^\circ$ & 9.69$^\circ$ & 1.34$^\circ$ \\
generic conditioned map & $c+\mathrm{MLP}([c,\allowbreak{}\cos\Delta,\allowbreak{}\sin\Delta])$, 8-256-256-6 & 69,638 & 1.72$^\circ$ & 1.32$^\circ$ & 2.59$^\circ$ & 2.47$^\circ$ & 2.02$^\circ$ \\
\bottomrule
\end{tabularx}
\end{table*}

\section{Positioning against the four literature lines}

The paper's questions --- existence, identifiability, realisation form, dynamics, construction --- decompose across four lines of work, and no single line contains the join. This appendix states, line by line, what each supplies, what it does not contain, and where a reader should look for the nearest collision with our claims. It is a table, not an argument; the argument is \S{}2.

\begin{table*}[t]
\centering
\scriptsize
\caption*{\textbf{Table G.1.} Positioning table. "Nearest collision" names the work a reviewer is most likely to raise against the corresponding claim of this paper, with the reason it does not anticipate it.}
\setlength{\tabcolsep}{2pt}
\begin{tabularx}{\textwidth}{@{}>{\raggedright\arraybackslash}X >{\raggedright\arraybackslash}X >{\raggedright\arraybackslash}X >{\raggedright\arraybackslash}X >{\raggedright\arraybackslash}X >{\raggedright\arraybackslash}X >{\raggedright\arraybackslash}X@{}}
\toprule
\textbf{line} & \textbf{representative works} & \textbf{central question in that line} & \textbf{what it supplies to this paper} & \textbf{what it does not contain} & \textbf{our increment} & \textbf{nearest collision} \\
\midrule
\textbf{Equivariance measurement} & Lenc and Vedaldi [1]; Gruver et al. [2]; Bruintjes et al. [3]; Romero and Lohit [4] & how equivariant is a given representation, and can a probe-space map make it more so? & the empirical object --- a frozen representation whose response to a known transformation can be measured --- and the practice of scoring it against a re-rendered reference & no information condition: equivariance is scored, never asked to exist; no account of which part of the algebra the score conflates; no composition behaviour & existence is an information condition on the encoder's fibres (\S{}3.2), priced exactly in the linear layer (Thms 1--2) and with the fixed-linear class's score exact in the rectifier layer (Props C--E); and a family of operators can be perfectly self-consistent while failing the transformation (\S{}5.4) & Gruver et al.'s Lie-derivative score is a \textit{local} equivariance error; it is silent on the global existence question, and our Proposition D shows the local quantity it measures belongs to the gate-change term, whose role in the failure is decided by the region-transition criterion \\
\textbf{Symmetry-based and equivariant architectures} & Cohen and Welling [9]; Weiler and Cesa [10]; Lengyel et al. [11]; Yang et al. [12]; Yang et al. [13]; Higgins et al. [15] & how do we \textit{build} a representation with a prescribed symmetry, and what can such a network express? & the representation-theoretic candidate forms --- invariant blocks and harmonic rotation blocks (Prop, \S{}3.6) --- and the demonstration that exact architectural equivariance is achievable when the symmetry is known a priori & the transformation must be known and built in; nothing about the structure a \textit{pretrained, opaque} network already carries; the group-versus-semigroup distinction is assumed rather than measured & the measured structure \textit{is} the blueprint: the paper reads the organisation off frozen networks and then builds the interface from the measurement (\S{}4, \S{}7), with the action never fitted & CEConv and its successors assume colour equivariance is desirable and impose it; we show that on the colour instance it is \textit{already approximately present} in ordinary backbones, and measure where it is not \\
\textbf{Colour and perceptual transformations} & Koenderink [17]; Duits et al. [18]; colour-appearance literature & how should a colour or scale transformation be represented analytically so that it is well behaved? & the reference transformations themselves --- the hue rotation and the diffusion semigroup --- and their algebraic character as implemented physics & no network-side condition; nothing about whether a learned representation carries the transformation, composes it, or can be edited through it & the same transformations used as ground-truth probes on frozen networks, with the algebra (group vs semigroup) as the \textit{independent variable} of the measurement & the dissipative family is the sharp case: the truth action composes to $1.8\times10^{-7}$, yet no basis tested carries a shared decay rate, so the carrier fails for a reason the analytic theory alone does not predict (L15) \\
\textbf{Causal abstraction and latent linearisation} & Geiger et al. [19], [20]; Park et al. [21]; Nadaf [22]; Venkatesh and Kurapath [23] & when do high-level interventions correspond to low-level ones, and when is a latent direction a well-defined intervention? & the interventional-consistency frame, of which our fibre condition is the transformation-action specialisation, and the caution that latent directions are not identifiable without a stated reference & single interventions rather than a whole algebra; no composition law and no price for violating it; no realisation-form question; no construction & the \textit{whole} algebra must descend (\S{}3.2), composition is priced quantitatively (Thm 4), identifiability is carried by the kernel and stabilisers of the representation (\S{}3.1), and the question continues to family capacity (Thm 3) and construction (\S{}7) & our fibre condition is formally Geiger et al.'s consistency condition specialised to transformation actions --- we say so in \S{}2.4 and claim only the specialisation's additional content: the algebra, the pricing, and the construction \\
\textbf{Dynamical / Koopman-style linearisation} & Koopman-style embedding work as surveyed in \S{}2.4 & when does a nonlinear system admit an invariant finite-dimensional subspace on which the evolution is linear? & the idea that a linear operator on an embedding can stand for a nonlinear action, which our band-limited global family instantiates in truncated form & the subspace is \textit{designed or learned} and the dynamics are temporal; the line does not ask whether a \textit{given, frozen} representation admits a linear realisation, nor what obstructs one & the realisation question is \textit{decided} for the given representation, not assumed: existence is an iff condition (Thm 1), the closure criterion decides which carrier coordinates can carry a fixed action (Props F--G), the capacity of the natural per-dimension class is bounded before fitting (Thm 3), and the failure is attributed to a measurable term (Prop D) & a truncated Koopman operator on features is one of our rejected global families (\S{}5.2) unless the orbit is band-limited --- the measured spectrum ($k\le2$ at $84$--$88\%$) is what makes the truncated form work \\
\bottomrule
\end{tabularx}
\end{table*}

\subsubsection*{G.1 Where this paper does \textit{not} compete}

Three boundaries are worth stating so that the positioning is not read as a claim of priority.

\setcounter{enumi}{0}
\begin{enumerate}
\item \textbf{Architecture design.} The paper does not propose a new layer, loss, or training scheme, and no claim here depends on one. The interface of \S{}7 is a read-out and a fixed action on \textit{existing} features; where it is trained (the read-in $E_\theta$), it is trained under ordinary supervision, and the paper's contribution is that the \textit{form} of the action is fixed by measurement rather than learned.
\item \textbf{Perceptual theory.} The hue rotation and the diffusion semigroup are used as exact, re-renderable references (Assumption A1); the paper takes their physical validity in the rendered domain as given, and treats questions of colour appearance, spectral rendering, or perceptual uniformity as out of scope (\S{}2.3).
\item \textbf{A universal claim about all networks.} Every general statement here is a statement about encoders and algebras under stated assumptions (Appendix B); every \textit{quantitative} statement is a measurement on the site and family it names. The claim--scope table (\S{}3.7) and Appendix A fix this per row, and \S{}8.3 states the two places where the theory's magnitude does not carry over.
\end{enumerate}

\end{document}